\documentclass[journal]{IEEEtran}

\usepackage[T1]{fontenc}
\usepackage[utf8]{inputenc}
\usepackage{url}
\usepackage{amsmath,amssymb,amsfonts}
\usepackage{mathtools}
\usepackage{bm}
\usepackage{graphicx}
\usepackage{array,multirow,makecell,tabularx,booktabs}
\usepackage{float}
\usepackage[caption=false,font=footnotesize]{subfig}
\newcolumntype{Y}{>{\raggedright\arraybackslash}X}

\usepackage[numbers,sort&compress]{natbib}

\usepackage[algo2e,ruled,vlined,linesnumbered]{algorithm2e}
\SetKwInOut{Inputs}{Inputs}
\SetKwInOut{Output}{Output}
\SetAlgoHangIndent{1em}
\SetAlgoInsideSkip{smallskip}
\DontPrintSemicolon

\usepackage{tikz}
\usetikzlibrary{arrows.meta,positioning,fit}

\usepackage[hidelinks,bookmarksopen,bookmarksnumbered]{hyperref}
\providecommand\BibTeX{{\rmfamily B\kern-.05em \textsc{i\kern-.025em b}\kern-.08em
T\kern-.1667em\lower.7ex\hbox{E}\kern-.125emX}}

\title{Variable-Resolution Virtual Map for Autonomous Exploration of Unmanned Surface Vehicles (USVs)}

\author{Ye~Li, Yewei~Huang, Yongchang~Xie, Wenlong~GaoZhang,\\
Alberto~Quattrini~Li, Brendan~Englot, and Yuanchang~Liu%
\thanks{Ye Li, Yongchang Xie, Wenlong GaoZhang, and Yuanchang Liu are with the Department of Mechanical Engineering, University College London, London, UK.}%
\thanks{Yewei Huang is with the Department of Mechanical Engineering, Stevens Institute of Technology, Hoboken, NJ, USA, and the Department of Computer Science, Dartmouth College, Hanover, NH, USA.}%
\thanks{Alberto Quattrini Li is with the Department of Computer Science, Dartmouth College, Hanover, NH, USA.}%
\thanks{Brendan Englot is with the Department of Mechanical Engineering, Stevens Institute of Technology, Hoboken, NJ, USA.}%
\thanks{Corresponding author: Yuanchang Liu, Department of Mechanical Engineering, University College London, London, UK (e-mail: yuanchang.liu@ucl.ac.uk).}}

\begin{document}
\markboth{VRVM for USVs}{Li et al.: VRVM for USVs}
\maketitle

\begin{abstract}
Autonomous exploration by unmanned surface vehicles (USVs) in near-shore waters requires reliable localisation and consistent mapping over extended areas, but this is challenged by GNSS degradation, environment-induced localisation uncertainty, and limited on-board computation. 
Virtual map–based methods explicitly model localisation and mapping uncertainty by tightly coupling factor-graph SLAM with a map uncertainty criterion.
However, their storage and computational costs scale with fixed-resolution workspace discretisations, leading to inefficiency in large near-shore environments.
Moreover, overvaluing feature-sparse open-water regions can increase the risk of SLAM failure as a result of inbalance between exploration and exploitation.
To address these limitations, we propose a Variable-Resolution Virtual Map (VRVM), a computationally efficient method for representing localisation uncertainty using sparse Gaussian virtual landmarks placed on an adaptive quadtree. 
The adaptive quadtree enables an area-weighted uncertainty representation that keeps coarse, far-field
virtual landmarks deliberately uncertain while allocating higher resolution to information-dense regions, and reduces the sensitivity of the map valuation to local refinements of the tree. 
An expectation–maximisation (EM) planner is adopted to evaluate pose and map uncertainty along frontiers using the VRVM, balancing exploration and exploitation.
We evaluate VRVM against several state-of-the-art exploration algorithms in the VRX
Gazebo simulator, using a high-fidelity marina environment across different testing scenarios with an increasing level of exploration difficulties. 
We further assess its exploration planning capabilities through real-world pond experiments with operator-controlled navigation.
The results indicate that our method offers safer
behaviour and better utilisation of on-board computation in GNSS-degraded near-shore environments.

\end{abstract}


\begin{figure*}[!t]
    \centerline{\includegraphics[width=\linewidth]{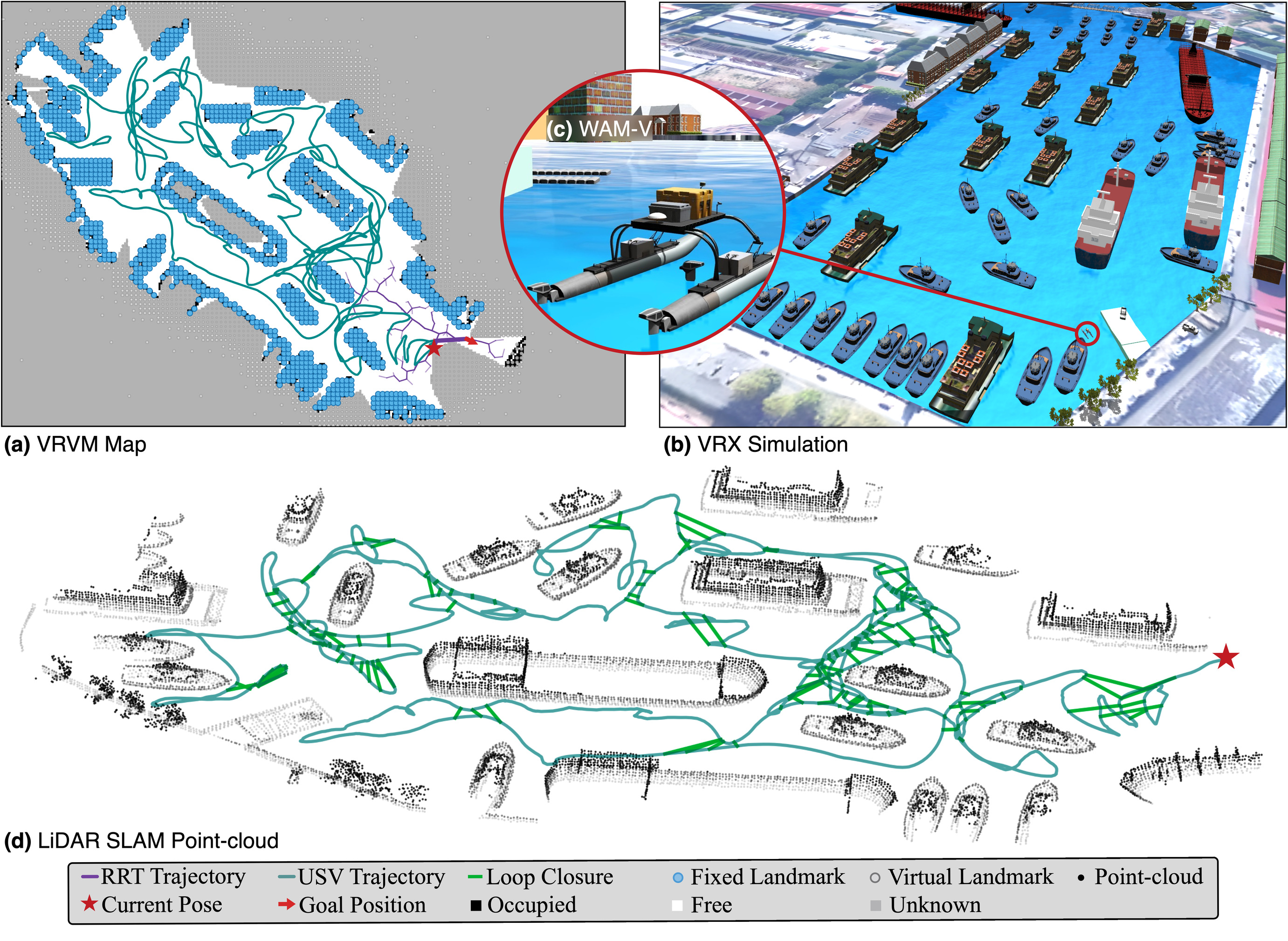}}
    \caption{Illustration of the WAM-V (c) in the $210\,\mathrm{m} \times 500\,\mathrm{m}$ Harbour Basin scene (b). The VRVM map (a) and corresponding LiDAR SLAM point cloud (d) are shown. The USV is depicted in red, with the trajectory history in green. White cells indicate observed space, and gray cells denote unobserved space. Ellipses represent the covariance associated with each virtual landmark.
}
    \label{fig:vrvm}
\end{figure*}

\section{INTRODUCTION}
Autonomous exploration of near-shore waters is vital for applications including harbour inspection, canal surveying, and berth monitoring, particularly because these environments pose major localisation challenges for unmanned surface vehicles (USVs).
Human-made structures, such as quay walls, bridges, and port superstructures, obstruct or reflect global navigation satellite system (GNSS) signals, resulting in significant biases and outages~\citep{makar2023harbour_gnss,pandele2020maritime}.
Moreover, harbour operations impose strict time constraints on USV missions, as vessel traffic can change rapidly and the presence of large moving vessels introduces significant safety risks, requiring missions to be completed both efficiently and accurately.

In such GNSS-degraded near-shore environments, USVs must rely on simultaneous localisation and mapping (SLAM) for pose estimation and mapping. 
However, extended waterways and open basins are often feature-sparse~\citep{wang2024usv_laserslam,marchel2020slam_nav_aids}, and the available geometric structure is highly uneven—dense and stable near shorelines and quays, but largely absent in interior open-water regions. 
This imbalance increases the risk of SLAM drift during open-water traversal, while exhaustive surveying of low-information regions wastes valuable mission time under strict operational constraints.
As a result, exploration strategies in these environments must jointly consider exploration efficiency and the robustness of reliable localisation and mapping.

Existing exploration strategies only partially address this regime.
Early frontier-based~\citep{yamauchi1997frontier, stachniss2004activeloopclosing, leung2006active} and next-best-view (NBV)~\citep{kriegel2015efficient, bircher2016rhnbv} methods rely on simple heuristic utility functions to enable fast exploration, but do not explicitly model localisation or mapping uncertainty.
As a result, when executed in near-shore environments, these methods often overvalue long excursions into open water, accumulating localisation drift and producing blurred or inconsistent map overlap, which leads to repeated and inefficient traversals.

Most information-theoretic methods~\citep{bourgault2002information, carlone2010application, charrow2015csqmi, zhang2020fsmi, asgharivaskasi2025riemannian} consider entropy associated with both localization and mapping, but operating directly on probabilistic belief representations is computationally expensive.
As a result, these approaches often rely on simplifying independence assumptions, such as weak coupling between the robot state and the map or independence among map grid cells.
Consequently, their performance degrades as long-range correlations accumulate in large environments, and may lead to localization drift in environments with strongly uneven geometric structures.

Supervised learning–based methods~\citep{bai2017toward, cao2025dare} accelerate decision making by learning from pre-computed, time-consuming utility functions. However, they introduce additional data collection and training overhead, often rely on GPU-class hardware for deployment, and raise generalisation concerns in the highly variable geometries typically found around ports and canals.
Deep reinforcement learning–based methods~\citep{chaplot2020anslam, cheng2025asymmetric} select goals in the robot’s vicinity and improve the exploration–exploitation trade-off through predictive look-ahead. Nevertheless, they are prone to local optima~\citep{placed2023survey}, and incorporating belief-space uncertainty into fixed-dimensional network inputs remains challenging.

Instead of evaluating map entropy based on an occupancy grid map, the virtual map quantifies localisation uncertainty over the visited area using marginal covariances associated with robot state estimates and the corresponding sensing model~\citep{wang2019autonomous}.
The localisation covariance of virtual landmarks is propagated to predict how future observations affect localisation uncertainty using an expectation–maximisation (EM) framework.
Experimental results demonstrate that the virtual map achieves a balance between map uncertainty reduction and exploration efficiency on both ground robots~\citep{wang2019virtualmap} and underwater unmanned vehicles~\citep{wang2022virtual}.
However, a uniform virtual map is less effective for USVs operating in unevenly structured coastal environments. In such settings, a grid-structured virtual map can lead to overly conservative behaviour when the vehicle repeatedly explores structure-free regions. Computational resources are expended maintaining low-entropy virtual landmarks in feature-poor areas, while high-uncertainty regions are not prioritised in a manner consistent with the vehicle’s actual sensing capability.

Built on virtual map, we propose a variable-resolution virtual map (VRVM).
While VRVM is applicable to various robotic platforms, it is specifically designed for USVs exploring large near-shore scenes, where uneven structures require a careful balance between exploration and localisation stability.
VRVM maintains Gaussian virtual landmarks on a quadtree and refines only uncertainty- or occupancy-ambiguous regions inside the sensor range.
This makes the per-cycle cost scale with the size of the observable region rather than with the  map discretisation. 
We introduce an area-weighted map valuation that reduces dependence on the current split pattern and avoids spuriously rewarding trajectories that pass through feature-sparse water.
On top of this representation, we use an EM planner that selects frontiers by combining trajectory uncertainty, predicted reductions in virtual-map uncertainty, and path cost.

As shown in Fig.~\ref{fig:vrvm}, we evaluate the VRVM algorithm by benchmarking it against a range of existing algorithms in the VRX Gazebo simulator~\citep{saldarriaga2025vrx}, using a LIO-SAM backbone~\citep{shan2020liosam}, across several representative near-shore environments. VRVM achieves a better balance between computational cost and the accuracy–coverage trade-off compared with other planners in these structurally uneven scenes. This is followed by full-scale field trials for real-world demonstration.
We summarise our main contributions as follows:
\begin{itemize}
\item To the best of our knowledge, VRVM is the first virtual map implementation achieving real-time performance on embedded systems.
\item An area-weighted map valuation strategy that balances exploration and exploitation in structurally unbalanced nearshore coastal environments.
\item Demonstration of sustained VRVM-based exploration planning during two 39-minute pond field trials with operator-controlled navigation, and 1.5 hours of autonomous exploration in a realistic $1000,\mathrm{m} \times 1000,\mathrm{m}$ near-coastal simulation environment.
\end{itemize}
Our code for the proposed VRVM framework %
will be publicly released upon acceptance. The remainder of this paper is organised as follows. Section~\ref{sec:related} reviews existing autonomous exploration algorithms. Section~\ref{sec:problem} describes the graph-based SLAM backbone and the virtual map representation. Section~\ref{sec:vrvm} presents the proposed VRVM algorithm and VRVM-based planner. Section~\ref{sec:experiments} reports the experimental results, followed by a discussion of the scope and limitations in Section~\ref{sec:discussion}.
Section~\ref{sec:conclusions} concludes the paper.

\section{Related Work} 
\label{sec:related}

Autonomous exploration for USVs relies on three core components: maintaining reliable localisation under degraded GNSS, constructing an accurate and consistent map, and extending coverage into previously unexplored water. 
Recent work has concentrated primarily on the first two.
Most methods~\citep{shen2023usvfusion, engstrom2022lidarslam, sawada2023shipslam} use LiDAR–SLAM to support USV localisation under GNSS-degraded or GNSS-denied conditions, particularly in inland waterways and harbour-docking scenarios. However, these approaches typically follow prescribed routes or berthing manoeuvres, and do not explicitly address trajectory-level decision making in marine environments.

From a decision making perspective, a wide range of planning and exploration methods have been proposed.
\citet{yamauchi1997frontier} proposes the frontier-based approach, in which the robot navigates towards the boundary between known and unknown regions in an occupancy map.
Originally developed for scene reconstruction in computer vision~\citep{connolly1985determination, border2024surface}, the next-best-view (NBV) planner selects viewpoints by maximising a notion of information (visibility) gain~\citep{bircher2016rhnbv}.
Next-best-trajectory (NBT) methods~\citep{lindqvist2024tree} extend NBV planners by constructing a sampling-based trajectory tree and selecting the feasible branch that maximizes cumulative information gain.
These methods are straightforward and provide fast and good coverage, and have been used on UGVs~\citep{wang2024exploration, zheng2025aage}, UAVs~\citep{best2024multi, zhang2024falcon} and USVs~\citep{song2024efficient}.
However, they do not explicitly model mapping or localisation quality, and may lead to an overestimation of information gain and the accumulation of localisation error in feature-poor environments.

Information-theoretic methods formulate exploration over a unified belief of the robot state and the map.
\citet{bourgault2002information} are among the earliest works to evaluate candidate actions using mutual information (MI) defined over an occupancy grid map.
\citet{carlone2010application, vallve2014dense, vallve2015active, carrillo2018autonomous,  popovic2020informative} further consider entropy-based criteria to balance mapping and localisation utilities.
\citet{saulnier2020information} and \citet{oleynikova2017voxblox} use signed distance fields for map representation to better align with Gaussian measurement models.
These methods discretise the workspace into dense, uniform-resolution grids and evaluate entropy-based utilities on these representations.
\citet{chen2024adaptive} formulate USV information gathering as a non-stationary Gaussian Process regression problem to model spatial variability in uneven offshore environments.

However, computing or approximating MI on dense maps or Gaussian Process posteriors is computationally expensive.
To improve efficiency, \citet{charrow2015csqmi} and \citet{zhang2020fsmi} propose Cauchy–Schwarz Quadratic Mutual Information (CSQMI) and Fast Shannon Mutual Information (FSMI), respectively, to accelerate MI computation.
\citet{Nelson-2018-120036} use the Information Bottleneck principle in conjunction with CSQMI to restrict the growth of computational and memory requirements.
Nevertheless, such efficiency improvements do not change the fact that information-theoretic methods still rely on dense grid representations that scale poorly in storage and inference, while the inconsistency between occupancy-grid-based map entropy and marginal covariance-based robot state entropy makes balancing mapping and localisation utilities difficult, particularly in the absence of explicit forward propagation of the full SLAM belief.

Virtual map-based methods~\citep{wang2019autonomous,wang2019virtualmap, wang2022virtual} are considered information-theoretic methods using expectation-maximisation (EM)~\citep{placed2023survey}.
The main difference between virtual map and other information-theoretic methods is the unified description of map and localisation uncertainty using covariance.
\citet{wang2019autonomous} first propose a virtual map tightly coupled with a landmark-based factor-graph SLAM system.
\citet{wang2019virtualmap} further update the virtual map to make it compatible with a pose-graph SLAM backbone and validates it in indoor environments.
The robustness of this algorithm has been demonstrated in underwater scenarios~\citep{wang2022virtual} and extended to planning~\citep{colladogonzalez2024vmauv} and multi-robot settings~\citep{huang2024multi}.
However, these methods still inherit the common disadvantages of uniform grid representations.

Various learning-based and deep reinforcement learning–based methods have also been proposed. Among them, supervised learning methods~\citep{bai2017toward, chen2019autonomous, cao2025dare} rely on expert demonstrations generated using hand-crafted utility functions, and train deep neural networks, graph neural networks or diffusion models to imitate the resulting trajectories.
While these approaches significantly reduce online computation and enable efficient long-horizon planning, their performance is fundamentally bounded by the quality and structure of the underlying expert policies.

Deep reinforcement learning methods learn a policy through interaction with the environment.
\citet{zhu2018deep} combine deep reinforcement learning with NBV planning to guide high-level exploration decisions.
\citet{chaplot2020anslam} present an active SLAM framework in which exploration behaviour is learned via reinforcement learning, with rewards defined purely by map coverage.
\citet{cao2024deep} introduce a frontier-driven exploration planner that reasons over an informative graph representation using an attention-based policy.
\citet{chen2025gleam} adopt a semantic map representation and learn exploration strategies that balance travel cost and exploration gain.
\citet{li2025learning} reason over a compact heterogeneous graph to enable scalable long-horizon exploration.
\citet{vutetakis2025active} learn a non-myopic NBV policy using active perception network, rather than information gain evaluation.
Despite their architectural differences, these methods rely on simplified reward formulations that primarily account for travel cost and exploration rate, without explicitly modeling localisation and mapping uncertainty. 
Furthermore, all of these approaches are trained and evaluated in indoor environments, leaving their scalability and robustness in large-scale, unstructured harbor environments an open question.

In summary, existing methods either focus on rapid exploration of the environment, or attempt to balance exploration and exploitation without explicitly accounting for strongly uneven geometric structure and the computational burden introduced by entropy-based utility evaluation.
This gap motivates the proposed variable-resolution virtual map (VRVM), which balances exploration and exploitation by selectively modelling uncertainty over the visible region and reweighting virtual map information to favour structurally informative areas during exploration.

\section{Problem Formulation}\label{sec:problem}
\begin{figure*}[!t]
    \centerline{\includegraphics[width=.95\linewidth]{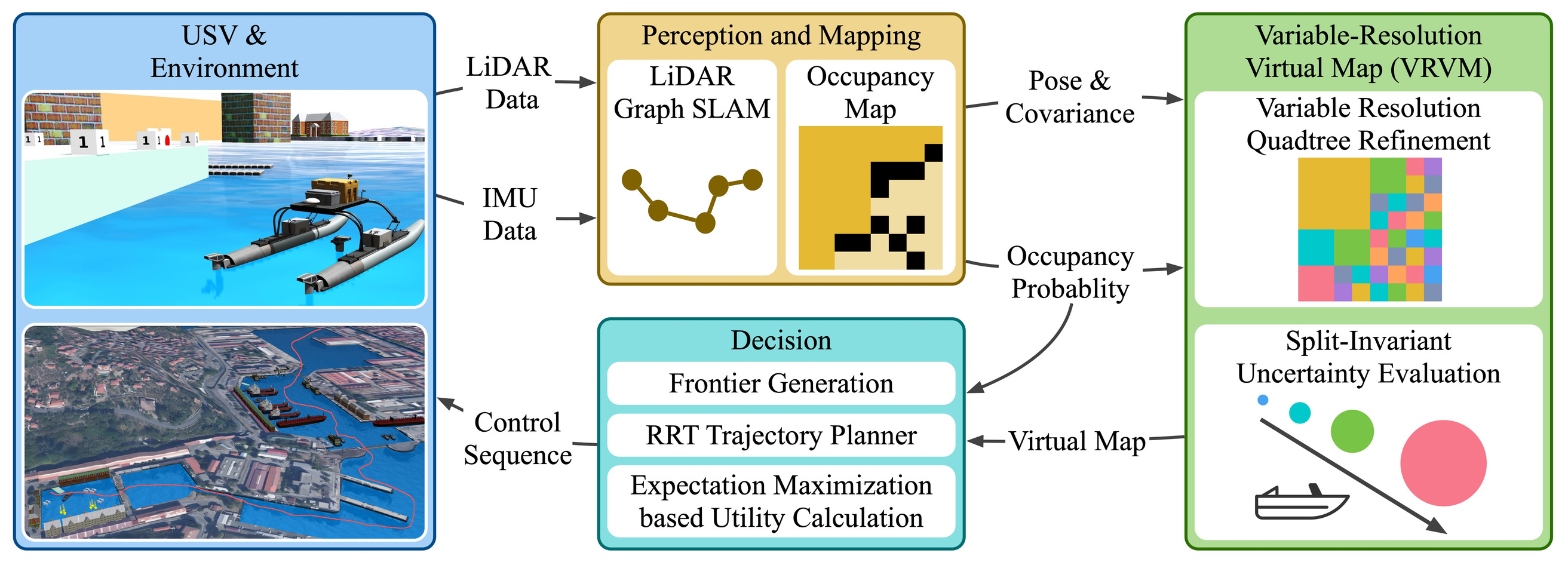}}
    \caption{Overview of the navigation pipeline. The factor graph produced by LiDAR-based Graph SLAM is used to construct the variable-resolution virtual map (VRVM), which provides uncertainty-aware information for decision making. (More complex)}
    \label{fig:pipeline} 
\end{figure*}

We consider an autonomous exploration problem that is tightly coupled with simultaneous localisation and mapping (SLAM) using factor-graph optimisation.
The USV explores within a fixed 2D workspace $\mathcal{W} \subset \mathbb{R}^2$ defined in the world frame $\{W\}$,  and carries LiDAR and IMU sensors.
The overall goal for the USV is to traverse the workspace, obtain the required coverage of nearby structures, and estimate their geometry with high mapping accuracy.
During the exploration, the robot decides its own incremental goal and plans the path accordingly.
Once the current goal is reached, a new set of goal candidates (\emph{frontiers}) is selected from both the boundary of the explored region and the structure-rich areas within the known map. 
This selection strategy accounts for the localisation uncertainty introduced by sensor noise and environmental variations, such as wind and hydrodynamic effects, and enables the USV to balance \emph{exploration} (gaining additional coverage) with \emph{exploitation} (revisiting known regions to support loop closures).
An overview of the exploration pipeline is presented in Fig.~\ref{fig:pipeline}.

\subsection{Robot and Sensor Model}
\label{sec:sensor}
The robot considered in this paper is a USV equipped with an IMU and a LiDAR sensor. 
We denote the USV state (position and heading) at time step $t$ by $\mathbf{x}_{t} = [\,x_t,\; y_t,\; \psi_t\,]^\top \in\mathbb{R}^3$.
$x_i$ and $y_i$ denote the USV position in the world frame $\{W\}$ and $\psi_i$ is the orientation.
The USV motion model is defined as:
\begin{equation}
    \mathbf{x}_{t+1} = f(\mathbf{x}_{t}, \mathbf{u}_{t}) + \mathbf{w}_{t},
\end{equation}
where $\mathbf{u}_{t}$ denotes the control input and $\mathbf{w}_{t} \sim \mathcal{N}(\mathbf{0}, \mathbf{Q}_{t})$ is the zero-mean process noise with covariance $\mathbf{Q}_{t}$.
$f(\cdot)$ denotes the discrete-time motion model.
 
Two types of sensor measurements are considered: the IMU odometry measurement 
$\mathbf{z}^{\mathrm{imu}}_{t}$ and the LiDAR measurement $\mathbf{z}^{\mathrm{lidar}}_{t}$.
The IMU odometry measurement model is defined as:
\begin{equation}
    \mathbf{z}^{\mathrm{imu}}_{t} 
        = h_{\mathrm{imu}}(\mathbf{x}_{t}, \mathbf{x}_{t+1}) 
        + \mathbf{n}^{\mathrm{imu}}_{t},
\end{equation}
where $\mathbf{n}^{\mathrm{imu}}_{t} \sim \mathcal{N}(\mathbf{0}, \mathbf{R}_{t}^{\mathrm{imu}})$ is zero-mean IMU measurement noise with covariance $\mathbf{R}_{t}^{\mathrm{imu}}$. 
The function $h_{\mathrm{imu}}(\cdot)$ returns the relative motion between two consecutive states obtained through IMU preintegration~\citep{forster2016manifold}.
The LiDAR sensor measurement model is defined as
\begin{equation}
    \mathbf{z}^{\mathrm{lidar}}_{t}
        = h_{\mathrm{lidar}}(\mathbf{x}_{t}, \mathcal{M})
        + \mathbf{n}^{\mathrm{lidar}}_{t},
\end{equation}
where $\mathbf{n}^{\mathrm{lidar}}_{t} \sim \mathcal{N}(\mathbf{0},\, \mathbf{R}_{t}^{\mathrm{lidar}})$ 
is zero-mean LiDAR measurement noise with covariance $\mathbf{R}_{t}^{\mathrm{lidar}}$.
The function $h_{\mathrm{lidar}}(\cdot)$ returns the LiDAR ranges at pose $\mathbf{x}_{t}$ given the occupancy map $\mathcal{M}$.
Thus, the inverse sensor model for LiDAR $g_{\mathrm{lidar}}(\cdot)$ is given by:
\begin{equation}
    p(\mathcal{M} \mid \mathbf{z}^{\mathrm{lidar}}_{t}, \mathbf{x}_{t})
    \propto 
    g_{\mathrm{lidar}}\!\left(\mathbf{z}^{\mathrm{lidar}}_{t}, \mathbf{x}_{t}\right),
    \label{eq:lidar}
\end{equation}

\subsection{Factor-Graph-Based SLAM}

\label{sec:slam}
A factor-graph-based SLAM framework is used to estimate the USV pose and its 
uncertainty, and to provide uncertainty criteria for both the virtual map and 
the EM planner.
At time step $t$, let $\mathcal{X}_{0:t} = \{\mathbf{x}_0, \dots, \mathbf{x}_t\}$ 
denote the sequence of USV states up to time $t$. 
Similarly, let $\mathcal{Z}_{0:t} = \mathcal{Z}^{\mathrm{imu}}_{0:t} \cup 
\mathcal{Z}^{\mathrm{lidar}}_{0:t}$ denote the set of all measurements up to time $t$.
The SLAM problem seeks the maximum a posteriori (MAP) estimate of the 
trajectory $\mathcal{X}_{0:t}$ given all measurements $\mathcal{Z}_{0:t}$:
\begin{align}\label{eq:map}
\mathcal{X}_{0:t}^\star
    &= \arg\max_{\mathcal{X}_{0:t}}
        p(\mathcal{X}_{0:t} \mid \mathcal{Z}_{0:t})\nonumber \\
    &= \arg\max_{\mathcal{X}_{0:t}}
        p(\mathcal{Z}_{0:t} \mid \mathcal{X}_{0:t}) \, p(\mathcal{X}_{0:t}).
\end{align}

Here, $\mathcal{Z}^{\mathrm{imu}}_{0:t} = \{ \mathbf{z}_i^{\mathrm{imu}}\}_{i=0}^{t}$ contains all accumulated IMU preintegration measurements, and 
$\mathcal{Z}^{\mathrm{lidar}}_{0:t} = 
\{\, \mathbf{z}_i^{\mathrm{scan}},\, \mathbf{z}_i^{\mathrm{loop}} \,\}_{i=0}^{t}$ 
contains the LiDAR scan-matching measurements and LiDAR loop-closure measurements.
Thus, we write the factor-graph SLAM posterior as
\begin{equation}
p(\mathcal{X}_{0:t} \mid \mathcal{Z}_{0:t})
\;\propto\;
\prod_k
\exp\!\Big(
-\tfrac{1}{2}
\bigl\| r_k(\mathcal{X}_{0:t}) \bigr\|_{\Lambda_k}^2
\Big),
\end{equation}
where $r_k(\mathcal{X}_{0:t}) = h_k(\mathcal{X}_{0:t}) - \mathbf{z}_k$ is the residual 
associated with measurement $\mathbf{z}_k \in \mathcal{Z}_{0:t}$, and 
$\Lambda_k = R_k^{-1}$ is the corresponding information matrix of the measurement noise.
The maximum-a-posteriori (MAP) estimate $\mathcal{X}_{0:t}^\star$ is obtained by solving the nonlinear least-squares problem
\begin{equation}
\label{eq:slam_ls_simple}
\mathcal{X}_{0:t}^\star
=
\arg\min_{\mathcal{X}_{0:t}}
\sum_{k}
\bigl\|
r_k(\mathcal{X}_{0:t})
\bigr\|_{\Lambda_k}^2.
\end{equation}

While the VRVM framework is agnostic to the underlying SLAM implementation, we adopt LIO-SAM~\citep{shan2020liosam}, a tightly coupled LiDAR–IMU odometry and mapping system that formulates lidar–inertial fusion as inference on a factor graph and solves it incrementally using the iSAM2~\citep{kaess2012isam2} algorithm. 
LIO-SAM maintains a 6-DoF pose trajectory in $SE(3)$ with factors constructed from IMU preintegration, LiDAR scan matching, and loop-closure constraints. 
The incremental solver maintains a sparse factorisation of the linearised normal equations and updates only the affected variables when new LiDAR or loop-closure factors arrive, enabling real-time operation.
Although the SLAM back-end operates in full 3D, we project each 6-DoF LIO-SAM estimate onto a planar state to support water-surface exploration and extract 
the corresponding $3\times 3$ marginal covariance of $(x_i, y_i, \psi_i)$.

We now describe how the SLAM results are used to compute the uncertainty 
criteria. Linearizing all measurement factors in the factor graph around the 
current estimate $\mathcal{X}_{0:t}^\star$ leads to the standard Gaussian 
approximation of the posterior,
\begin{equation}
p(\mathcal{X}_{0:t} \mid \mathcal{Z}_{0:t})
\;\approx\;
\mathcal{N}\!\left(\mathcal{X}_{0:t}^\star,\, 
\mathbf{\Sigma}_{\mathcal{X}}\right),
\end{equation}
where $\mathbf{\Sigma}_{\mathcal{X}}$ denotes joint covariance matrix of the USV state variables.

Instead of forming $\mathbf{\Sigma}_{\mathcal{X}}$ explicitly, we follow standard practice in iSAM2 and query the solver for the marginal covariance blocks corresponding to the individual optimised states. 
For each optimised state $\mathbf{x}_i$, we obtain the $3\times 3$ marginal covariance
$\mathbf{\Sigma}_{\mathbf{x}_i}$.
For virtual map updates, we require only the positional uncertainty, so we 
extract the $2\times 2$ positional block
$\mathbf{\Sigma}_{\mathbf{p}_i}$
from $\mathbf{\Sigma}_{\mathbf{x}_i}$ and use it in the inverse sensor model to update the virtual map.
The pair $(\mathbf{x}_i,\, \mathbf{\Sigma}_{\mathbf{p}_i})$ therefore defines a 
Gaussian belief over the USV state at time step $i$. 
This belief is used as the input to the virtual map representation and to the utility evaluation during exploration.

\subsection{Environment and Map representation}
\label{sec:virtual_map}
Sensor noise and environmental factors such as wind and hydrodynamic disturbance introduce significant uncertainty into the mapping process. 
To account for this, we represent the environment using two data structures: an occupancy grid map $\mathcal{M}$ that captures the observed geometry, and a \emph{virtual map} $\mathcal{V}$ that quantifies the reliability (map accuracy) of the mapped area. 
We then use this reliability measure as a planning criterion.
The virtual map, first proposed by~\citet{wang2019autonomous}, is a grid-based representation that quantifies map accuracy by propagating the covariance of robot states and observed landmarks into the workspace through the inverse sensor model.
In this paper, we employ a landmark-free adaptation of the virtual map~\citep{wang2019virtualmap} 
to ensure compatibility with our SLAM pipeline.

The workspace $\mathcal{W}$ is represented by a virtual map 
$\mathcal{V} = \{\mathbf{v}_k\}$, where each grid cell $\mathbf{v}_k$ is a \emph{virtual landmark} whose position is modelled as a 2D Gaussian. 
We distinguish between \emph{actual landmarks} $\mathbf{l}_k$, corresponding to confidently occupied cells, and \emph{virtual landmarks} $\mathbf{v}_k$, 
which model the uncertainty of all remaining cells (free, uncertain, or unknown).

Let $m_k \in \mathcal{M}$ denote the occupancy state of the $k$-th grid cell, with $P(m_k = 1)$ the probability that it is occupied. We define an indicator for whether a cell hosts a virtual landmark as
\begin{equation}
P(\mathbf{v}_k = 1)
=
\begin{cases}
0, & P(m_k = 1) \ge \theta_{\mathrm{occ}}, \\[0.3ex]
1, & P(m_k = 1) < \theta_{\mathrm{occ}} \text{ or cell unknown},
\end{cases}
\label{eq:virtual_indicator}
\end{equation}
where $\theta_{\mathrm{occ}} \in (0,1)$ is an occupancy threshold.
For brevity, we refer to this quantity as the virtual-map indicator $P_{\mathrm{v}}(\mathbf{v}_k)$.
In this way, geometric structure that is confidently observed (high-occupancy cells) is handled by the SLAM and occupancy map as actual landmarks, while the virtual map concentrates on representing mapping uncertainty in partially observed regions. 
For a unified and simplified map representation, we treat each actual landmark $\mathbf{l}_k$ as a virtual landmark $\mathbf{v}_k$, whose associated 2D Gaussian remains unupdated.

\subsection{Frontier Selection via Expectation–Maximisation}\label{sec:frontier&em}
During exploration, once the USV reaches its current goal at time step $t$, 
the expectation–maximisation (EM) planner is invoked to select the next optimal control sequence $\pi^{\star}$. Candidate control sequences are generated from a set of frontiers. We use two types of frontiers: \emph{exploring frontiers}, which lie on the boundary between observed and unknown regions, and \emph{exploit frontiers}, which guide revisits to previously mapped structures. 
For each frontier $\mathbf{a}_i$, a candidate control sequence $\pi_i$ is 
produced using the occupancy map to ensure safe navigation.

For a given candidate $\pi_i$, let $\mathcal{X}_{t:t+h}$ denote the hypothetical states generated by applying $\pi_i$ over a planning horizon of length $h$.
The resulting end-of-horizon trajectory is defined as
\begin{equation}
\mathcal{X}_{0:t+h} = \mathcal{X}_{0:t} \cup \mathcal{X}_{t:t+h},
\end{equation}

We evaluate the quality of a hypothetical trajectory by maximising the expected complete-data log-likelihood of the measurements and the latent virtual-map variables. For a candidate trajectory $\mathcal{X}_{0:t+h}$, the EM objective is
\begin{align}
\mathcal{X}_{0:t+h}^\star
  = \arg\max_{\mathcal{X}_{0:t+h}}
     \, \mathbb{E}_{\mathcal{V}_t \mid \mathcal{Z}_{0:t}}
        \left[ \log p(\mathcal{Z}_{0:t}, \mathcal{V}_t \mid \mathcal{X}_{0:t+h}) \right],
\end{align}
where $\mathcal{V}_t$ denotes the virtual map inferred from all past measurements $\mathcal{Z}_{0:t}$.
The EM procedure consists of two steps. The \emph{E-step} computes the expected complete-data log-likelihood:
\begin{equation}
Q(\mathcal{X}_{0:t+h} \mid \mathcal{X}_{0:t})
= \mathbb{E}_{\mathcal{V}_t \mid \mathcal{Z}_{0:t}, \mathcal{X}_{0:t}}
  \left[ \log p(\mathcal{Z}_{0:t}, \mathcal{V}_t \mid \mathcal{X}_{0:t+h}) \right].
\end{equation}
The \emph{M-step} updates the hypothetical trajectory by maximising this expectation:
\begin{equation}
\mathcal{X}_{0:t+h}^\star
= \arg\max_{\mathcal{X}_{0:t+h}}
  Q(\mathcal{X}_{0:t+h} \mid \mathcal{X}_{0:t}).
\end{equation}

To avoid the exponential growth in the number of possible virtual-map states, we replace the \emph{E-step} with a classification step (\emph{C-step}) that directly computes a deterministic estimate of the virtual map:
\begin{align}
\mathcal{V}_t
&\approx \arg\max_{\mathcal{V}} p(\mathcal{V} \mid \mathcal{Z}_{0:t}, \mathcal{X}_{0:t}), \nonumber \\
&= g_{\mathrm{lidar}}(\mathcal{Z}_{0:t}, \mathcal{X}_{0:t}^{\star}).
\label{eqn:e-step}
\end{align}
The computation of the virtual map is described in Sec.~\ref{sec:vrvm}.

Subsequently, in the \emph{M-step}, we update the hypothetical trajectory by
maximising the complete-data log-likelihood under the virtual map obtained in
the C-step:
\begin{equation}
\mathcal{X}_{0:t+h}^\star
= \arg\max_{\mathcal{X}_{0:t+h}}
    \log p(\mathcal{Z}_{0:t}, \mathcal{V}_t \mid \mathcal{X}_{0:t+h}).
\end{equation}

Given this deterministic map, the \emph{M-step} performs a 
maximum a posteriori (MAP) update of the hypothetical trajectory:
\begin{align}
\label{eqn:mstep-map}
\mathcal{X}_{0:t+h}^\star
= \arg\max_{\mathcal{X}_{0:t+h}}
\left[
    \log p(\mathcal{Z}_{0:t}, \mathcal{V}_t , \mathcal{X}_{0:t+h})
\right],  \\
p(\mathcal{Z}_{0:t}, \mathcal{V}_t , \mathcal{X}_{0:t+h})
= p(\mathcal{Z}_{0:t}, \mathcal{V}_t \mid \mathcal{X}_{0:t+h})\,
  p(\mathcal{X}_{0:t+h}).
\end{align}
With the optimised trajectory $\mathcal{X}_{0:t+h}^\star$, we update the 
virtual map $\mathcal{V}_{t+h}$ using the inverse LiDAR sensor model 
$g_{\mathrm{lidar}}(\cdot)$ and compute the corresponding mapping utility based on a per-landmark evaluation criterion $\phi(\cdot)$:
\begin{align}
\label{eq:utility_map}
    U_{\mathrm{map}}(\pi) 
    &= \phi(\mathcal{V}_{t+h}), \nonumber\\
    &= \sum_{\mathbf{v}_i \in \mathcal{V}_{t+h}} \phi(\mathbf{v}_i).
\end{align}
This mapping utility $U_{\mathrm{map}}(\pi)$ forms one component of the overall 
planning objective. The complete utility function consists of three terms: the 
mapping utility $U_{\mathrm{map}}(\pi)$, the trajectory utility 
$U_{\mathrm{traj}}(\pi)$, and the energy utility $U_{\mathrm{length}}(\pi)$:
\begin{align}
    \pi^\star 
    &= \arg\max_{\pi \in \Pi} U(\pi),  \\
    U(\pi) &=  \big( U_{\mathrm{map}}(\pi) 
          + U_{\mathrm{traj}}(\pi)
          + U_{\mathrm{length}}(\pi) \big).
          \label{eq:utility}
\end{align}
Further details on the utility computation are provided in 
Sec.~\ref{sec:vrvm}.

\section{Variable-Resolution Virtual Map} 
\label{sec:vrvm}

Variable-resolution virtual map
(VRVM) runs in a receding-horizon~\citep{brugali2025mobile} loop that couples a variable-resolution virtual map with visibility-limited prediction and an EM planner. 
During each control cycle, the quadtree map is adaptively refined based on depth-aware, uncertainty-driven rules, where only visible leaves are updated using the inverse sensor model.
The uncertainty of reliably occupied leaves containing obstacles is held fixed to prevent redundant updates.
This quadtree map serves as a reference for ranking kinodynamically feasible trajectories using a unified utility, $U(\pi)$, that combines the end-of-horizon SLAM optimised pose log-determinant ($U_{\mathrm{traj}}(\pi)$), the area-weighted map log-determinant over the visible set ($U_{\mathrm{map}}(\pi)$), and a path-length penalty ($U_{\mathrm{length}}(\pi)$).
The resulting formulation maintains computational complexity approximately linear in the visible-set size and remains compatible with incremental smoothing factor graph SLAM.

In this section, we introduce the proposed VRVM algorithm. We first describe the virtual-landmark representation and its
initialisation and update rules, which apply to both the classical uniform
virtual map (UVM) and the VRVM. We then present the variable-resolution
formulation and its implementation.

\subsection{Virtual Landmark}\label{sec:virtual_landmark}
As discussed in Sec.~\ref{sec:virtual_map}, the position of each virtual landmark is modelled as a 2D Gaussian,
\begin{equation}
\mathbf{v}_q \sim \mathcal{N}(\boldsymbol{\mu}_q, \boldsymbol{\Sigma}_q),
\qquad
\boldsymbol{\mu}_q\in \mathbb{R}^2,\;
\boldsymbol{\Sigma}_q \in \mathbb{R}^{2\times 2}.
\end{equation}
At the beginning of exploration, all virtual landmarks are initialised with an isotropic Gaussian prior
\begin{equation}
\boldsymbol{\Sigma}_{q} = \boldsymbol{\Sigma}_{0} = \begin{bmatrix}
\sigma_0^2 & 0 \\
0 & \sigma_0^2,
\end{bmatrix},
\end{equation}
which expresses high and direction-independent positional uncertainty.
The value of $\sigma_0$ is selected to be sufficiently large relative to the covariance produced by a typical single range measurement under standard pose uncertainty. This follows the guideline that the prior should remain less informative than any practicable measurement update~\citep{wang2022virtual}.

To keep the computation local, the virtual map is updated only along the USV trajectory using the inverse sensor model $g_{\mathrm{lidar}}(\cdot)$. 
Given a trajectory $\mathcal{X}_{0:t}$, let $\mathbf{p}_k \in \mathbb{R}^2$ denote the position of state $\mathbf{x}_k \in \mathcal{X}_{0:t}$. 
For a LiDAR sensing range $R$, the set of virtual landmarks visible from $\mathbf{x}_k$ is defined as
\begin{equation}
\label{eq:visible_set}
\mathcal{V}(\mathbf{x}_k)
\triangleq
\Bigl\{
\mathbf{v}_q \;\big|\;
\|\boldsymbol{\mu}_q - \mathbf{p}_k\|_2
\le R + \sqrt{2}\,h_q
\Bigr\},
\end{equation}
where $h_q$ is the half-side length of virtual landmark $\mathbf{v}_q$. The additional term
$\sqrt{2}\,h_q$ is the radius of the cell’s circumscribed circle and ensures
that any cell whose area intersects the sensing disc of radius $R$ is included in the visible set.
Similarly, the set of virtual landmarks observed along the trajectory $\mathcal{X}_{0:t}$ is defined as
\begin{equation}
\mathcal{V}(\mathcal{X}_{0:t}) 
\;\triangleq\; 
\bigcup_{\mathbf{x}_k \in \mathcal{X}_{0:t}} \mathcal{V}(\mathbf{x}_k)
\end{equation}

We now describe how each virtual landmark cell $\mathbf{v}_q \in \mathcal{V}(\mathcal{X}_{0:t})$
is updated. 
Let $\mathbf{x}_k^\star \in \mathcal{X}^\star_{0:t}$ denote the optimised
USV state at time $k$, and let
$\bar{\mathbf{z}}_k^{\mathrm{lidar}}
= \mathbf{z}_k^{\mathrm{lidar}} - \mathbf{n}_k^{\mathrm{lidar}}$
be the noise-free LiDAR measurement.
Using the inverse sensing model introduced previously, we linearise
$g_{\mathrm{lidar}}(\mathbf{x}, \mathbf{z})$ around the operating point
$(\bar{\mathbf{x}}, \bar{\mathbf{z}}) = (\mathbf{x}_k^\star, \bar{\mathbf{z}}_k^{\mathrm{lidar}})$:
\[
\mathbf{y} \approx
g_{\mathrm{lidar}}(\bar{\mathbf{x}}, \bar{\mathbf{z}})
+ \mathbf{A}\,(\mathbf{x} - \bar{\mathbf{x}})
+ \mathbf{B}\,(\mathbf{z} - \bar{\mathbf{z}}),
\]
where the Jacobians are
\begin{equation}
\mathbf{A}
\triangleq
\frac{\partial g_{\mathrm{lidar}}}{\partial \mathbf{x}}
\bigg|_{(\bar{\mathbf{x}}, \bar{\mathbf{z}})},
\qquad
\mathbf{B}
\triangleq
\frac{\partial g_{\mathrm{lidar}}}{\partial \mathbf{z}}
\bigg|_{(\bar{\mathbf{x}}, \bar{\mathbf{z}})}.
\end{equation}

The positional and measurement noise from time step $k$ are propagated to the virtual landmark as a
covariance matrix $\boldsymbol{\Sigma}_y$:
\begin{equation}
\label{eq:sigma_v}
\boldsymbol{\Sigma}_y
=
\mathbf{A}\,\boldsymbol{\Sigma}_{\mathbf{p}_k}\,\mathbf{A}^\top
+
\mathbf{B}\,\mathbf{R}_k^{\mathrm{lidar}}\,\mathbf{B}^\top,
\end{equation}
with corresponding information matrix
$\boldsymbol{\Omega}_y = \boldsymbol{\Sigma}_y^{-1}$.

A virtual landmark $\mathbf{v}_q$ is typically observed at multiple time steps, and its covariance $\boldsymbol{\Sigma}_q$ must be updated accordingly. For
computational efficiency, we adopt the incremental information–filter update used in~\citep{wang2019autonomous}. For each visible virtual landmark
$\mathbf{v}_q$, we maintain its belief in information form,
$\boldsymbol{\Omega}_q = \boldsymbol{\Sigma}_q^{-1}$. Given the composite
covariance $\boldsymbol{\Sigma}_y$ in Eq.~\ref{eq:sigma_v}, we update all
virtual landmarks $\mathbf{v}_q \in \mathcal{V}(\mathbf{x}_k)$ using the additive information rule:
\begin{equation}
\label{eq:info_update}
\boldsymbol{\Omega}_q \leftarrow \boldsymbol{\Omega}_q + \boldsymbol{\Omega}_y,
\qquad
\boldsymbol{\Sigma}_q \leftarrow \boldsymbol{\Omega}_q^{-1},
\quad
\forall\, \mathbf{v}_q \in \mathcal{V}(\mathbf{x}_k).
\end{equation}
Because $\boldsymbol{\Sigma}_y \succ 0$, this update is monotone in the Loewner
order~\citep{horn2013matrix}, ensuring that $\log\det\boldsymbol{\Sigma}_q$ is
non-increasing with each effective observation. The log-determinant of the
virtual-landmark covariance provides a consistent D-optimality measure
of uncertainty reduction.
Thus, we define the mapping uncertainty criterion for a virtual landmark $\mathbf{v}_q$ as
\begin{equation}
    \phi(\mathbf{v}_q) \equiv -\log\!\det(\bm{\Sigma}_q).
\end{equation}

The only quantity passed from the SLAM factor graph to the virtual map is the position $\mathbf{p}_k$ and the  covariance $\boldsymbol{\Sigma}_{\mathbf{p}_k}$ that appears in
Eq.~\ref{eq:sigma_v}. 
When loop-closure factors are added to the SLAM factor
graph, the resulting reduction in marginal position covariance is directly reflected in the propagated virtual-landmark covariance $\boldsymbol{\Sigma}_y$. 
As illustrated in Fig.~\ref{fig:pipeline}, virtual
landmarks are stored and updated exclusively within the exploration module and are not introduced as factors in the SLAM graph. 
This separation ensures that coverage and localisation are coupled through a unified information-theoretic
objective, while avoiding any feedback of virtual-map variables into the SLAM back-end.

\subsection{Uniform Virtual Map}
\label{sec:uvm}

\begin{figure}[!t]
    \centerline{\includegraphics[width=\linewidth]{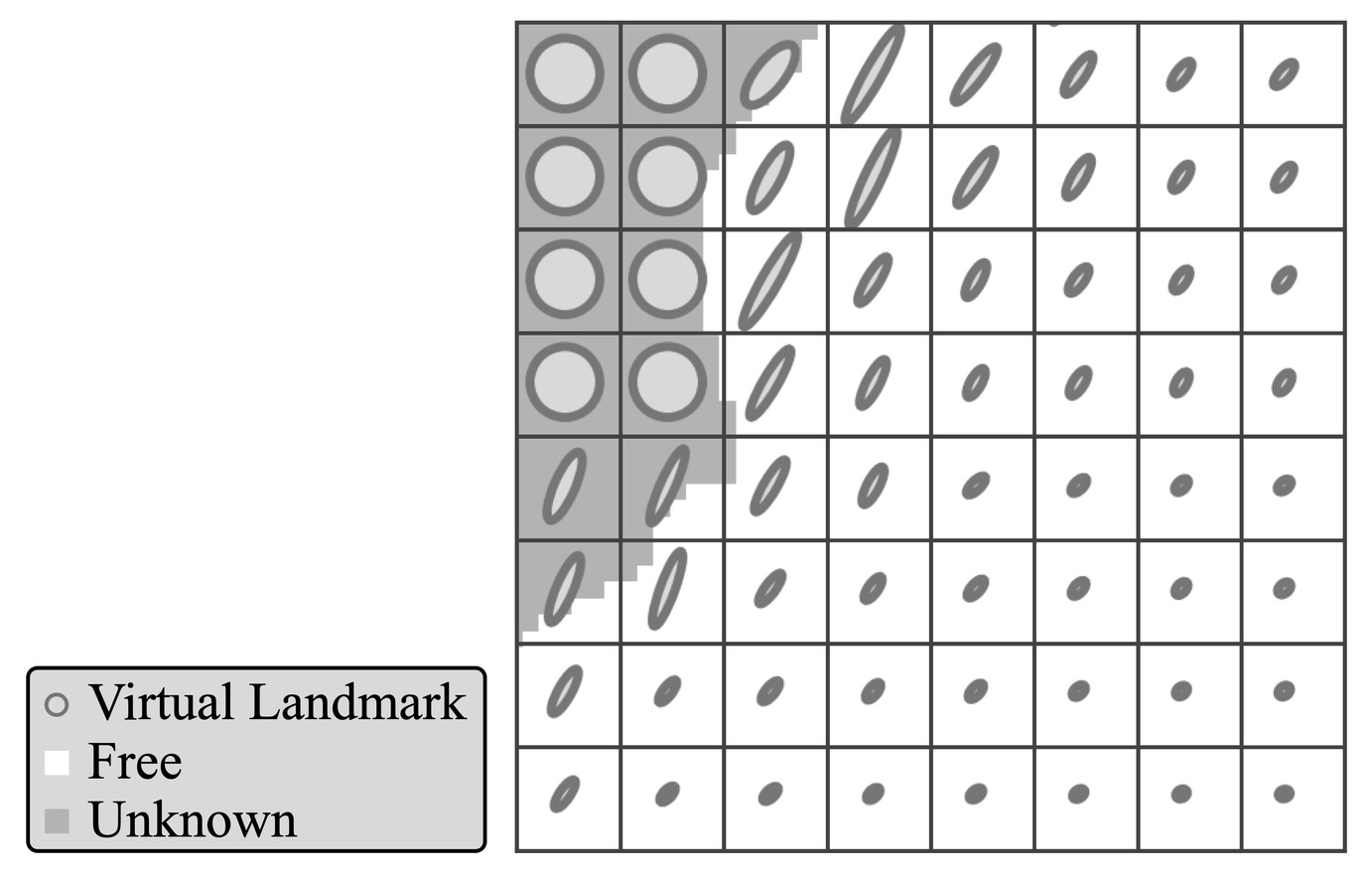}}
    \caption{Visualization of a portion of the classical uniform virtual map (UVM).}
    \label{fig:uvm} 
\end{figure}

As shown in Fig.~\ref{fig:uvm}, the original virtual map~\citep{wang2022virtual} adopts a uniform grid structure.
At the beginning of exploration, the workspace $\mathcal{W}$ is evenly partitioned into a fixed number of $K$ cells. The resulting uniform virtual map is expressed as
\begin{equation}
    \mathcal{V} = \{\mathbf{v}_k \mid k = 0, ..., K-1\}
\end{equation}
where each $\mathbf{v}_k$ denotes a virtual cell anchored at a fixed spatial location. The virtual cell resolution may be identical to, or different from, the resolution of the occupancy grid used for mapping. Because the grid is fixed for the entire mission, the number and placement of virtual cells do not adapt to changes in viewpoint or scene complexity.
Fig. ~\ref{fig:uvm} illustrates a portion of the boundary between explored and unexplored regions under this uniform structure.

\begin{figure}[!t]
    \centerline{\includegraphics[width=\linewidth]{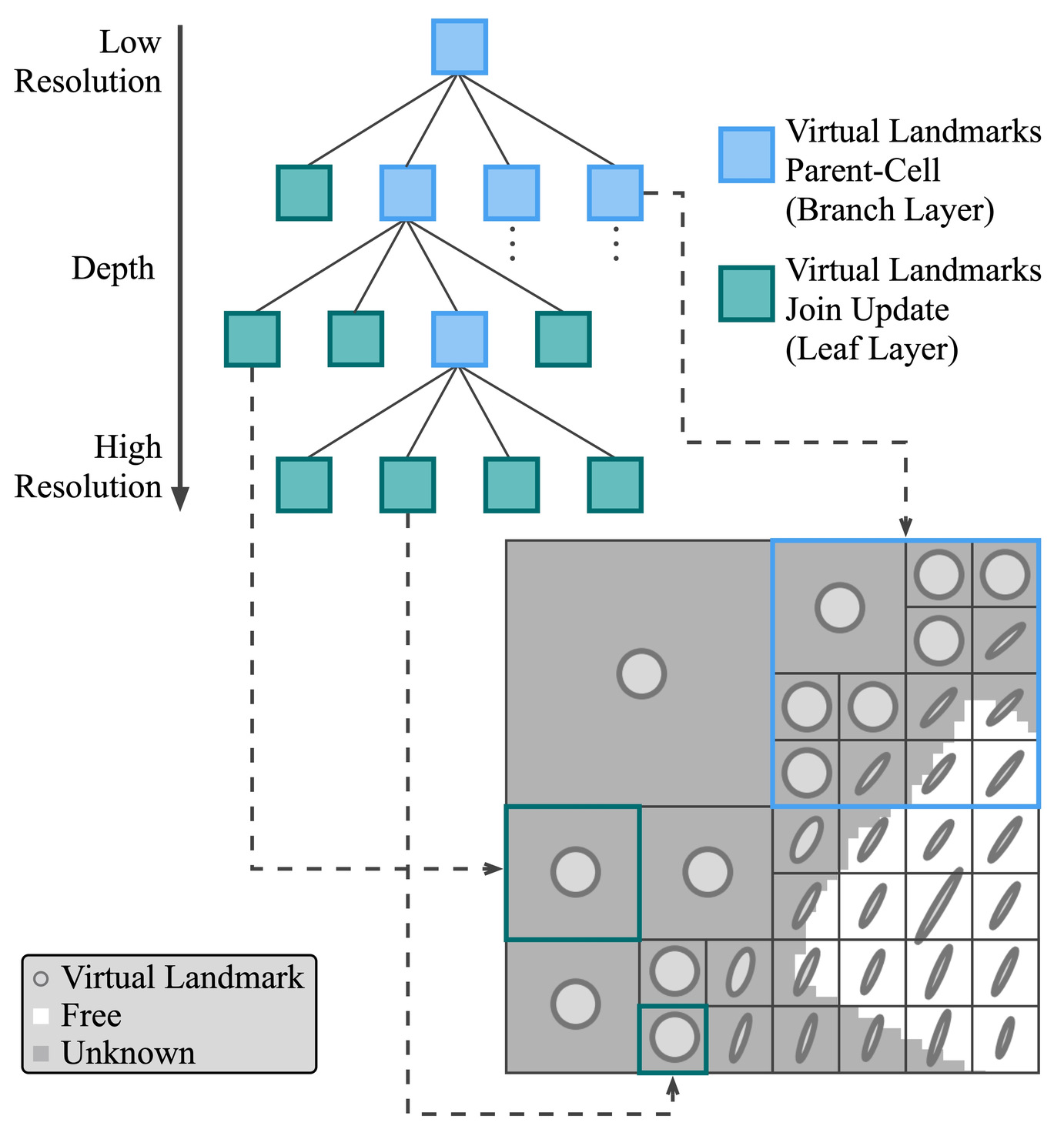}}
    \caption{Visualization of a portion of the variable-resolution virtual map hierarchy over a quadtree. A correspondence between the leaf and branch cells in the quadtree and their physical surface regions is shown.}
    \label{fig:quadtree} 
\end{figure}

\subsection{Variable-Resolution Quadtree Refinement and Occupancy Locking}
\label{sec:quadtree_lock}
Our variable-resolution virtual map is implemented using a quadtree rather than a uniform grid. 
In this quadtree $\mathcal{V}$ (Fig.~\ref{fig:quadtree}), each leaf $\mathbf{v}_q$ represents a virtual landmark and carries an area indicator specifying the spatial coverage of the leaf. 
During exploration, the quadtree is refined adaptively to ensure that the virtual map provides the resolution needed for planning. 

At the start of exploration, the virtual map consists of a quadtree with a single leaf. 
As the robot begins to observe and explore the environment, the
tree is recursively refined. 
As shown in Fig.~\ref{fig:vrvm}, refinement is applied only where it benefits planning: regions with high posterior uncertainty are maintained at high resolution, whereas structure-free regions remain coarse.

A leaf $\mathbf{v}_q$ is split only when it is observed and its local
uncertainty $\boldsymbol{\Sigma}_q$ remains high at its current resolution, or
when its occupancy state is ambiguous.
Formally,
\begin{equation}
\label{eq:split_rule}
\det(\boldsymbol{\Sigma}_q) \;\ge\; \tau_{\det}(d_q)
\quad \text{or} \quad
\bigl|\,P_{\mathrm{v}}(\mathbf{v}_q) - 0.5\,\bigr| \;\le\; \tau_p ,
\end{equation}
where $d_q$ denotes the depth of leaf $q$. We employ a depth-aware uncertainty
threshold of the form $\tau_{\det}(d_q) = \tau_{\det}^{0}\,4^{-d_q}$, where
$\tau_{\det}^{0}$ is the base threshold. The term $P_{\mathrm{v}}(\mathbf{v}_q)\in[0,1]$
denotes the occupancy probability at the leaf centre, as defined in
Eq.~\ref{eq:virtual_indicator}.

A refinement (leaf-splitting) termination criterion is also enforced. A leaf
$\mathbf{v}_q$ may be split only if both
\begin{equation}
    h_q > \frac{r_0}{2}
\qquad\text{and}\qquad
d_q < d_{\max},
\end{equation}
that is, splitting is permitted only when the half leaf size $h_q$ is larger than the minimum
admissible resolution $ \frac{r_0}{2}$ and the quadtree depth limit $d_{\max}$ has not been reached.
Equivalently, refinement is terminated once either
\begin{equation}
h_q \le \frac{r_0}{2}
\qquad\text{or}\qquad
d_q \ge d_{\max}.
\end{equation}
The parameters $\tau_{\det}^{0}$, $\tau_p$, and $d_{\max}$ are predefined
constants. When a split occurs, four children are created at the centres of the
subcells and are initialised according to
\begin{equation}
\boldsymbol{\Sigma}_{\mathrm{child}} = \boldsymbol{\Sigma}_{0},
\qquad
P_{\mathrm{v}}(\mathbf{v}_{\mathrm{child}}) = P_{\mathrm{v}}(\mathbf{v}_{q}) .
\end{equation}
The full refinement procedure is summarised in
Alg.~\ref{alg:refine}.

Leaves whose centres fall in confidently occupied cells are locked by setting 
\begin{equation}
\boldsymbol{\Sigma}_q = 
\boldsymbol{\Sigma}_{\mathrm{fix}} = 
\begin{bmatrix}
\sigma_\mathrm{fix}^2 & 0 \\
0 & \sigma_\mathrm{fix}^2,
\end{bmatrix},
\end{equation}
 and excluding them from subsequent updates and from the map uncertainty term. Locking is applied after visible-set updates in each cycle, as shown in Alg.~\ref{alg:lock}.
Let $N_{\mathrm{split}}$ denote the number of splits performed in a cycle.
The computational complexity of one refinement cycle is
\begin{equation}
    O_{\mathrm{refine}} = O(N_{\mathrm{split}}) + O\!\left(|\mathcal{V}(\mathbf{x}_k)|\right),
\end{equation}
and is typically dominated by the size of the visible set.

\begin{algorithm2e}[t]
  \caption{Adaptive Quadtree Refinement (VRVM)}
  \label{alg:refine}
  \Inputs{Root $\mathbf{v}_{q_{\mathrm{root}}}$; size floor $r_0$; depth cap $d_{\max}$; thresholds $\tau_{\det}^{0}; \tau_p$; init variance $\bm{\Sigma}_{0}$.}
  \Output{Updated leaf set $\mathcal{V}$.}
  \SetKwFunction{Refine}{Refine}
  \SetKwFunction{Split}{Split}
  \SetKwProg{Fn}{Function}{:}{end}
  \Fn{\Refine{$\mathbf{v}_q$}}{
    \uIf{$\mathbf{v}_q$ is not a leaf}{
      \ForEach{child $\mathbf{v}_c$ of $\mathbf{v}_q$}{\Refine{$\mathbf{v}_c$}}
    }\ElseIf{$\bigl(\det(\bm{\Sigma}_q)\ge\tau_{\det}^{0}4^{-d_q} \ \lor \ |P_{\mathrm{v}}(\mathbf{v}_q)-0.5|\le\tau_p\bigr) \ \land \ h_q>r_0/2 \ \land \ d_q<d_{\max}$}{
        \Split{$\mathbf{v}_q$} \tcp*{into four children}
        \For{$i\leftarrow 1$ \KwTo $4$}{
          $\mathbf{v}_c \leftarrow \text{child}_i(\mathbf{v}_{q})$;\ 
          $P_{\mathrm{v}}(\mathbf{v}_c)\leftarrow P_{\mathrm{v}}(\mathbf{v}_q)$ \ 
          $\bm{\mu}_{c} \leftarrow \text{subcell centre}$;\
          $\bm{\Sigma}_{c} \leftarrow \bm{\Sigma}_{0}$ \ 
          \Refine{$\mathbf{v}_c$}
          }
      }
    }
  \BlankLine
  \Refine{$\mathbf{v}_{q_{\mathrm{root}}}$}
\end{algorithm2e}

\begin{algorithm2e}[t]
  \caption{Projective Occupancy Locking}
  \label{alg:lock}
  \Inputs{Projected occupancy grid $\mathcal{M}$; threshold $\theta_{\mathrm{occ}}$; fixed variance $\bm{\Sigma}_{\mathrm{fix}}$.}
  \Output{Locked leaves removed from future updates}.
  \SetKwFunction{WorldToGrid}{WorldToGrid}
  \ForEach{finest-resolution leaf $\mathbf{v}_q$, $\mathbf{v}_q \in \mathcal{V}$}{
    $(i,j) \leftarrow \WorldToGrid\big(\bm{\mu}_q, \mathcal{M}\big)$ %
    \If{$\mathcal{M}[i,j] \ge \theta_{\mathrm{occ}}$}{
      $\bm{\Sigma}_q \leftarrow \bm{\Sigma}_{\mathrm{fix}}$ \tcp* {mark $q$ as fixed}
    }
  }  
\end{algorithm2e}

\subsection{Area-Weighted Map Valuation}
\label{sec:area_weight}
As shown in Eq.~\ref{eq:utility_map}, the mapping accuracy is quantified by the mapping utility $U_{\mathrm{map}}(\pi)$, where the uncertainty evaluation function $\phi(\mathbf{v}_q)$ is defined as a log-determinant metric.
For the uniform virtual map, the virtual landmarks are evenly distributed, and each one contributes with equal weight. 
In contrast, for the variable-resolution representation, directly summing $\log\det(\boldsymbol{\Sigma}_q)$ over leaves is split-sensitive: dividing a coarse leaf into several finer leaves with similar covariances increases the total simply because the number of terms increases, which can bias decisions at earlier stages. We therefore seek a valuation that is less dependent on the current split structure while still promoting viewpoints that reduce uncertainty in high-entropy regions.

For a set of virtual landmarks $\mathcal{V}$, we define the split-invariant weight of a leaf $\mathbf{v}_q \in \mathcal{V}$ as
\begin{equation}
w_{\mathrm{area}}(\mathbf{v}_q, \mathcal{V}) \;=\;
\frac{A_q}{\sum_{\mathbf{v}_r \in \mathcal{V}} A_r},
\end{equation}
where $A_q = 4h_q^2$ denotes the area associated with leaf $\mathbf{v}_q$. 
By construction, these weights are normalised such that 
\begin{equation}
    \sum_{\mathbf{v}_r \in \mathcal{V}} w_{\mathrm{area}}(\mathbf{v}_r, \mathcal{V}) = 1.
\end{equation} 
The area-weighted map uncertainty of the virtual map is then given by
\begin{equation}
\label{eq:J_area_x_fixed}
J_{\mathrm{area}}(\mathcal{V})
\;=\;\sum_{\mathbf{v}_q \in \mathcal{V}}
w_{\mathrm{area}}(\mathbf{v}_q, \mathcal{V})\,
\phi(\mathbf{v}_q).
\end{equation}
$\phi(\mathbf{v}_q)$ is the mapping uncertainty criterion defined in Sec.~\ref{sec:virtual_landmark}.
If a parent leaf is split into children whose covariances are locally similar to the parent covariance, the weighted sum in
Eq.~\ref{eq:J_area_x_fixed} remains approximately unchanged, so
$J_{\mathrm{area}}(\mathcal{X}_{0:t})$ is locally invariant to the current quadtree split pattern.

For a candidate control sequence
$\pi$ with an end-of-horizon trajectory $\mathcal{X}_{0:t+h}$, we propagate our variable-resolution virtual map along the
discretised trajectory using the information-form update in
Sec.~\ref{sec:virtual_landmark}. The mapping utility $U_{\mathrm{map}}(\pi)$ is then formulated using area-weighted map valuation criteria.

The mapping utility $U_{\mathrm{map}}(\pi)$ consists of two components: the end-of-horizon virtual map uncertainty $J_{\mathrm{area}}(\mathcal{V}_{t+h})$, and the mapping accuracy gain achieved by executing the control policy $\Delta J_{\mathrm{gain}}(\pi)$:
\begin{equation}
U_{\mathrm{map}}(\pi) = J_{\mathrm{area}}(\mathcal{V}_{t+h}) + \Delta J_{\mathrm{gain}}(\pi).
\end{equation}
To emphasise uncertainty reduction along the executed trajectory, we define the split-invariant gain $\Delta J_{\mathrm{gain}}(\pi)$ as:
\begin{equation}
\Delta J_{\mathrm{gain}}(\pi)
= \sum_{k=0}^{t+h} \left[\gamma_k \, J_{\mathrm{area}}\!\big(\mathcal{V}_{t+h}(\mathbf{x}_k)\big)
\;-\;  J_{\mathrm{area}}\!\big(\mathcal{V}_t(\mathbf{x}_k)\big)\right].
\end{equation}
where $\mathcal{V}_{t}(\mathbf{x}_k)$ and $\mathcal{V}_{t+h}(\mathbf{x}_k)$ denote the sets of virtual landmarks visible from $\mathbf{x}_k$ in the virtual maps constructed before and after executing the control sequence $\pi$, respectively.
$\gamma_k \in (0,1]$ is a discount factor to downweigh samples
that require longer travel distance or have higher predicted pose
uncertainty.
We summarise our calculation of $\Delta J_{\mathrm{gain}}(\pi)$ in  Alg.~\ref{alg:areagain}.
In our implementation, we simplify the formulation by using a fixed $\gamma_k$.
The virtual map itself is still updated according to
Sec.~\ref{sec:vrvm}; this area-weighted criteria
is used only to rank candidate control sequences during utility calculation.

\begin{algorithm2e}[t]
\caption{Split-Invariant Mapping Accuracy Gain}
\label{alg:areagain}
\KwIn{Initial trajectory guess $\bar{\mathcal{X}}_{0:t+h}$; optimised trajectory $\mathcal{X}^{\star}_{0:t+h}$; virtual maps before and after executing policy $\pi$, $\mathcal{V}_{t}$ and $\mathcal{V}_{t+h}$; weights $\{\gamma_k\}_{k=0}^{t+h}$.}
\KwOut{Split-invariant gain $\Delta J_{\mathrm{gain}}(\pi)$.}

$\Delta J_{\mathrm{gain}} \leftarrow 0$\;

\For{$k \leftarrow 0$ \KwTo $t+h$}{
  $\mathcal{V}_{t}(\mathbf{x}_k) \leftarrow \mathrm{Visible}(\mathcal{V}_{t}, \mathbf{x}_k)$,\;
  $\mathcal{V}_{t+h}(\mathbf{x}_k) \leftarrow \mathrm{Visible}(\mathcal{V}_{t+h}, \mathbf{x}_k)$\;
  \tcp* {Extract visible sets}

  $S_t \leftarrow \sum_{\mathbf{v}_q \in \mathcal{V}_{t}(\mathbf{x}_k)} A_q$,\;
  $S_{t+h} \leftarrow \sum_{\mathbf{v}_q \in \mathcal{V}_{t+h}(\mathbf{x}_k)} A_q$\;
  \tcp* {Area-normalisation terms}

  $J_t(\mathbf{x}_k) \leftarrow - \sum_{\mathbf{v}_q \in \mathcal{V}_{t}(\mathbf{x}_k)} \frac{A_q}{S_t}\,\log\!\det(\bm{\Sigma}_q)$,\;
  $J_{t+h}(\mathbf{x}_k) \leftarrow - \sum_{\mathbf{v}_q \in \mathcal{V}_{t+h}(\mathbf{x}_k)} \frac{A_q}
  {S_{t+h}}\,\log\!\det(\bm{\Sigma}_q)$\;
  \tcp* {Are-weighted uncertainties}
  $\Delta J_{\mathrm{gain}} \leftarrow \Delta J_{\mathrm{gain}} + \gamma_k\,J_{t+h}(\mathbf{x}_k) - J_t(\mathbf{x}_k)$\;
}
\Return $\Delta J_{\mathrm{gain}}$\;
\end{algorithm2e}

\begin{figure}[!t]
    \centerline{\includegraphics[width=\linewidth]{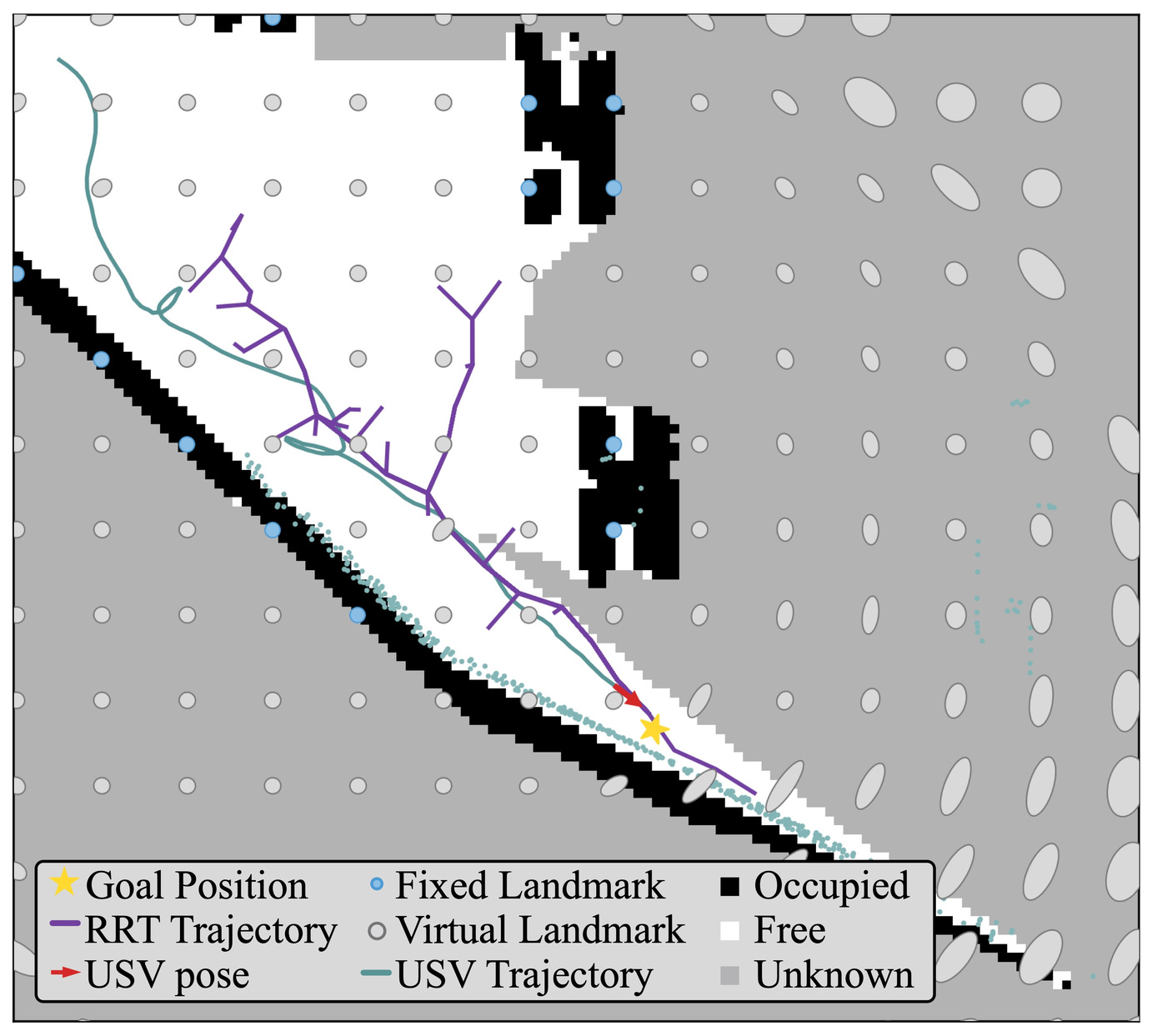}}
    \caption{Example of the EM–planner selecting a goal and generating an RRT trajectory over the VRVM. The USV trajectory (teal), RRT tree (purple), and goal position (yellow star) are shown together with fixed landmarks (blue), virtual landmarks (gray), and the occupancy map, where black, white, and gray denote occupied, free, and unknown regions.}
    \label{fig:RRT} 
\end{figure}

\begin{figure*}[!t]
  \centering

  \subfloat[{Marina world of VRX.}]{%
\begin{minipage}[t]{.7\textwidth}
    \centering
    \includegraphics[width=\linewidth]{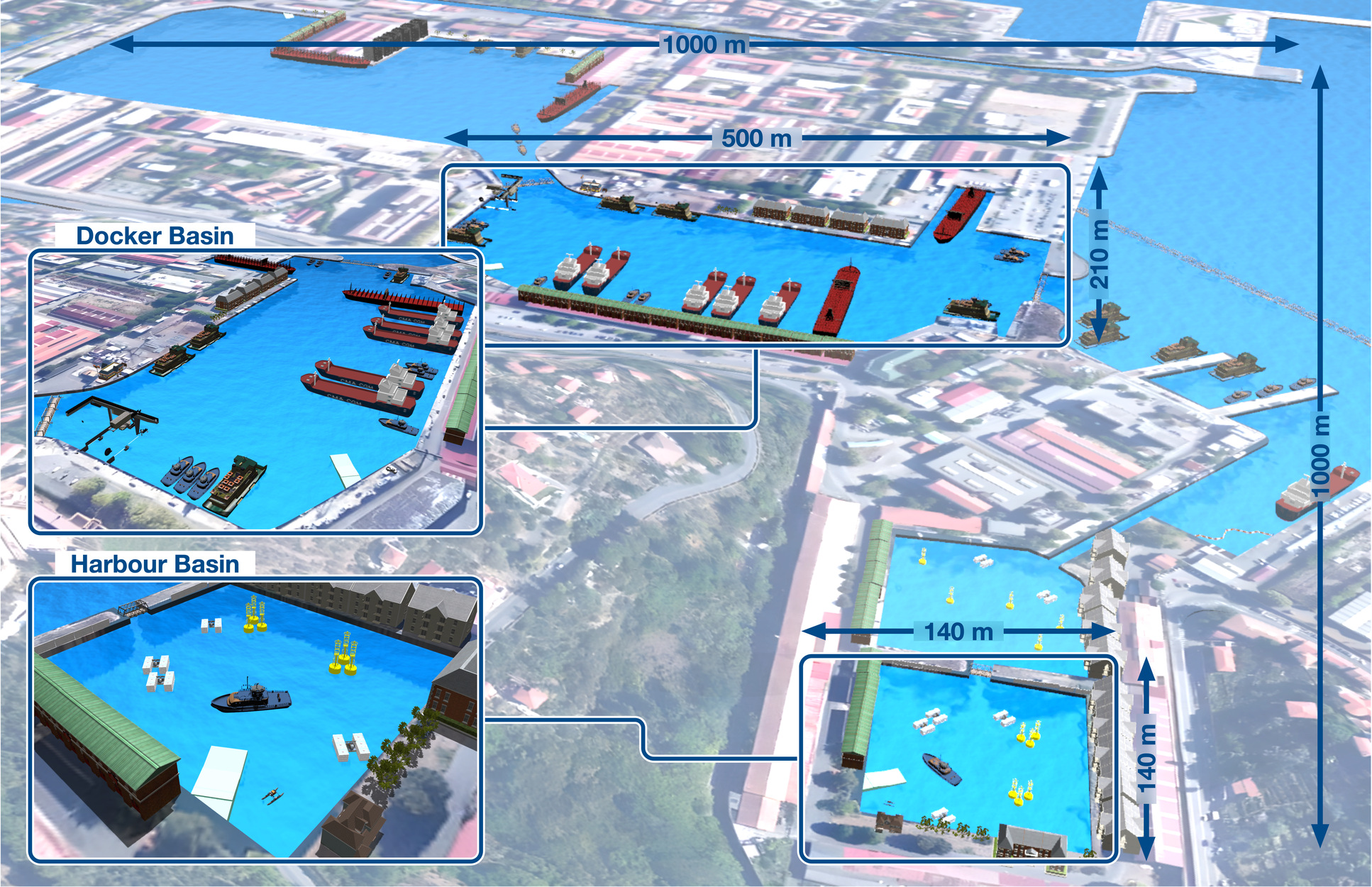}
    
  \end{minipage}%
}\hfill
  \subfloat[{Google map of the harbour.}]{%
\begin{minipage}[t]{.2778\linewidth}
    \centering
    \includegraphics[width=\linewidth]{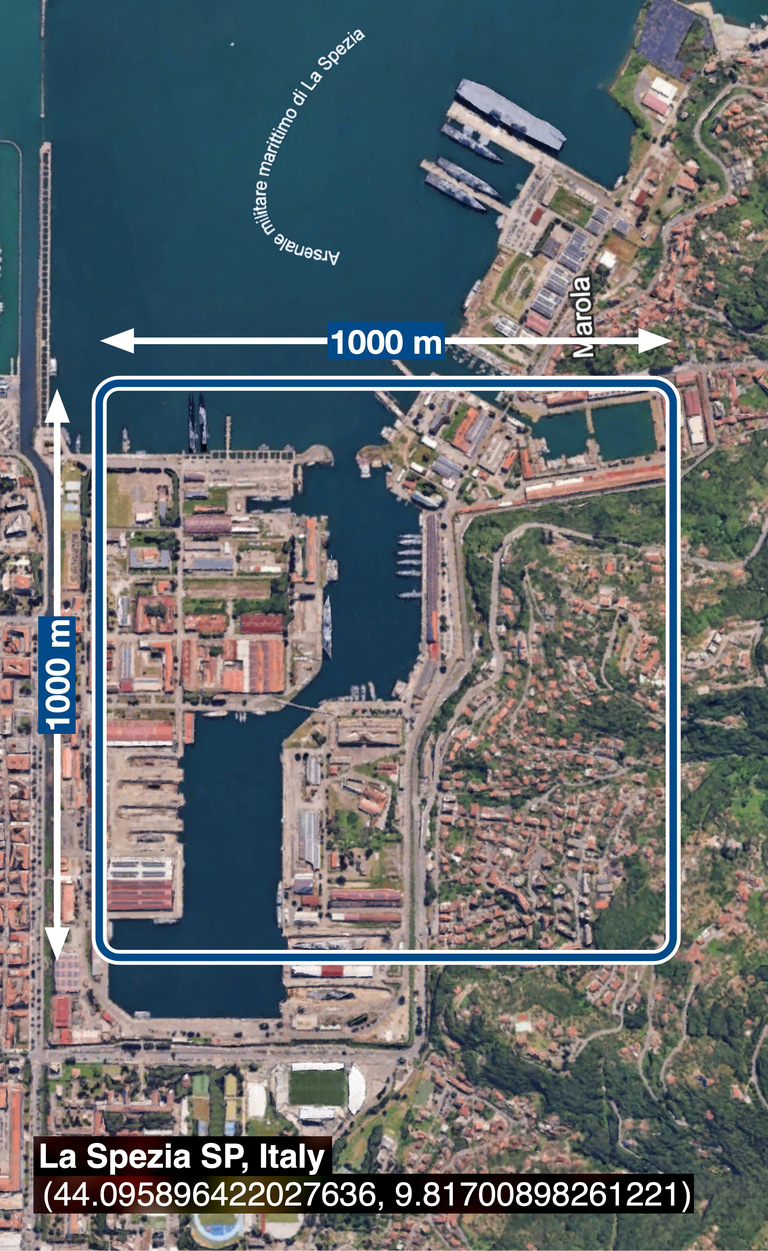}

  \end{minipage}%
}
  \caption{Overview of the simulated maritime environment with corresponding real-world scene taken as reference. Detailed insets illustrate the Docker Basin and Harbour Basin with annotated spatial dimensions.}
  \label{fig:scene_sum}
\end{figure*}

\subsection{Planning and Utility Calculation}
\label{sec:em_planner}
In this section, we summarise the overall planning pipeline of the VRVM. As described in Sec.~\ref{sec:frontier&em}, at each control cycle a finite candidate set $\Pi=\{\pi\}$ is assembled by sampling exploring frontiers located at the boundary between known and unknown regions of the occupancy grid, together with exploit frontiers near previously observed structures to encourage loop closures and improve localisation. For each frontier, a kinodynamically feasible path is generated using a goal-biased RRT alogrithm~\citep{urmson2003approaches} followed by trajectory short-cutting. Candidates that violate collision constraints or curvature limits are discarded. Fig.~\ref{fig:RRT} illustrates the set of end-of-horizon trajectories generated by the RRT planner, shown in purple.

The pose covariance of each end-of-horizon trajectory is propagated using a factor graph optimiser based on the latest occupancy map; in our implementation, we use iSAM2~\citep{kaess2012isam2}. The VRVM is then updated accordingly using a visible-set update along the predicted trajectory.
We evaluate a scalar utility function that balances three objectives: pose uncertainty reduction, area-weighted map uncertainty reduction, and path cost. The planner selects the feasible control sequence with the highest utility value, as defined in Eq.~\ref{eq:utility}.
The trajectory and energy utility terms are defined as:
\begin{align}
U_{\mathrm{traj}}(\pi) &= -\log \det\bigl(\boldsymbol{\Sigma}_{\mathbf{p},t+H}(\pi)\bigr), \\
U_{\mathrm{length}}(\pi) &= -\alpha\mathrm{length}(\pi),
\end{align}
where $\boldsymbol{\Sigma}_{\mathbf{p},t+H}(\pi)$ denotes the predicted pose covariance at the end of the trajectory, and $\alpha$ is a scalar weight that penalises longer paths.

\begin{table*}[t]
\centering
\caption{Key hyper-parameters of the compared exploration baselines. We report only parameters needed to interpret evaluation and discussion (representation/uncertainty modelling, candidate-set size and scoring trade-offs, and replanning triggers).}
\label{tab:baseline_params}
\vspace{-0.5em}
\footnotesize
\setlength{\tabcolsep}{3.0pt}
\renewcommand{\arraystretch}{1.18}

\begin{tabularx}{\textwidth}{
>{\raggedright\arraybackslash}p{0.20\textwidth}
Y Y Y Y Y}
\toprule
\multirow{2}{*}{\textbf{Category / Parameter}} &
\multicolumn{2}{c}{\textbf{Virtual-map utilities (EM planner)}} &
\multicolumn{3}{c}{\textbf{Grid-based baselines}} \\
\cmidrule(lr){2-3}\cmidrule(lr){4-6}
& \textbf{VRVM} & \textbf{UVM} & \textbf{NF} & \textbf{NBV} & \textbf{FSMI} \\
&   &\citep{wang2022virtual} & \citep{ijrr2013efficient}  & \citep{ral2019NBV} & \citep{zhang2020fsmi} \\
\midrule

Workspace bounds [m] &
\makecell[l]{$[-600,600]$\\$\times$ $[-600,600]$} &
\makecell[l]{$[-600,600]$\\$\times$ $[-600,600]$} &
-- & -- & -- \\

Base resolution [m] &
$1.0$ & $1.0$ & $1.0$ & $1.0$ & $1.0$ \\

Prior std.\ $\sigma_0$ [m] &
$0.5$ & $0.5$ & -- & -- & -- \\

VM update &
\makecell[l]{update $1.0$} &
\makecell[l]{update $1.0$} &
-- & -- & -- \\

CI fusion &
\makecell[l]{on (logdet)\\$\Delta\omega{=}0.02$, $\kappa{=}1.0$} &
\makecell[l]{on (logdet)\\$\Delta\omega{=}0.02$, $\kappa{=}1.0$} &
-- & -- & -- \\

Variable resolution / locking &
\makecell[l]{quadtree, $d_{\max}{=}7$\\occ lock: $occ_{\mathrm{th}}{=}51$} &
\makecell[l]{uniform grid\\(no split /\\no area weight)} &
-- & -- & -- \\

Sensor model for scoring &
\makecell[l]{range $30$ m\\FOV $360^\circ$} &
\makecell[l]{range $30$ m\\FOV $360^\circ$} &
-- &
\makecell[l]{range $30$ m\\FOV $360^\circ$} &
\makecell[l]{range $30$ m\\FOV $360^\circ$} \\

\midrule

Replanning trigger \& reach tol. &
\makecell[l]{plan $5$ Hz\\$tol_{xy}{=}4$ m} &
\makecell[l]{plan $5$ Hz\\$tol_{xy}{=}4$ m} &
\makecell[l]{timer $5$ Hz\\yaw $360^\circ$} &
\makecell[l]{auto trigger\\yaw $20^\circ$\\speed $0.3$ m/s} &
\makecell[l]{timer $5$ Hz\\yaw $360^\circ$} \\

Candidate set size (per cycle) &
\makecell[l]{RRT paths $100$\\max iter $1000$} &
\makecell[l]{RRT paths $100$\\max iter $1000$} &
\makecell[l]{max frontiers $100$\\min cluster $8$\\4-connected} &
\makecell[l]{frontier ds $100$\\radii $[5,10,15]$ m} &
\makecell[l]{max cand.\ $100$\\radii $[5,10,15]$ m} \\

Candidate geometry / discretisation &
\makecell[l]{RRT step $4$ m\\sample rad.\ $80$ m} &
\makecell[l]{RRT step $4$ m\\sample rad.\ $80$ m} &
\makecell[l]{A* on inflated grid\\edge step $80$ m} &
\makecell[l]{view step $3^\circ$\\ray step $80$ m} &
\makecell[l]{cand.\ step $15^\circ$\\ray step $80$ m\\ray ang.\ step $6^\circ$} \\

Path post-processing &
\makecell[l]{shortcut iters $200$\\spline $\Delta t{=}0.1$} &
\makecell[l]{shortcut iters $200$\\spline $\Delta t{=}0.1$} &
-- & -- & -- \\

\midrule

Scoring form and weights &
\makecell[l]{EM weights:\\$\alpha{=}0.3$, $\beta{=}0.5$\\$\gamma{=}1.0$, $g_d{=}0.1$\\VM agg.: \\\textbf{area-weighted}} &
\makecell[l]{EM weights:\\VM agg.: uniform} &
\makecell[l]{frontier goal:\\$g^\ast=\arg\min L(g)$} &
\makecell[l]{gain--cost:\\$w_u{=}1.0$, $w_{ov}{=}0.5$\\$\alpha{=}0.05$, $\beta{=}0.2$} &
\makecell[l]{MI--cost:\\$\lambda_L{=}0.08$, $\beta{=}0.2$\\priors\\$(0.03,0.97,0.5)$\\log-odds $(+2,-2)$} \\

\bottomrule
\end{tabularx}
\vspace{-0.8em}
\end{table*}

We execute $\pi^\star$ in a receding-horizon fashion: after traversing a short segment, the SLAM smoother assimilates new odometry and loop-closure factors, the VRVM fuses visible-set information, the candidate frontiers are refreshed, and the VRVM-based EM planner evaluates the candidate control sequences.

The computation required to evaluate the utilities of all candidates scales with the number of sampled frontiers. The prediction cost is
$
O\!\Big(\sum_{\pi\in\Pi}\sum_{k} \big|\mathcal{V}(\mathbf{x}_k)\big|\Big)
$,
with only $2{\times}2$ operations on VRVM leaves, plus the incremental smoothing updates provided by the SLAM back-end. 
By introducing the area-weighted map valuation and the visible-set update, we reduce the computational cost relative to EM planners whose operations scale with the map size.

\begin{figure*}[!t]
    \centering
  \subfloat[{Environment overview of the dense harbor environment.\label{fig:dense_scene_overview}}]{%
\begin{minipage}[t]{.48\linewidth}
    \centering
    \includegraphics[width=\linewidth]{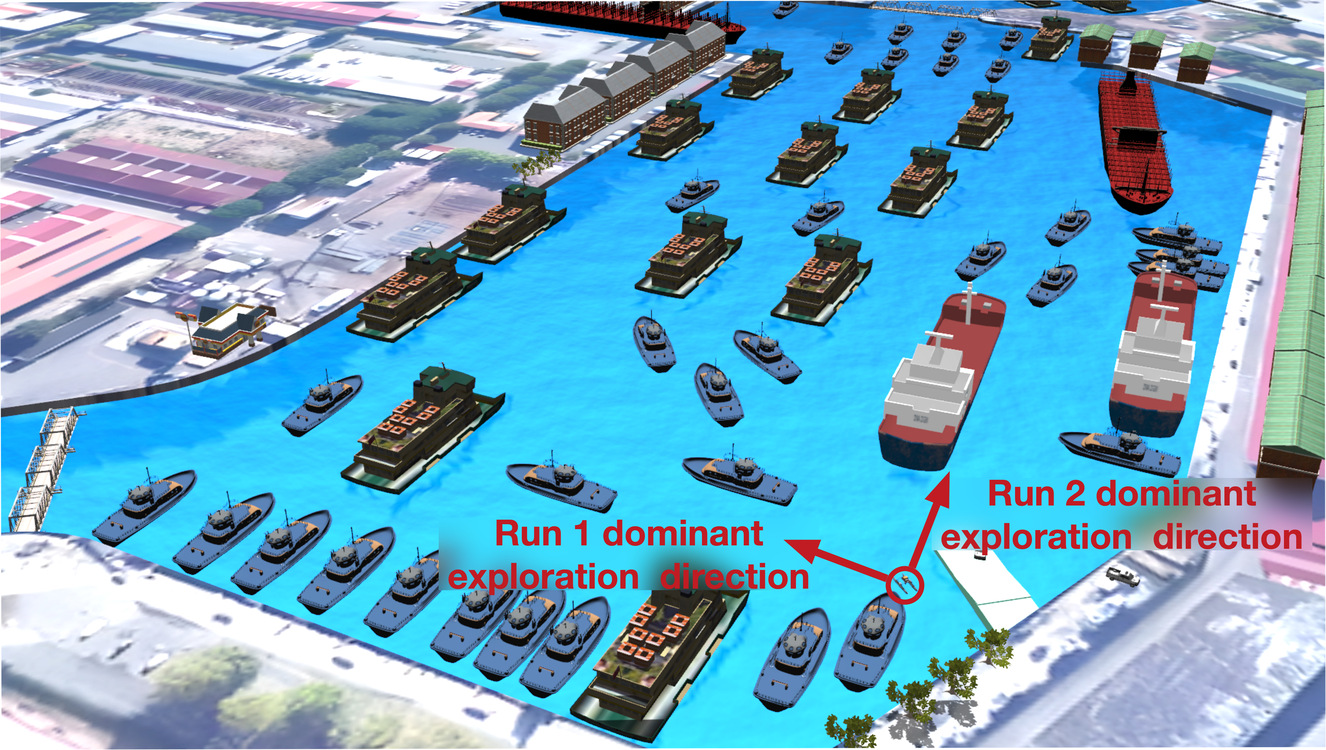}

  \end{minipage}%
}\hfill
\subfloat[{Some marine obstacles.\label{fig:dense_scene_google}}]{%
\begin{minipage}[t]{.48\linewidth}
    \centering
    \includegraphics[width=\linewidth]{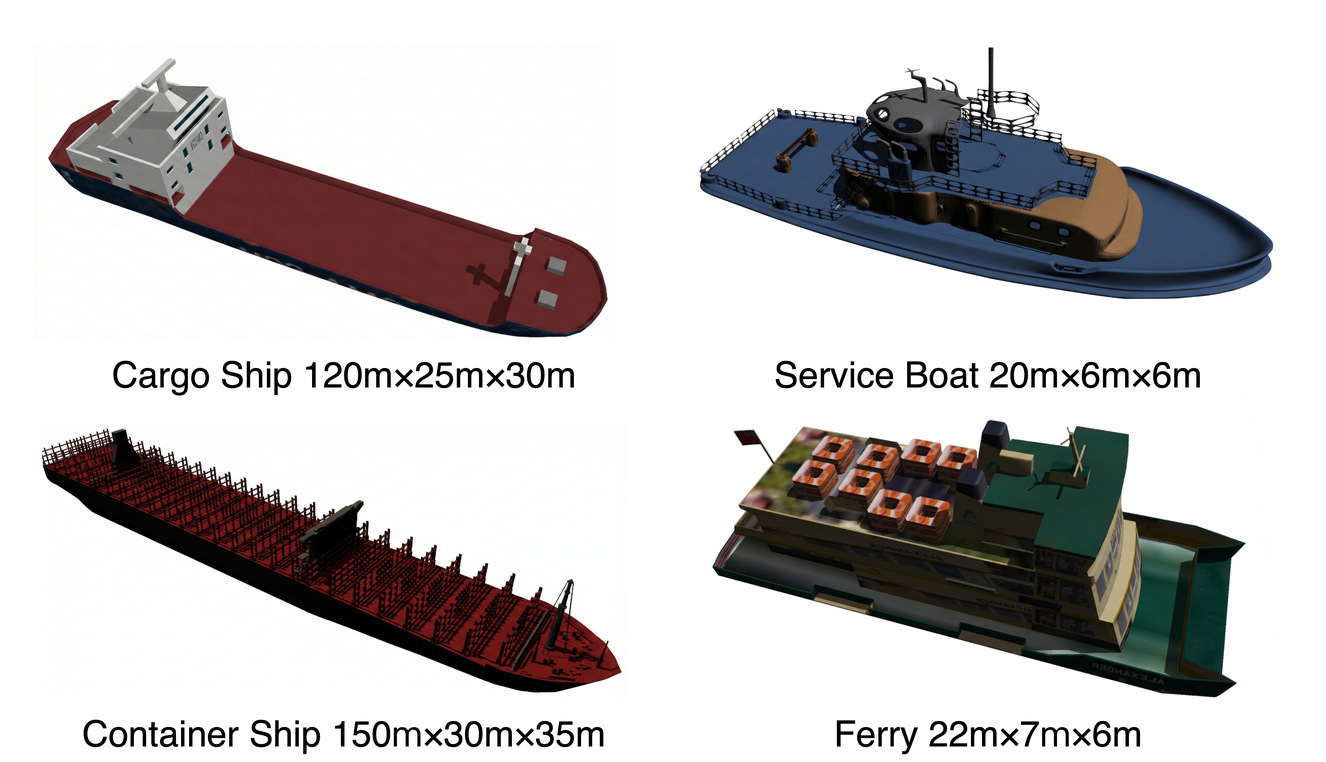}

  \end{minipage}%
}\par\medskip
\subfloat[{Run 1 using VRVM.\label{fig:vrvm_dense_a_env1}}]{%
\begin{minipage}[t]{.48\linewidth}
    \centering
    \includegraphics[width=\linewidth,height=.52\textheight,keepaspectratio]{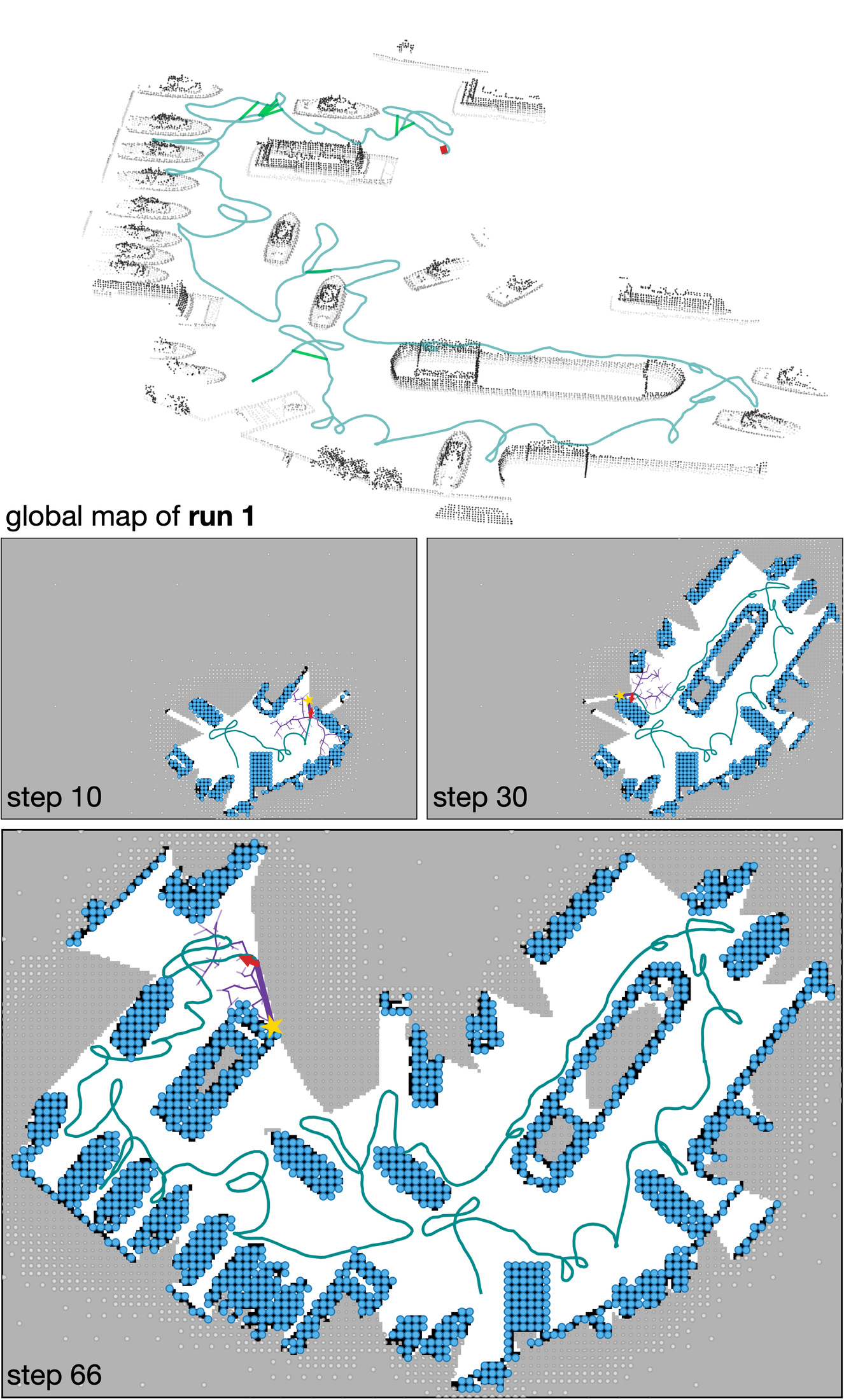}

  \end{minipage}%
}\hfill
\subfloat[{Run 2 using VRVM.\label{fig:vrvm_dense_a_env2}}]{%
\begin{minipage}[t]{.48\linewidth}
    \centering
    \includegraphics[width=\linewidth,height=.52\textheight,keepaspectratio]{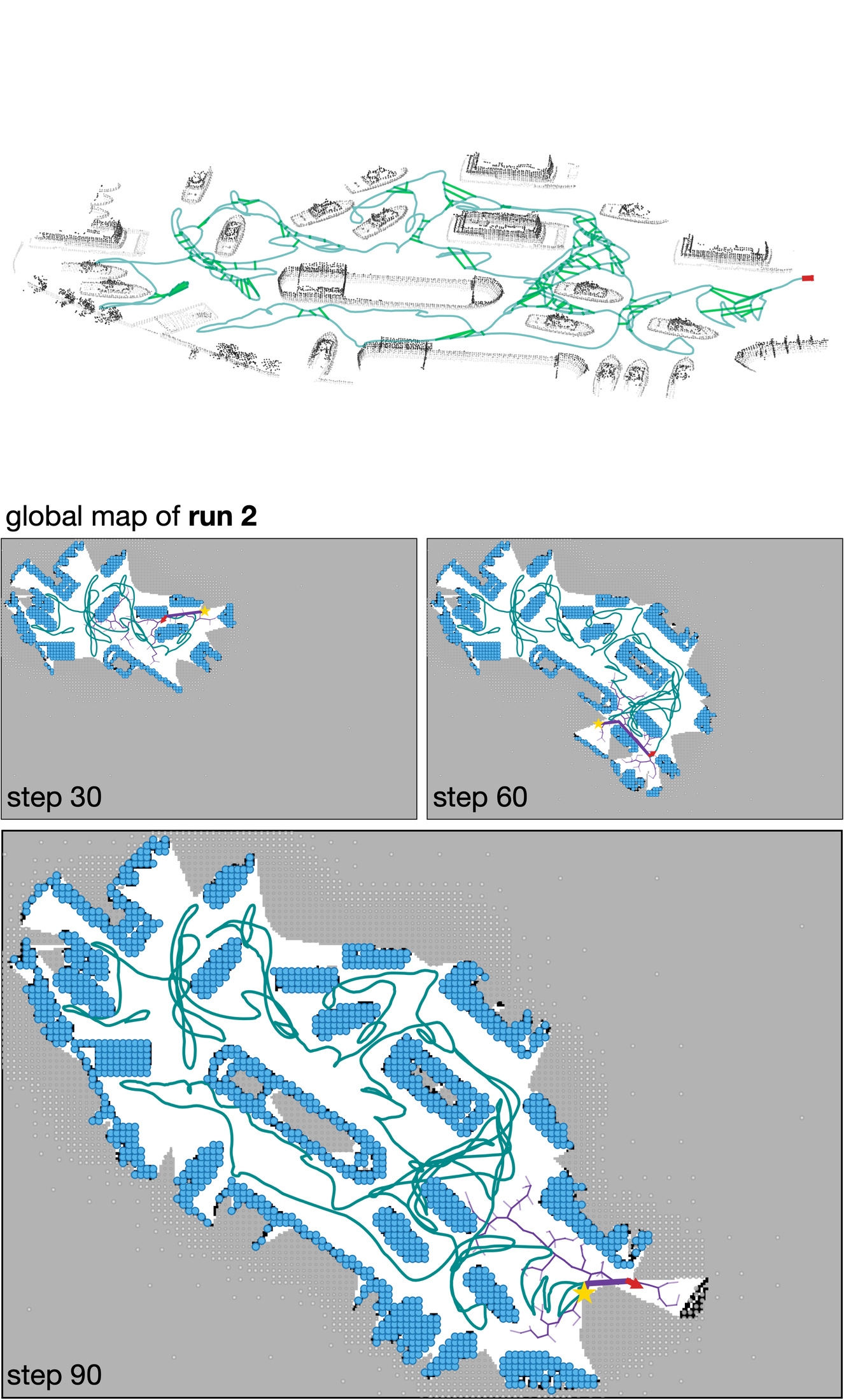}

  \end{minipage}%
}\par\smallskip
\begin{minipage}[t]{\linewidth}
    \centering
    \includegraphics[width=.8\linewidth]{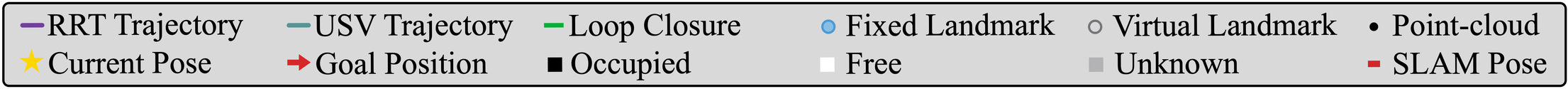}
  \end{minipage}
    \caption{Exploration in the dense Docker Basin environment. (a) Environment overview with the start position and the dominant exploration direction in each run. Run~1 snapshots are shown at steps 10, 30, and 66 with the final SLAM point-cloud map; Run~2 snapshots are shown at steps 30, 60, and 90 with the final SLAM point-cloud map.}
  \label{fig:exploration_vrvm_dense}
\end{figure*}

\section{Experimental Evaluation}
\label{sec:experiments}
We evaluate the proposed VRVM framework through controlled simulation benchmarks and two BlueBoat field runs.
The simulation benchmarks examine the trade-off between coverage expansion and localisation and mapping accuracy, together with computational demand.
The field runs examine onboard operation and goal selection without GNSS input to SLAM or planning, with manual RC steering following planner guidance.

Since achieving controlled, repeatable results in maritime environments is often precluded by traffic, safety constraints, and the difficulty of acquiring high-fidelity ground truth, we rely on the Virtual RobotX (VRX)~\citep{bingham19toward}, a {gazebo}\footnote{\url{https://gazebosim.org/home}} 3D simulator for large-scale benchmarking.
Furthermore, we evaluate the computational burden on both a desktop PC and a constrained embedded system to assess the framework’s suitability for long-horizon operation on resource-constrained robotic hardware.

\subsection{Simulation Setup}
\label{sec:exp_setup}
We conduct the simulation experiments in the VRX environment using the standard WAM-V USV platform equipped with a 16-beam 3D LiDAR and an IMU, as shown in Fig.~\ref{fig:vrvm}.
To capture the structural diversity of real near-coastal waters, we use a large, composite water-area setting (Fig.~\ref{fig:scene_sum}). It contains multiple regions with distinct geometric features, separated by narrow passages that constrain transitions and induce qualitatively different localisation conditions.
This layout is specifically designed to test the exploitation-exploration trade-off.

For state estimation, we use LIO-SAM~\citep{shan2020liosam} with GNSS inputs disabled; its 6-DoF estimates are projected onto the horizontal plane to form the planner’s $SE(2)$ state. In parallel, we integrate the registered point clouds into a fixed-resolution 3D OctoMap~\citep{hornung2013octomap} and then project it to a 2D occupancy grid. This grid is the shared interface used by all planners for collision checking and coverage, so that differences in results cannot be attributed to inconsistent map sources or evaluation back-ends.

We compare VRVM against four representative exploration baselines: Next Frontier (NF)~\citep{ijrr2013efficient}, Next-Best-View (NBV)~\citep{ral2019NBV}, Fast Shannon Mutual Information (FSMI)~\citep{zhang2020fsmi}, and UVM~\citep{wang2022virtual}. These methods share the same upstream SLAM backbone, the same occupancy-grid interface, and the same motion and safety constraints; their differences arise only from how uncertainty is modelled and how the utility is constructed and optimised. 
To isolate the impact of the variable-resolution representation itself, VRVM and UVM use the same virtual-map footprint, update frequency, and measurement-noise model. 
Key hyper-parameters are reported in Table~\ref{tab:baseline_params}.

Under these controlled conditions, we report metrics that jointly reflect task outcome and computational burden, including coverage, map error, planning time, and memory growth.
For the simulation benchmarks, runs terminate when the planner fails to find a new exploration path or localisation failure occurs.

Desktop experiments are performed on an Intel i7-12700H CPU without GPU acceleration, running ROS Noetic on Ubuntu 20.04. 
We further deploy the VRVM/UVM exploration stack on an embedded platform—{Raspberry Pi 4 Model B (4GB)}\footnote{\url{https://www.raspberrypi.com/products/raspberry-pi-4-model-b/}} .
As prior work establishes that NF, NBV, and FSMI incur low per-iteration computational and memory overhead, 
the embedded evaluation focuses specifically on the scalability gap between UVM and VRVM during sustained, wide-area exploration on resource-constrained hardware. %

\subsection{USV Model and Simulation Platform}
\label{sec:usv_model_sim}
Although a USV is a six-degree-of-freedom (6-DoF) rigid body in general,
we restrict the planning model to horizontal-plane motion.
Heave, roll, pitch, and wave-induced motions are treated as disturbances
that primarily affect sensing, while their impact on planar navigation is mitigated by the low-level controller.

Adopting the USV state definition introduced in Sec.~\ref{sec:sensor}, the planar configuration is parameterised by
$\mathbf{x}=[x,y,\psi]^\top$.
The body-fixed planar velocity is
$\boldsymbol{\nu}=[u,v,r]^\top \in \mathbb{R}^3$,
where $u$ and $v$ denote surge and sway velocities,
and $r$ denotes the yaw rate.
The horizontal-plane manoeuvring model for surface vessels are governed by the standard formulation~\citep{fossen2011handbook}:
\begin{equation}
\mathbf{M}\dot{\boldsymbol{\nu}}
+\mathbf{C}(\boldsymbol{\nu})\boldsymbol{\nu}
+\mathbf{D}(\boldsymbol{\nu})\boldsymbol{\nu}
=\boldsymbol{\tau}+\mathbf{w},
\label{eq:usv_dyn}
\end{equation}
where $\mathbf{M}$ is the inertia matrix,
$\mathbf{C}(\cdot)$ collects Coriolis and centripetal terms,
$\mathbf{D}(\cdot)$ represents hydrodynamic damping,
$\boldsymbol{\tau}$ is the generalised control input,
and $\mathbf{w}$ models environmental disturbances
(in this work, primarily wind and waves).
While the exploration objective is formulated over the planar pose
$\mathbf{x}$, Equation~\ref{eq:usv_dyn} defines the underlying motion-feasible constraints.

Propulsion is realised via two aft thrusters in a differential configuration.
Accordingly, we adopt an underactuated planar input
$\boldsymbol{\tau}=[\tau_u,\;0,\;\tau_r]^\top$,
where $\tau_u$ is the surge force and $\tau_r$ is the yaw moment.
Let $T_L$ and $T_R$ denote the left and right thrust magnitudes,
and $b$ the lateral separation between the thrusters. The mapping is
\begin{equation}
\tau_u = T_L + T_R,\qquad
\tau_r = \frac{b}{2}(T_R - T_L).
\label{eq:diff_thrust}
\end{equation}

For planning with step size $\Delta t$, we use the discrete-time propagation
\begin{equation}
\mathbf{x}_{k+1}
=\mathbf{x}_k+\Delta t\,\mathbf{R}(\psi_k)\boldsymbol{\nu}_k,
\label{eq:disc_kin}
\end{equation}
together with bounded-speed and bounded-turn-rate constraints
\begin{equation}
|u_k|\le u_{\max},\quad |r_k|\le r_{\max},
\label{eq:bounds_ur}
\end{equation}
to ensure feasibility on the WAM-V platform.

Actuation is implemented through the Gazebo thrust interface
(\nolinkurl{usv_gazebo_thrust_plugin}), which maps normalised
per-thruster commands
into applied forces.
Planner-generated waypoints are converted into left/right thrust commands
by a shared reference-tracking controller.
All methods use identical actuation settings and controller parameters,
so performance differences reported in this section
are attributable to the exploration and planning layer rather than
low-level control.

\subsection{Test Scenarios}
\label{sec:test_scenarios}

As shown in Fig.~\ref{fig:scene_sum}, the \emph{Marina} world\footnote{\url{https://github.com/osrf/vorc}} from VRX is selected as the global testbed. 
It provides a realistic simulation of the harbour environment in La Spezia, Italy, covering approximately $1000\,\mathrm{m}\times1000\,\mathrm{m}$.
Unless otherwise stated, to ensure repeatability, we keep the environmental conditions fixed across runs, including the wave and flow settings (wave height 0.3\,$\mathrm{m}$, current speed 1.5\,$\mathrm{m/s}$, wind speed 3.0\,$\mathrm{m/s}$).

The environment is populated with six representative maritime obstacles (Fig.~\ref{fig:exploration_vrvm_dense}): 
(i) \emph{Tugboat} (service vessel), (ii) \emph{Container Barge} (barge-like cargo platform), (iii) \emph{Cargo Ship} (small freighter), (iv) \emph{Bulk Carrier} (large ship hull), (v) \emph{Navigation Buoy} (fixed marker buoy), and (vi) \emph{Floating Platform} (modular raft/solar platform). 
We derive two task-focused sub-areas from Marina to support different evaluation goals (Fig.~\ref{fig:scene_sum}). 
The \emph{Harbour Basin} (approximately $140\,\mathrm{m}\times 140\,\mathrm{m}$) is used for functionality validation.
Harbour Basin is deliberately kept small and structure-rich using obstsacle types (i), (v), and (vi), so that the LiDAR--IMU SLAM backbone remains reliable throughout the run. %
Removing SLAM degradation allows a controlled initial validation of VRVM's key functional behaviour.

The \emph{Docker Basin} region (approximately $210\,\mathrm{m}\times 500\,\mathrm{m}$) serves as comparative evaluation of all methods. 
Utilizing obstacle types (i)--(iv), it is intentionally designed with long traversals through open water and occlusion corridors, so that localisation degradation can occur.
To probe robustness under different structural supports, we instantiate three Docker Basin variants (Fig.~\ref{fig:benchmark_sparse}, \ref{fig:benchmark_moderate}, \ref{fig:benchmark_dense}) by changing the obstacle density. As the scene becomes sparser, the USV is more likely to enter regions with insufficient geometric features within its sensing range. Such conditions increase the probability of localisation loss and, consequently, challenge the exploration policies.

\subsection{Weight Study}
\label{sec:weight_study}

\begin{figure*}[!t]
  \centering

    \subfloat[{$\lambda_{\mathrm{map}} = 1$\label{fig:gamma1}}]{%
\begin{minipage}[t]{\textwidth}
    \centering
    \includegraphics[width=1\linewidth]{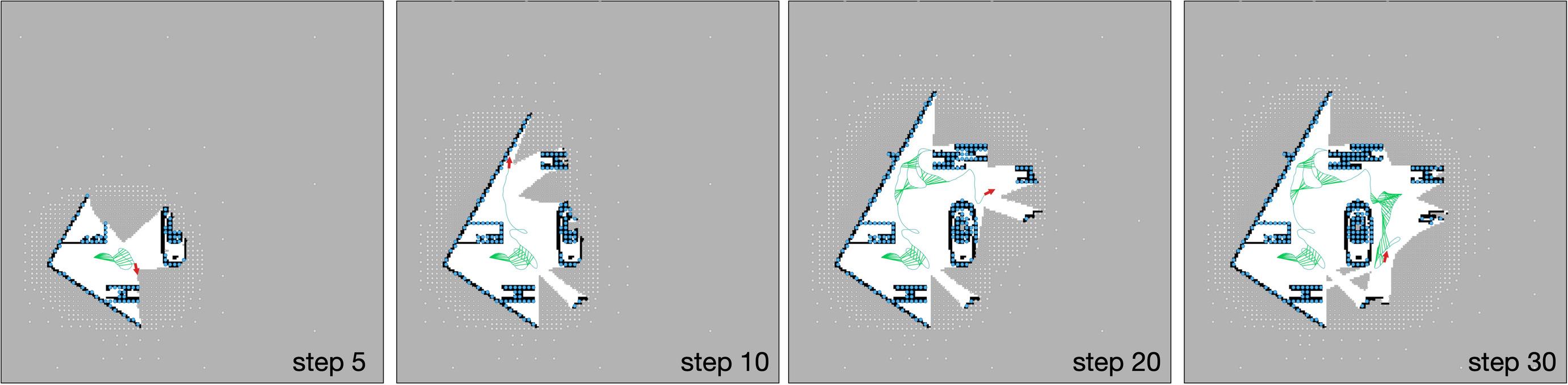}

  \end{minipage}%
}\par\smallskip
\subfloat[{$\lambda_{\mathrm{map}} = 5$\label{fig:gamma5}}]{%
\begin{minipage}[t]{\textwidth}
    \centering
    \includegraphics[width=1\linewidth]{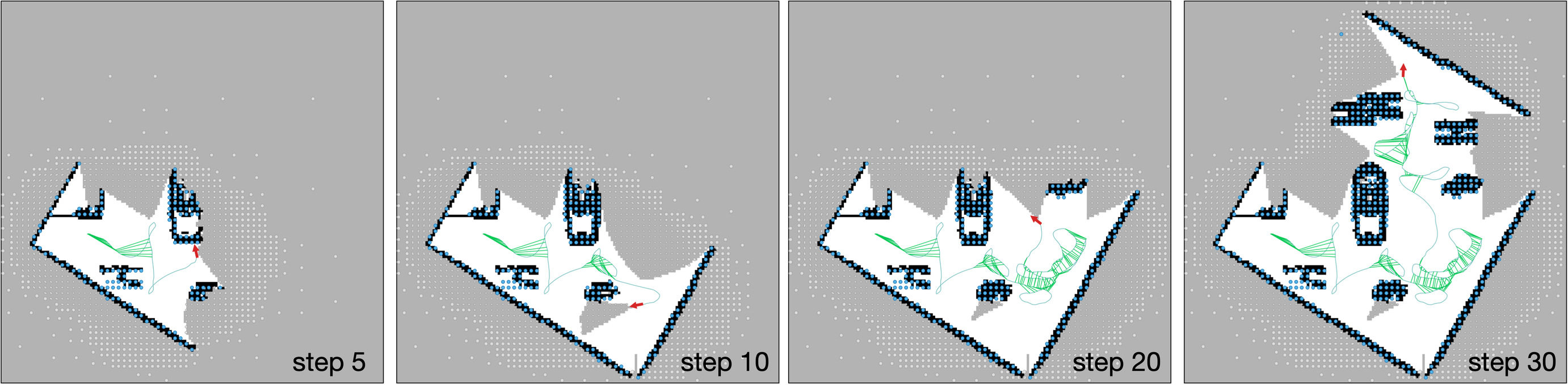}

  \end{minipage}%
}\par\smallskip
\subfloat[{$\lambda_{\mathrm{map}} = 15$\label{fig:gamma15}}]{%
\begin{minipage}[t]{\textwidth}
    \centering
    \includegraphics[width=1\linewidth]{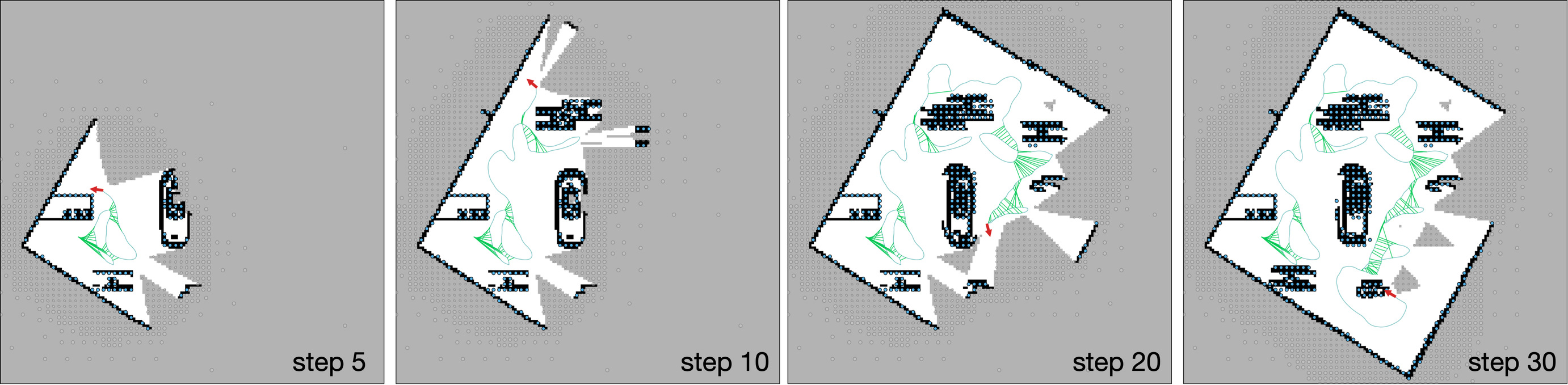}

  \end{minipage}%
}\par\smallskip
\begin{minipage}[t]{\textwidth}
    \centering
    \includegraphics[width=.6\linewidth]{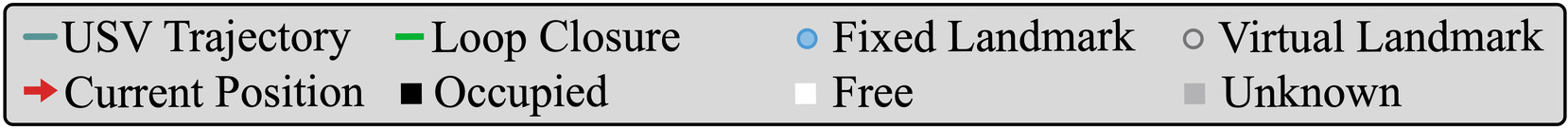}
  \end{minipage}

  \caption{Effect of the mapping weight $\lambda_{\mathrm{map}}$ on exploration behaviour in the square-sized Harbour Basin scene.
  Each row fixes $\lambda_{\mathrm{map}} \in \{1, 5, 15\}$, while columns show planning steps 5, 10, 20, and 30.}
  \label{fig:area_weight}
\end{figure*}

\begin{figure*}[!t]
  \begin{minipage}[t]{\linewidth}
    \centering
    \includegraphics[width=.9\linewidth,height=.78\textheight,keepaspectratio]{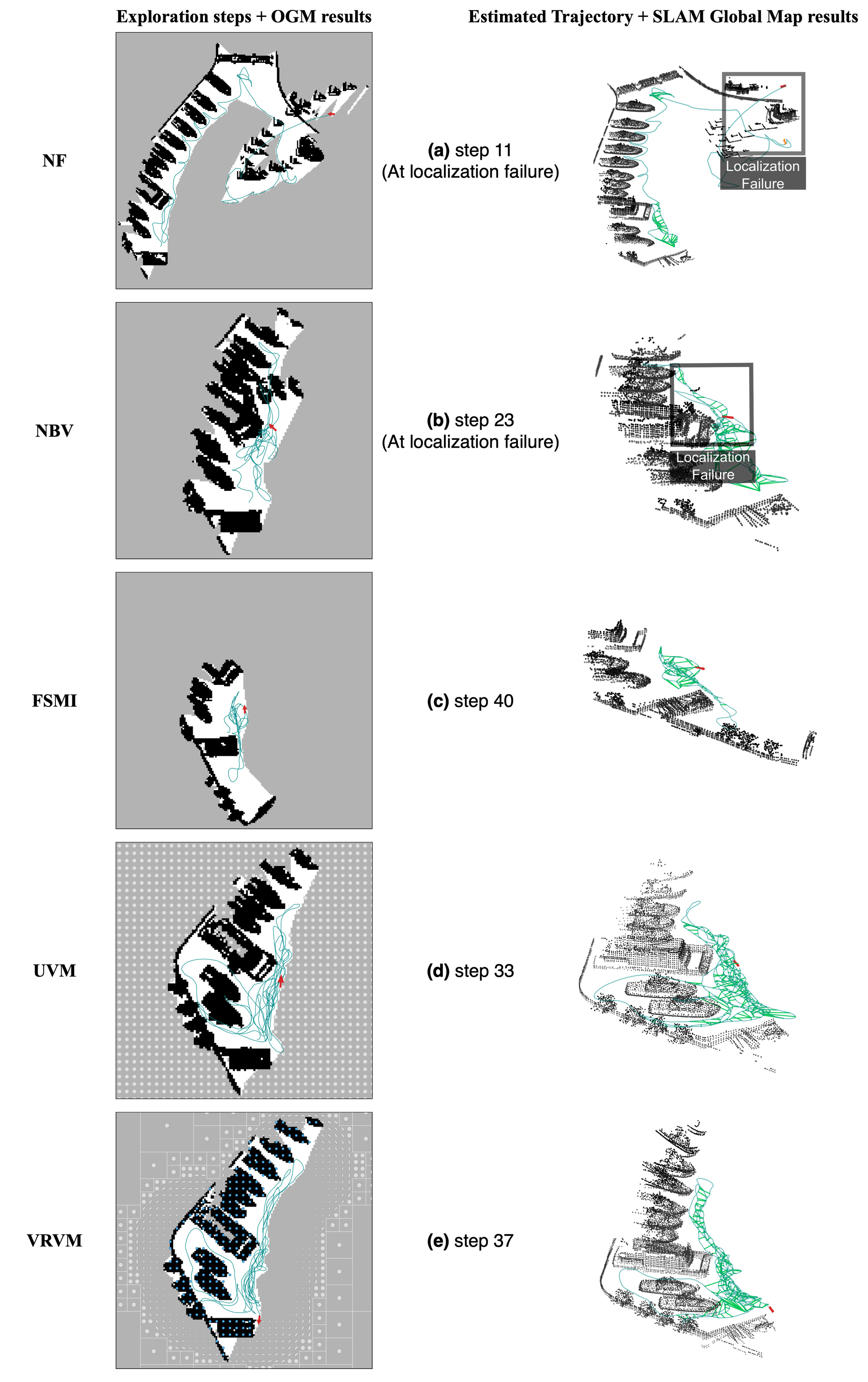}
  \end{minipage}
  \begin{minipage}[t]{\linewidth}
    \centering
    \includegraphics[width=.6\linewidth]{fig/SquareSizeMapGamma/legend_harbour_basin.jpg}
  \end{minipage}
\caption{Comparison of five exploration methods in the \textbf{sparse} Harbour Basin scenario (left: final exploration trajectory over the occupancy grid map; right: estimated SLAM trajectory and map).} 
\label{fig:benchmark_sparse} 
\end{figure*}

\begin{figure*}[!t]
  \begin{minipage}[t]{\linewidth}
    \centering
    \includegraphics[width=.9\linewidth,height=.78\textheight,keepaspectratio]{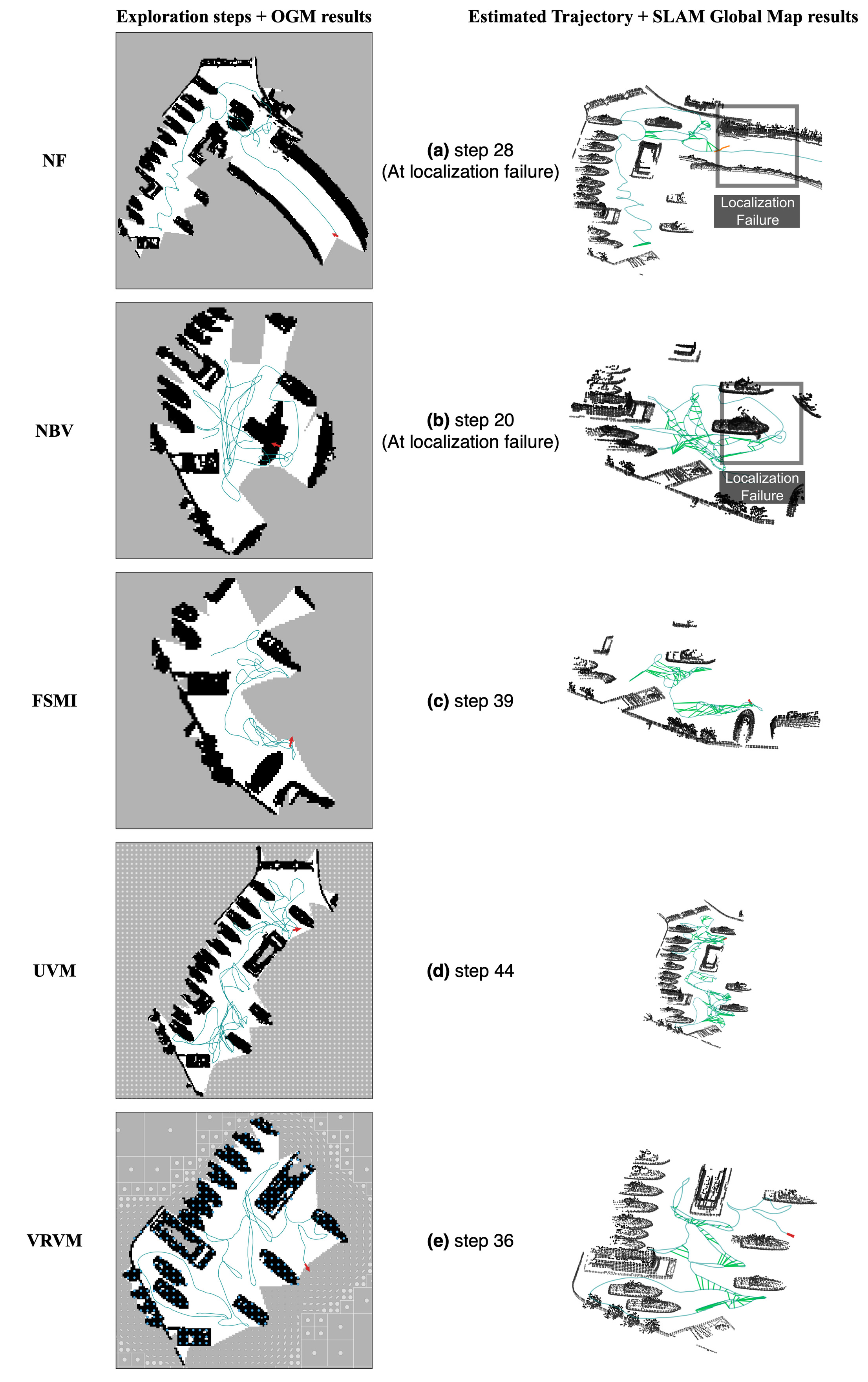}
  \end{minipage}
  \begin{minipage}[t]{\linewidth}
    \centering
    \includegraphics[width=.6\linewidth]{fig/SquareSizeMapGamma/legend_harbour_basin.jpg}
  \end{minipage}
\caption{Comparison of five exploration methods in the \textbf{moderate} Harbour Basin scenario (left: final exploration trajectory over the occupancy grid map; right: estimated SLAM trajectory and map).}
    \label{fig:benchmark_moderate} 
\end{figure*}

\begin{figure*}[!t]
  \begin{minipage}[t]{\linewidth}
    \centering
    \includegraphics[width=.9\linewidth,height=.78\textheight,keepaspectratio]{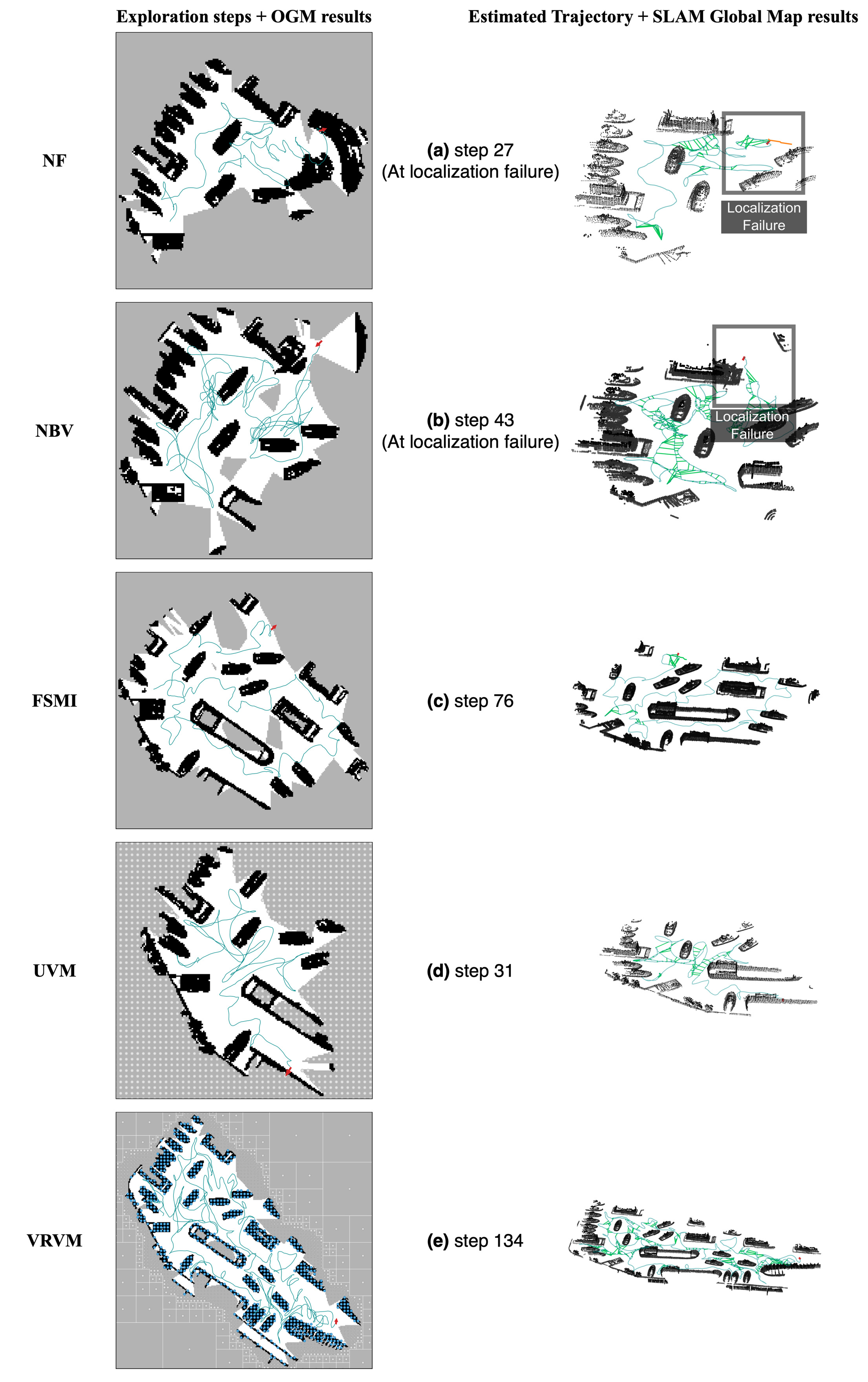}
  \end{minipage}
  \begin{minipage}[t]{\linewidth}
    \centering
    \includegraphics[width=.6\linewidth]{fig/SquareSizeMapGamma/legend_harbour_basin.jpg}
  \end{minipage}
    \caption{Comparison of five exploration methods in the \textbf{dense} Harbour Basin scenario (left: final exploration trajectory over the occupancy grid map; right: estimated SLAM trajectory and map).}
    \label{fig:benchmark_dense}     
\end{figure*}

To analyze the influence of the mapping weight $\lambda_{\mathrm{map}}$,
we conduct a parameter study in the \emph{Harbour Basin} scene.
All parameters other than $\lambda_{\mathrm{map}}$ are held fixed.
We evaluate three representative values, $\lambda_{\mathrm{map}}\in\{1,5,15\}$, and visualise the exploration state at planning steps $\{5,10,20,30\}$ in Fig.~\ref{fig:area_weight}. 

When $\lambda_{\mathrm{map}}=1$, the utility is dominated by the localisation- and cost-related terms, so the planner tends to favour short, low-risk traversals that maintain a well-conditioned trajectory estimate. As shown in Fig.~\ref{fig:gamma1}, exploration expands steadily but remains conservative: progress is concentrated around nearby frontiers, and the explored region grows without aggressively seeking out the most mapping-informative boundaries.

Increasing the mapping weight to $\lambda_{\mathrm{map}}=5$ yields a more balanced exploration behaviour. In Fig.~\ref{fig:gamma5}, the vehicle expands the explored region more decisively, reaching a wider set of frontiers under the same number of planning steps. Qualitatively, this setting encourages trajectories that better exploit available structure within the sensor range while still maintaining reasonable travel efficiency, resulting in faster spatial growth of the explored free space. %

With a large mapping weight $\lambda_{\mathrm{map}}=15$, the planner becomes strongly mapping-driven. As illustrated in Fig.~\ref{fig:gamma15}, candidate selection is biased toward frontiers expected to provide the largest mapping benefit under the VRVM. This promotes more aggressive expansion towards structure-bearing boundaries and reduces time spent on trajectories offering negligible information gain. In this controlled basin, the resulting behaviour produces rapid spatial coverage by step 30, at the expense of longer traversals and reduced emphasis on purely cost-efficient local expansion.

Overall, this study confirms that $\lambda_{\mathrm{map}}$ provides an intuitive and effective control over exploration behaviour: smaller values yield conservative, cost- and localisation-dominated exploration, whereas larger values drive more mapping-oriented expansion. Based on the qualitative trade-off observed in Harbour Basin, we set $\lambda_{\mathrm{map}}=5$ for the remaining simulation experiments unless stated otherwise.

\subsection{Benchmark Results}
\begin{figure*}[!t]
  \centering

  \subfloat[{Map Error - Sparse\label{fig:map_error_1_1}}]{%
\begin{minipage}[t]{0.31\textwidth}
    \centering
    \includegraphics[width=\textwidth]{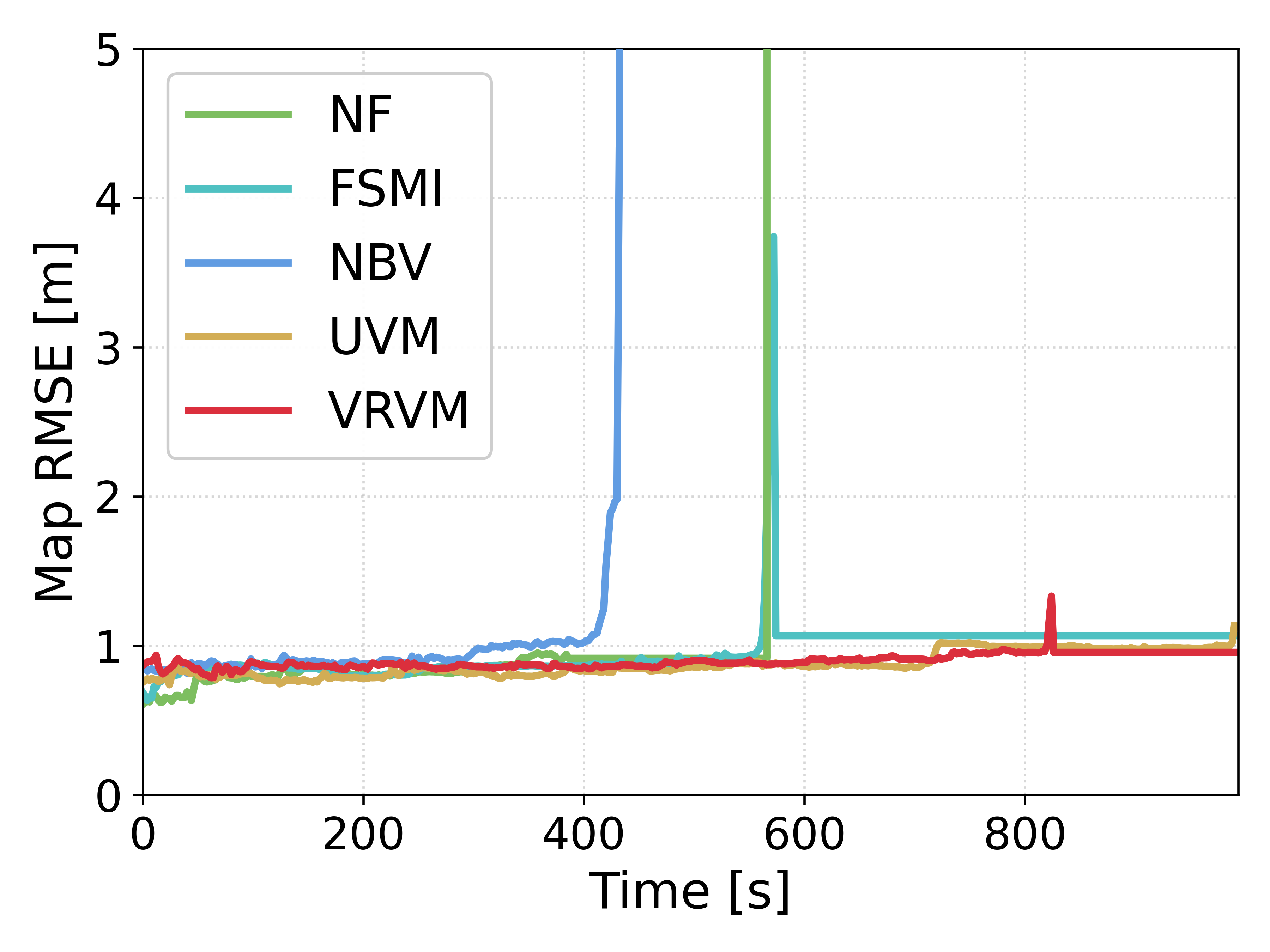}

  \end{minipage}%
}
  \hfill
  \subfloat[{Map Error - Moderate\label{fig:map_error_1_2}}]{%
\begin{minipage}[t]{0.31\textwidth}
    \centering
    \includegraphics[width=\textwidth]{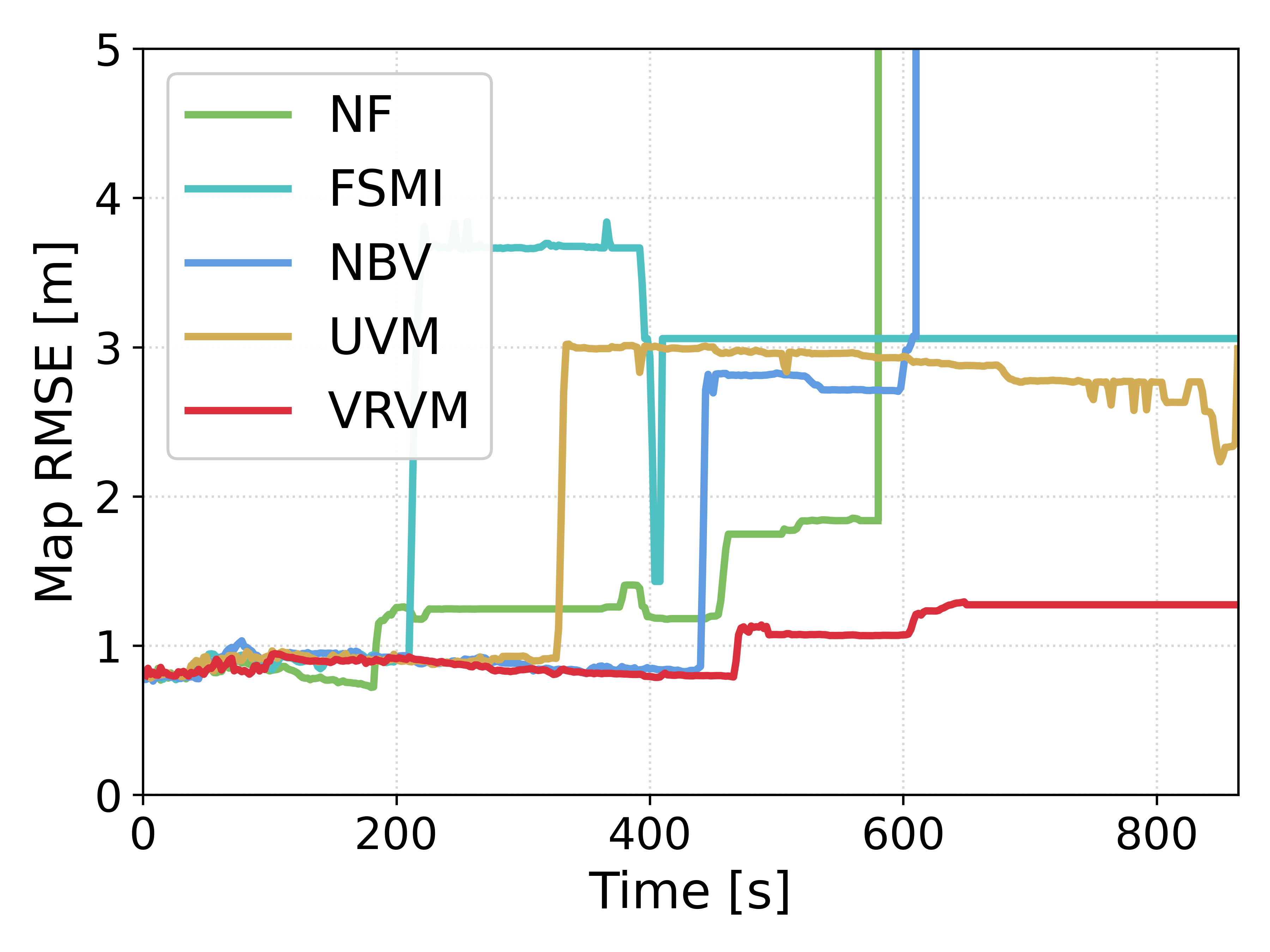}

  \end{minipage}%
}
  \hfill
  \subfloat[{Map Error - Dense\label{fig:map_error_1_3}}]{%
\begin{minipage}[t]{0.31\textwidth}
    \centering
    \includegraphics[width=\textwidth]{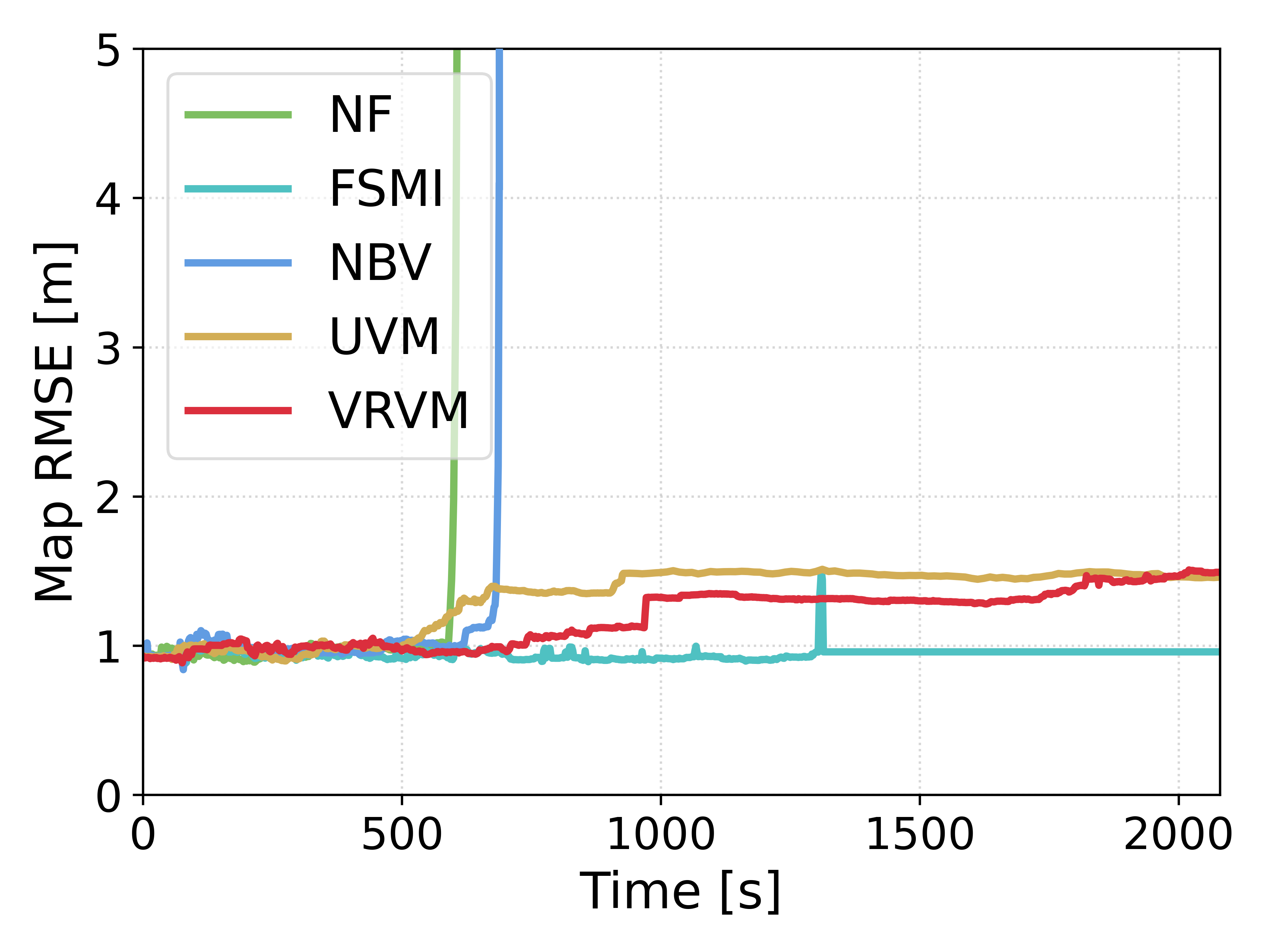}

  \end{minipage}%
}

  \subfloat[{Map Coverage - Sparse\label{fig:map_cov_1_1}}]{%
\begin{minipage}[t]{0.31\textwidth}
    \centering
    \includegraphics[width=\textwidth]{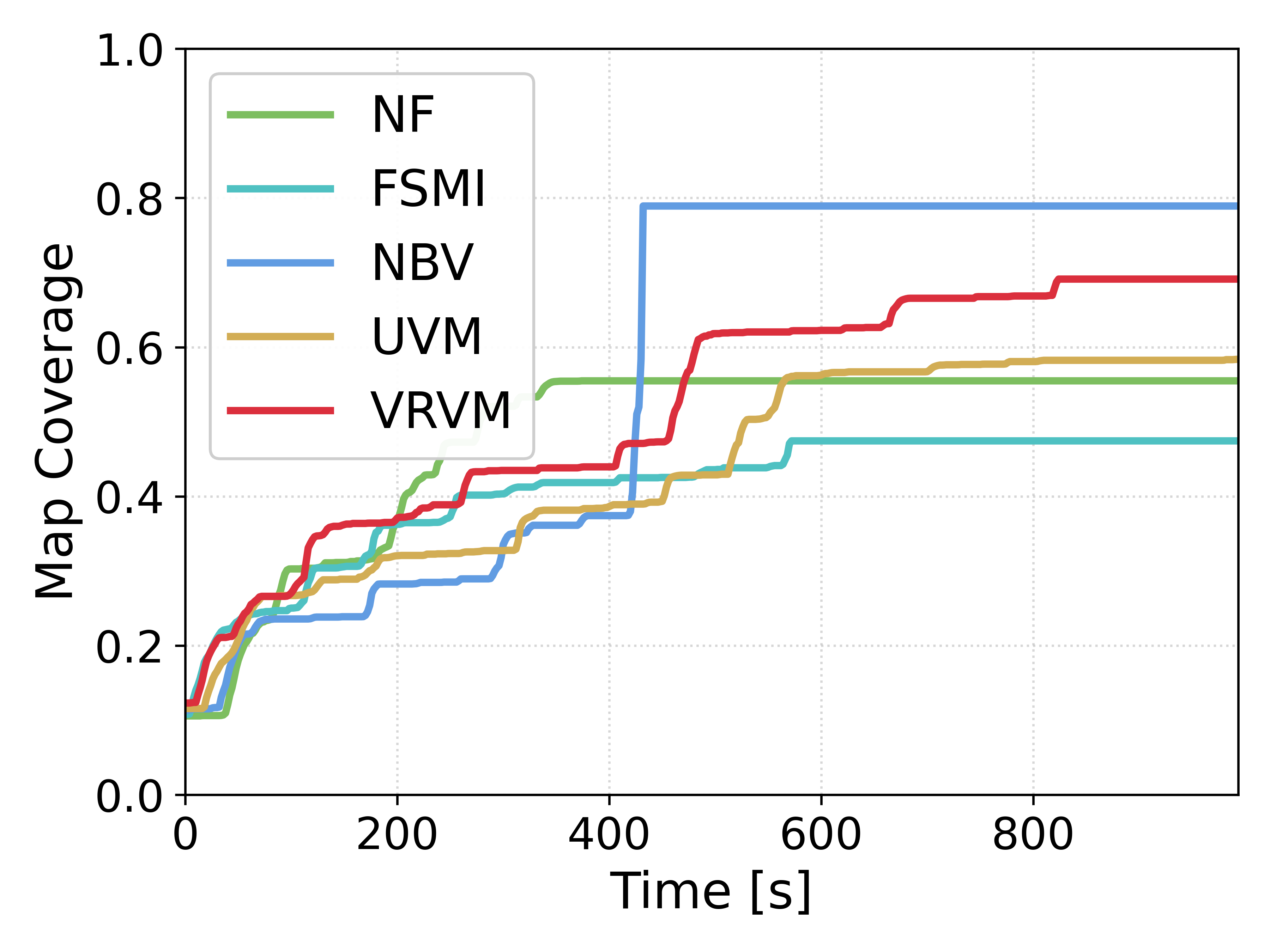}

  \end{minipage}%
}
  \hfill
  \subfloat[{Map Coverage - Moderate\label{fig:map_cov_1_2}}]{%
\begin{minipage}[t]{0.31\textwidth}
    \centering
    \includegraphics[width=\textwidth]{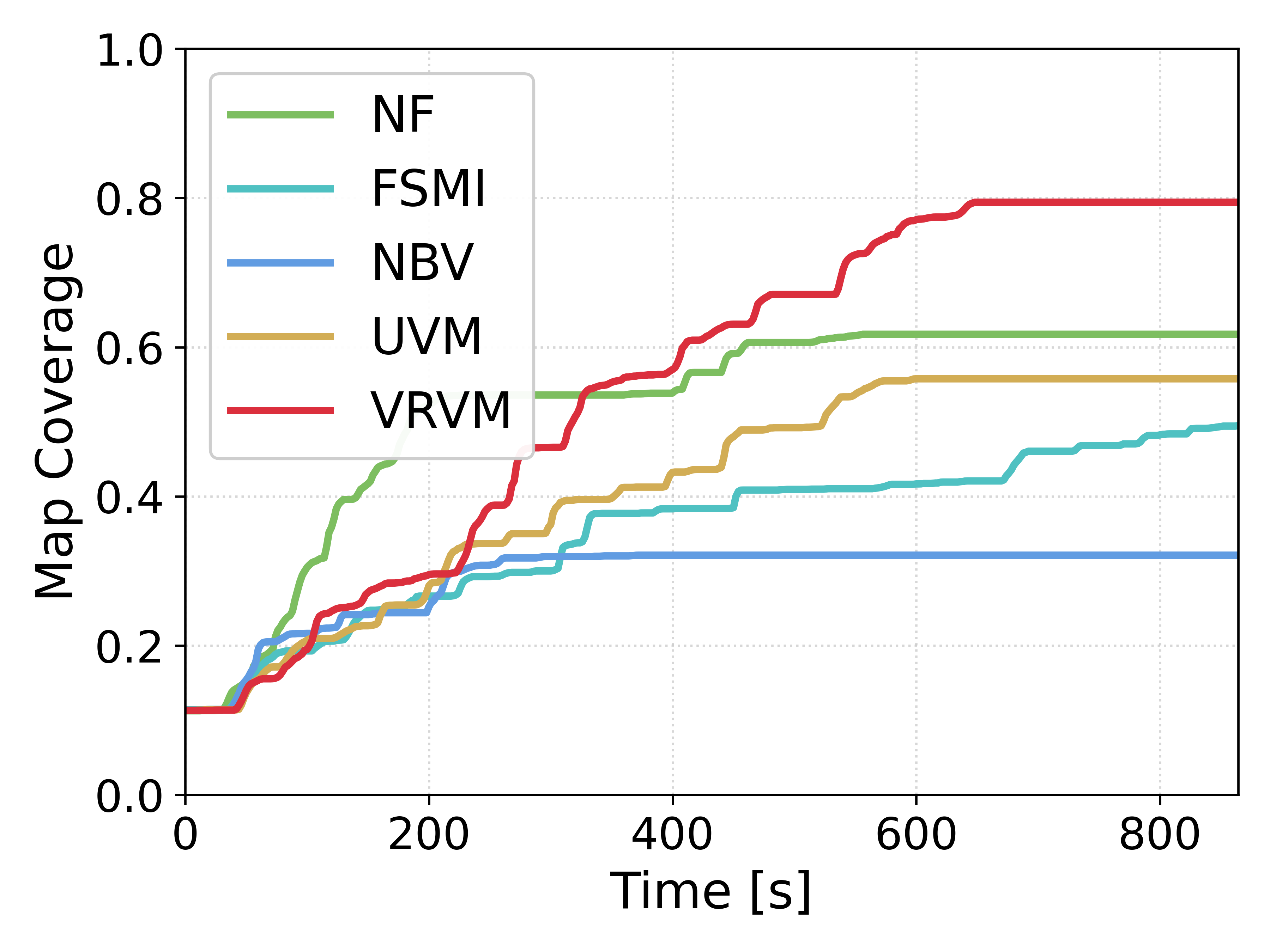}

  \end{minipage}%
}
  \hfill
  \subfloat[{Map Coverage - Dense\label{fig:map_cov_1_3}}]{%
\begin{minipage}[t]{0.31\textwidth}
    \centering
    \includegraphics[width=\textwidth]{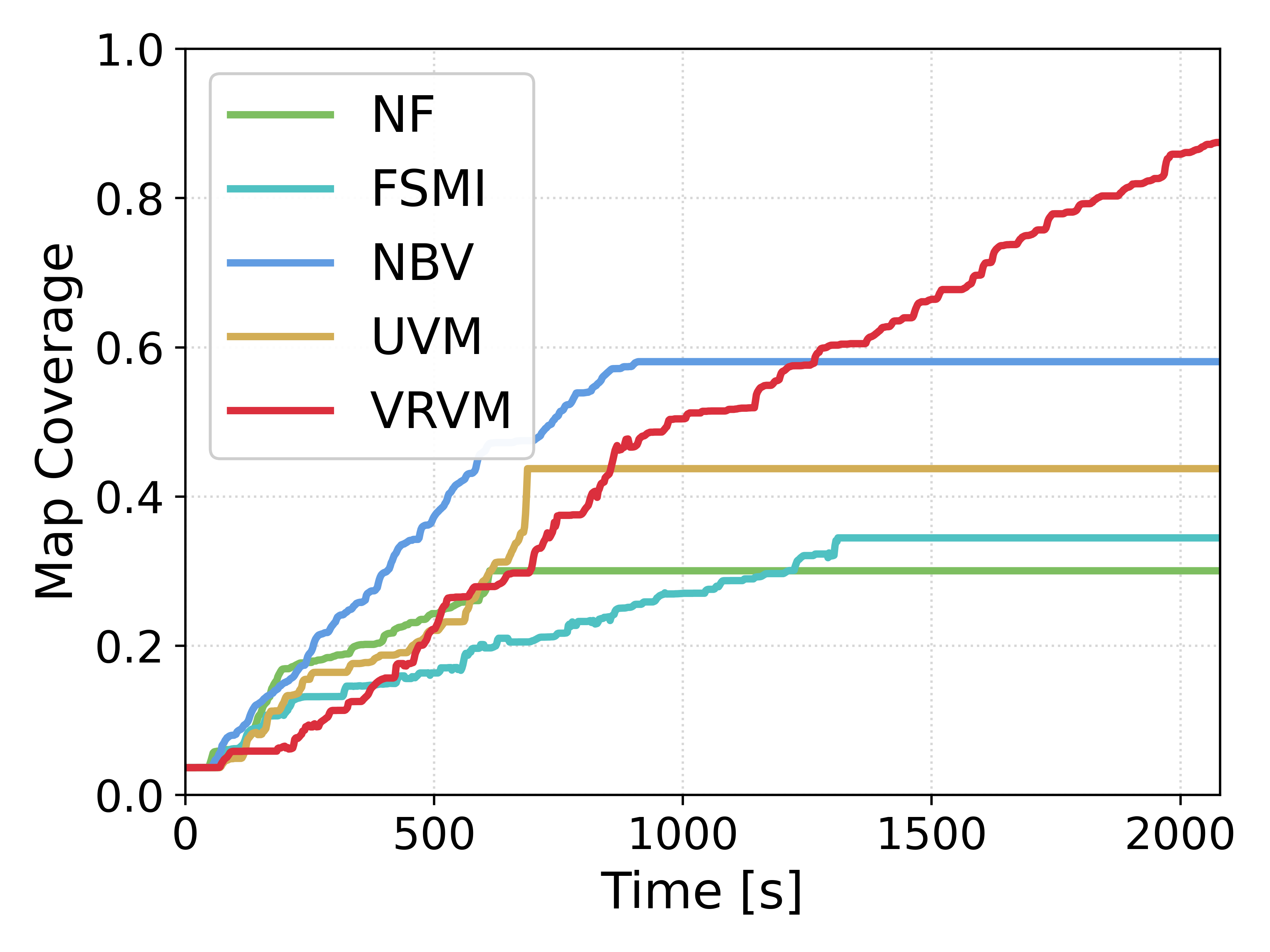}

  \end{minipage}%
}

  \caption{
Map error (RMSE) and union-normalised map coverage over time in Harbour Basin scenes with sparse, moderate, and dense structure using various methods. Each subplot compares Frontier, FSMI, NBV, the Uniform Virtual Map (UVM), and the proposed VRVM.
}

  \label{fig:mapping_error_coverage_all}

\end{figure*}
\subsubsection{Benchmark Utilities}
As mentioned in Sec.~\ref{sec:exp_setup}, we evaluate five exploration strategies
in the Docker Basin scene under three clutter levels: sparse, moderate, and dense. 
All methods use the same SLAM back-end and collision-aware navigation stack.
Let \(g\) denote a candidate goal position, \(\pi(g)\) the collision-free path produced by planning algorithm on an inflated traversable grid, we define the utility for all methods as follows:

\emph{NF}~\citep{ijrr2013efficient} enumerates reachable frontier cells and selects the one with minimum travel cost:
\begin{align}
g^\ast = \arg\max_{g}U_{\text{NF}}(g),\\
U_{\text{NF}}(g) = -L(g).
\end{align}

\emph{NBV}~\citep{ral2019NBV} follows an ``information–cost'' trade-off. In our implementation, the information term is approximated by \(G(g)\), defined as the number of unique unknown cells expected to become observable from a candidate viewpoint via ray-casting in the sensor field-of-view. To reduce redundancy, an overlap penalty is applied, including a minimum goal distance $L(g)$ and a short recent-goal penalty $\psi(g)$ to discourage repeatedly viewing already observed space.
\begin{align}
g^\ast &= \arg\max_g U_{\text{NBV}}(g),\\
U_{\text{NBV}}(g) &= G(g) - \alpha\,L(g) - \beta\,\Delta\psi(g).
\end{align}

\emph{FSMI}~\citep{zhang2020fsmi} replaces \(G(g)\) with $\text{MI}(g)$, an explicit mutual-information style score computed as expected entropy reduction accumulated along ray casts: 
\begin{align}
g^\ast &= \arg\max_g U_{\text{MI}}(g),\\
U_{\text{MI}}(g) &= \text{MI}(g) - \lambda\,L(g) - \beta\,\Delta\psi(g).
\end{align}

\emph{UVM}~\citep{wang2022virtual} and \emph{VRVM} use the same virtual-map-based utility defined in Sec.~\ref{sec:vrvm}. 
The only difference is the area weighting: UVM is uniform (no area weight), whereas VRVM introduces variable-resolution area weights.

\subsubsection{Analysis}
Qualitative comparisons of the final exploration result over the occupancy grid (left) and the corresponding SLAM trajectory and map estimate (right) are shown in Fig.\ref{fig:benchmark_sparse}, \ref{fig:benchmark_moderate} and \ref{fig:benchmark_dense}.
Runs that terminate due to loss of reliable SLAM state estimation are marked as \emph{Localisation Failure}.

In the sparse (Fig.~\ref{fig:benchmark_sparse}) and moderate (Fig.~\ref{fig:benchmark_moderate})  configuration, the two geometry-driven baselines, NF and NBV, are the most vulnerable to localisation failure.
In both cases, the resulting maps show clear signs of inconsistency around the failure region. In contrast, FSMI, UVM, and VRVM 
produce coherent reconstructions, indicating that their goal selection better maintains estimation stability while still expanding exploration.
The dense Docker Basin (Fig.~\ref{fig:benchmark_dense}) represents the most challenging configuration due to the presence of closely clustered obstacles.
NF and NBV fail at an early stage of exploration, while FSMI and VRVM successfully explore the majority of the environment. 
UVM remains stable but is unable to identify further exploration paths relatively early in the process, suggesting that removing area weighting can lead to earlier saturation of exploration progress under heavy clutter.

The qualitative outcomes above are consistent with the quantitative results in Fig.~\ref{fig:mapping_error_coverage_all}.
In all settings, NF and NBV exhibit abrupt Root Mean Square Error (RMSE) blow-ups associated with \emph{Localisation Failure}, whereas the other methods maintain bounded localisation error until exploration terminates.
The map coverage criteria shows how quickly each method expands and whether this expansion is sustained. 
Due to the specific characteristics of the environment, map coverage is evaluated only over areas containing structures.
Under this criterion, VRVM achieves the best overall performance among the compared methods.

\begin{figure*}[!t]
    \centering

    \subfloat[{UVM host RSS\label{fig:uvm_host_rss}}]{%
\begin{minipage}[t]{0.24\textwidth}
        \centering
        \includegraphics[width=\linewidth]{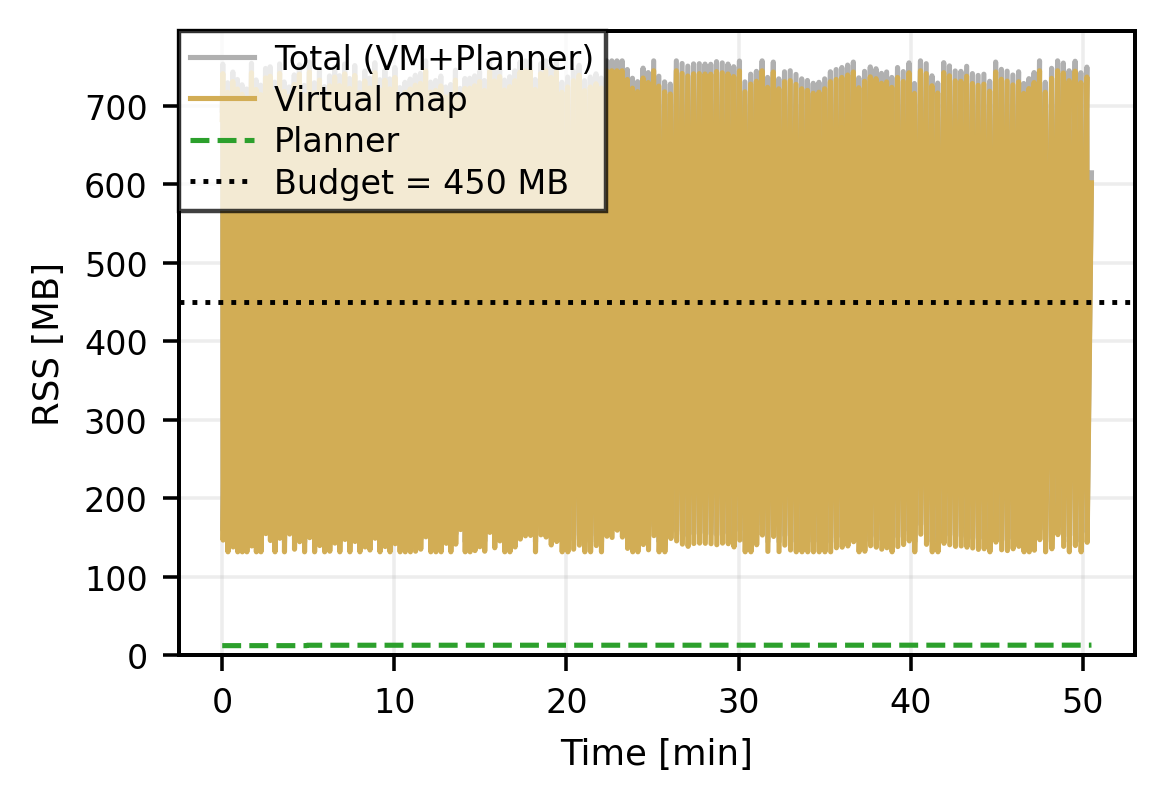}

    \end{minipage}%
}\hfill
    \subfloat[{UVM pi RSS\label{fig:uvm_pi_rss}}]{%
\begin{minipage}[t]{0.24\textwidth}
        \centering
        \includegraphics[width=\linewidth]{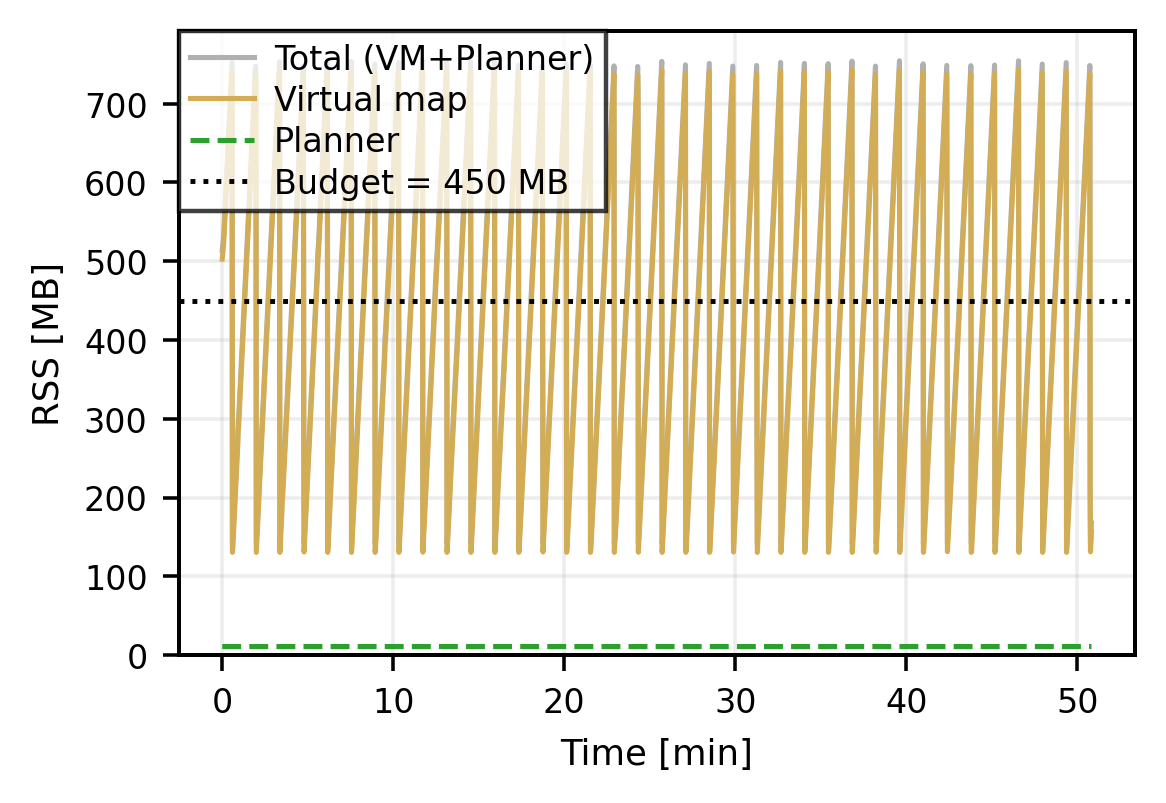}

    \end{minipage}%
}\hfill
    \subfloat[{VRVM host RSS\label{fig:vrvm_host_rss}}]{%
\begin{minipage}[t]{0.24\textwidth}
        \centering
        \includegraphics[width=\linewidth]{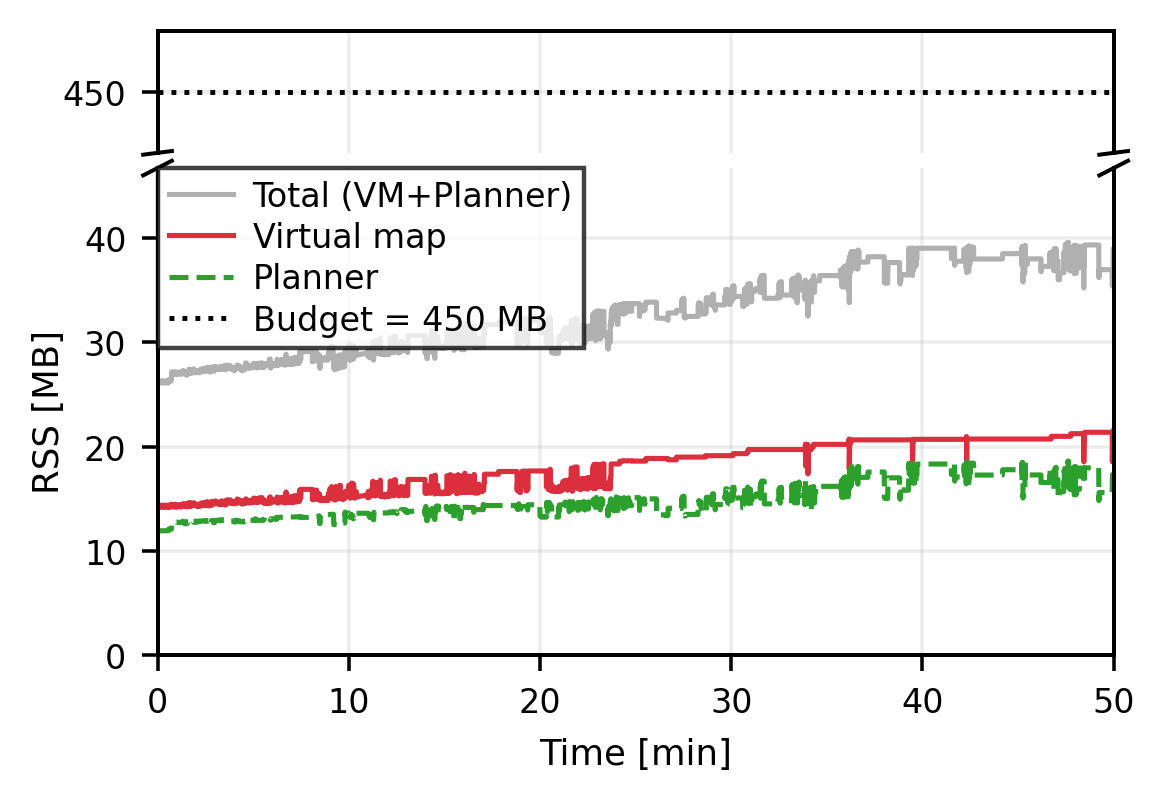}

    \end{minipage}%
}\hfill
    \subfloat[{VRVM pi RSS\label{fig:vrvm_pi_rss}}]{%
\begin{minipage}[t]{0.24\textwidth}
        \centering
        \includegraphics[width=\linewidth]{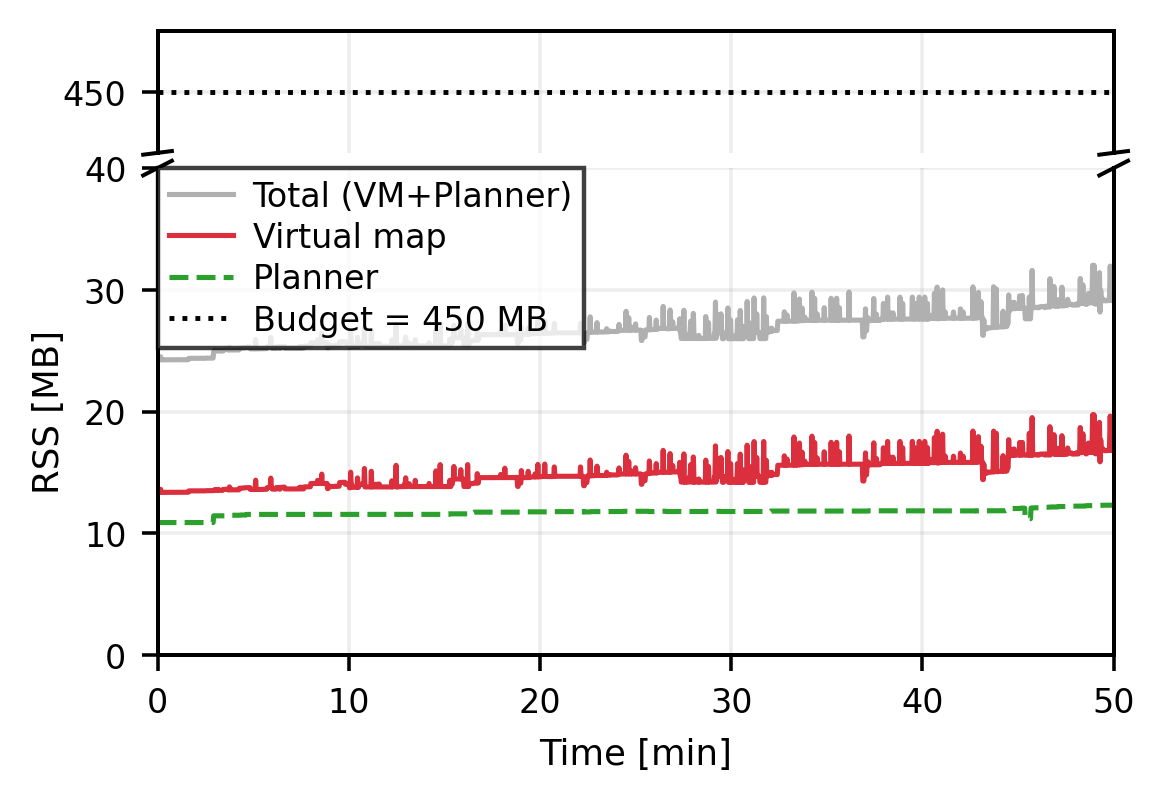}

    \end{minipage}%
}

    \vspace{2mm}

    \subfloat[{UVM host step\label{fig:uvm_host_step}}]{%
\begin{minipage}[t]{0.24\textwidth}
        \centering
        \includegraphics[width=\linewidth]{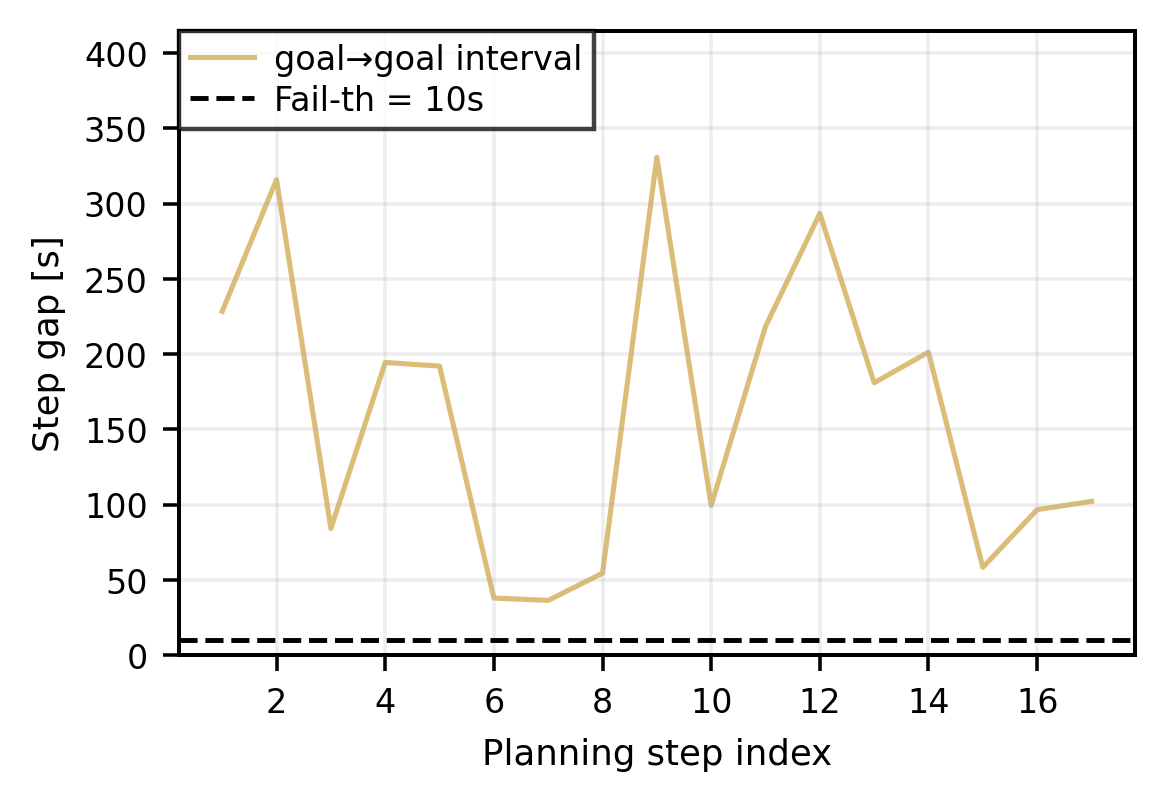}

    \end{minipage}%
}\hfill
    \subfloat[{UVM pi step\label{fig:uvm_pi_step}}]{%
\begin{minipage}[t]{0.24\textwidth}
        \centering
        \includegraphics[width=\linewidth]{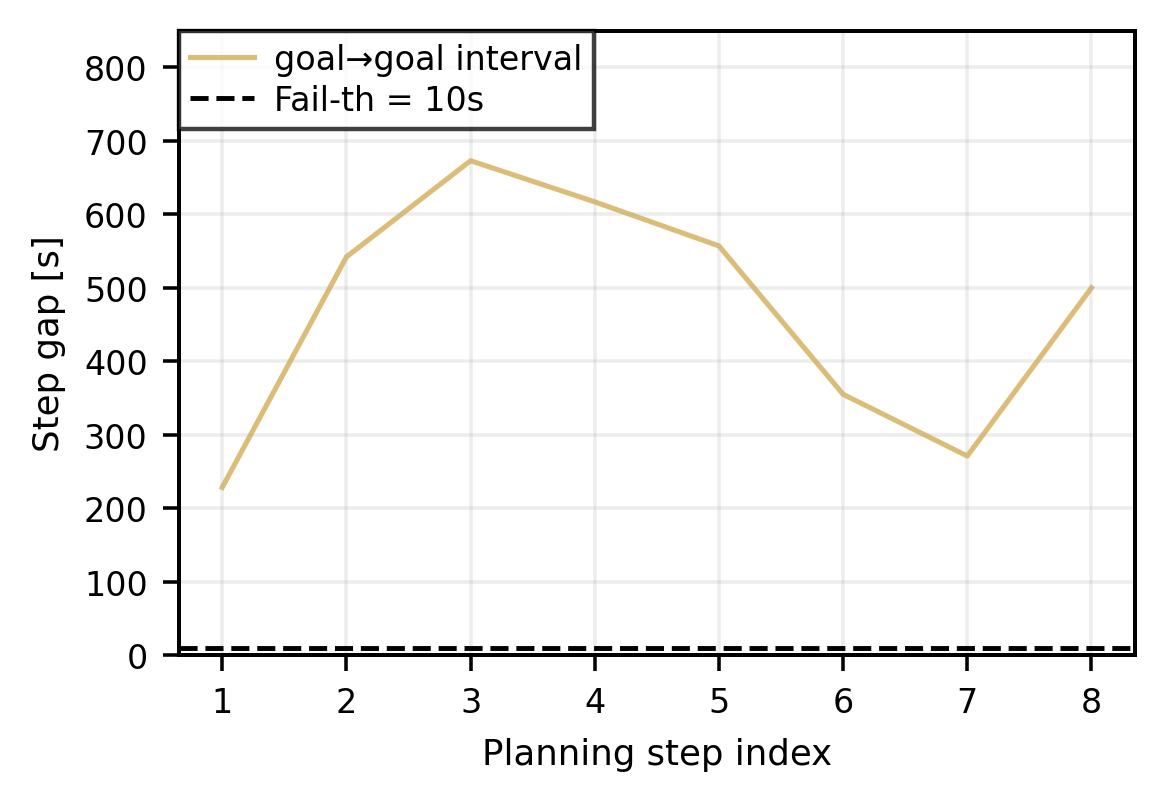}

    \end{minipage}%
}\hfill
    \subfloat[{VRVM host step\label{fig:vrvm_host_step}}]{%
\begin{minipage}[t]{0.24\textwidth}
        \centering
        \includegraphics[width=\linewidth]{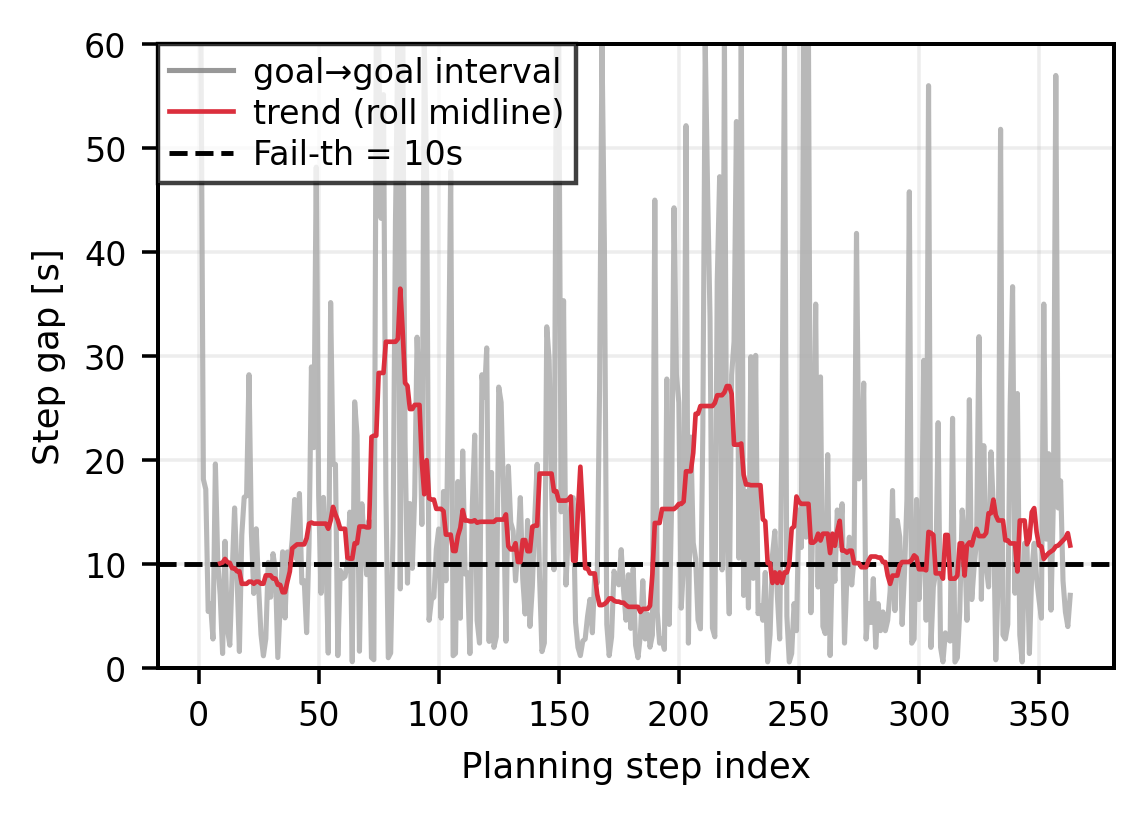}

    \end{minipage}%
}\hfill
    \subfloat[{VRVM pi step\label{fig:vrvm_pi_step}}]{%
\begin{minipage}[t]{0.24\textwidth}
        \centering
        \includegraphics[width=\linewidth]{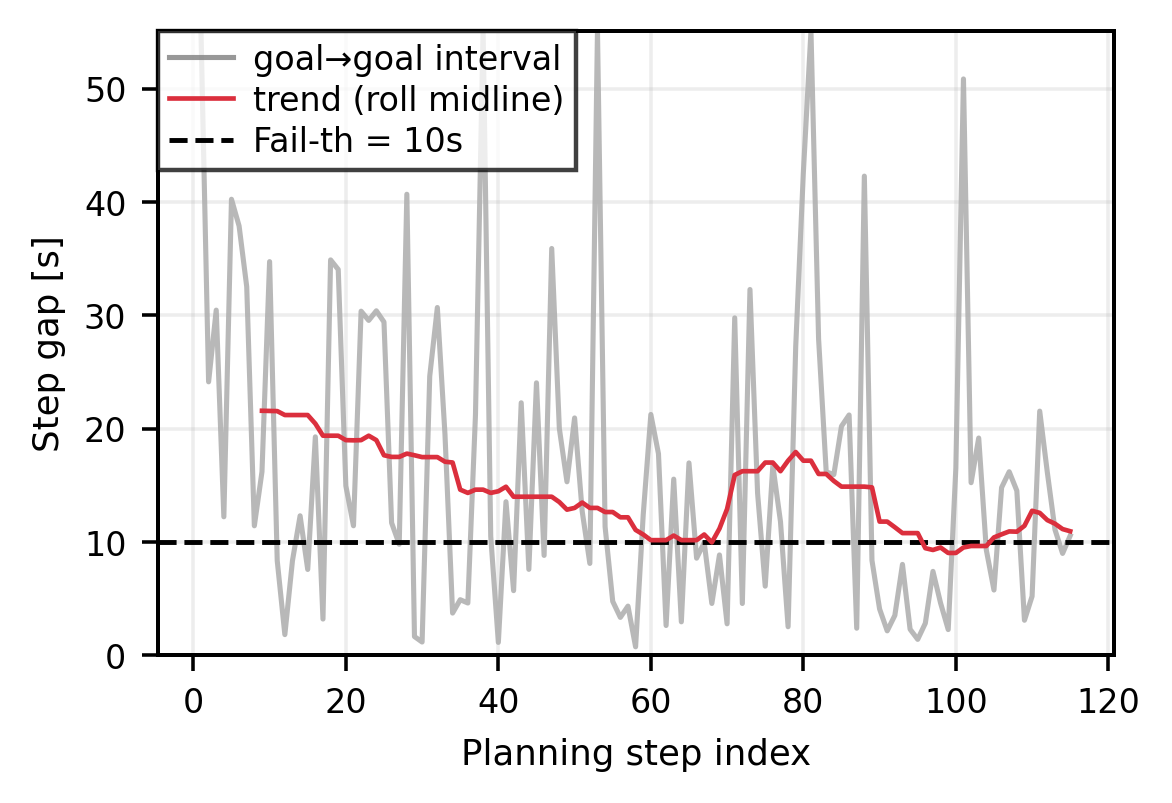}

    \end{minipage}%
}

    \caption{Performance comparison of UVM and VRVM in the port-area scene on desktop host machine and Raspberry Pi platforms. The top row presents Resident Set Size (RSS) memory usage relative to a 450 MB budget, while the bottom row displays planning step gap intervals against a 10 s failure threshold.}
    \label{fig:pi_and_host_rss_step}
\end{figure*}

\begin{figure*}[!t]
  \centering

  \subfloat[{\label{fig:peak-rss-run1}}]{%
\begin{minipage}[b]{0.384\textwidth}
    \centering
    \includegraphics[width=\linewidth]{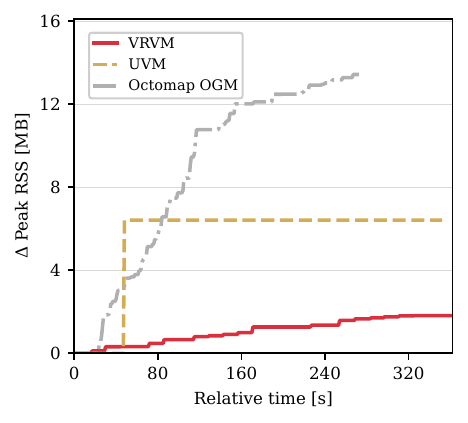}
    
  \end{minipage}%
}\hspace{0.032\textwidth}
\subfloat[{\label{fig:peak-rss-run2}}]{%
\begin{minipage}[b]{0.384\textwidth}
    \centering
    \includegraphics[width=\linewidth]{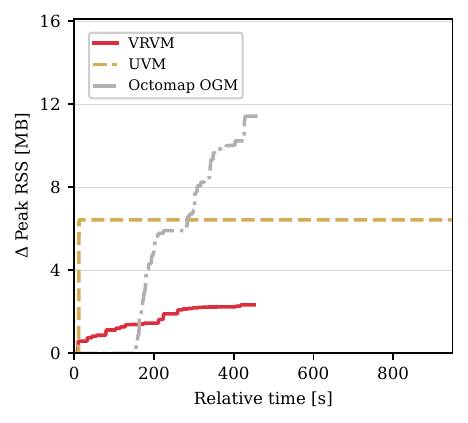}
    
  \end{minipage}%
}\par\medskip
\subfloat[{\label{fig:peak-rss-run3}}]{%
\begin{minipage}[b]{0.384\textwidth}
    \centering
    \includegraphics[width=\linewidth]{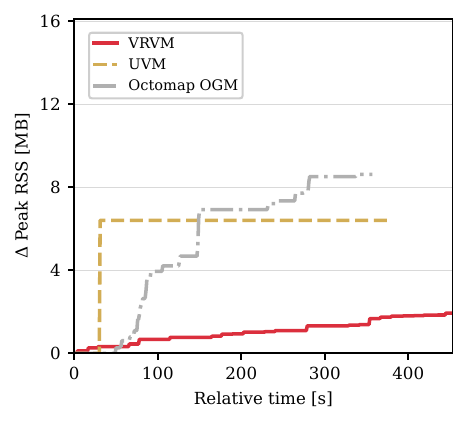}
    
  \end{minipage}%
}\hspace{0.032\textwidth}
\subfloat[{\label{fig:peak-rss-run4}}]{%
\begin{minipage}[b]{0.384\textwidth}
    \centering
    \includegraphics[width=\linewidth]{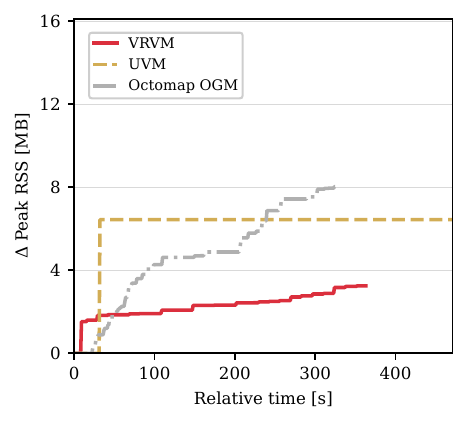}
    
  \end{minipage}%
}\par
\caption{Change in peak resident set size (RSS) over time for VRVM, the Uniform Virtual Map, and an OctoMap-based occupancy grid map across four independent runs in the Harbour Basin scene. }
  \label{fig:peak-rss-2x2}
\end{figure*}

\begin{figure*}[!t]
    \centering
    \subfloat[{\label{fig:fig9_1}}]{%
\begin{minipage}[b]{0.384\textwidth}
        \centering
        \includegraphics[width=\linewidth]{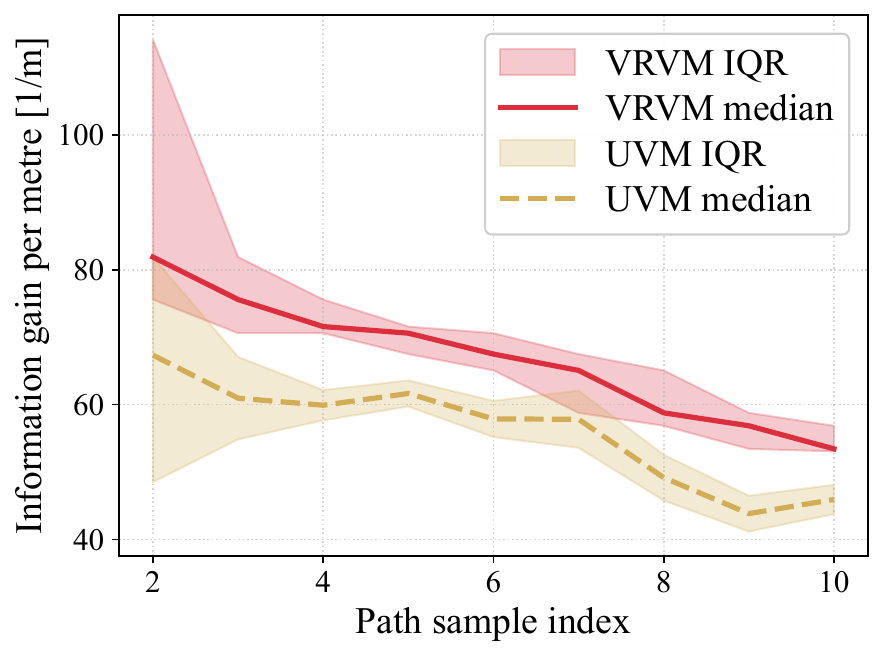}
        
    \end{minipage}%
}
    \hspace{0.032\textwidth}
    \subfloat[{\label{fig:fig9_2}}]{%
\begin{minipage}[b]{0.384\textwidth}
        \centering
        \includegraphics[width=\linewidth]{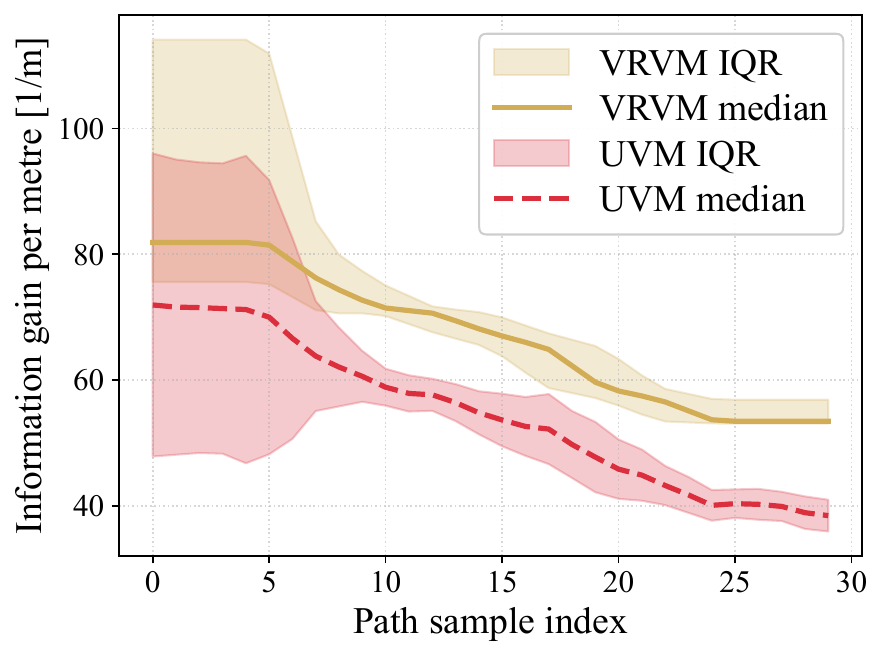}
        
    \end{minipage}%
}
    \caption{Information gain per metre along sampled paths in the Square basin Scene (left) and the Harbour Basin scene (right).} 
    \label{fig:fig9}
\end{figure*}

\subsection{Compute Scaling}
We evaluate compute scaling by comparing UVM and VRVM in the full $1000\,\mathrm{m}\times1000\,\mathrm{m}$ port-area scene.
To cover the entire area, both UVM and VRVM use the same virtual-map array size of $1200\,\mathrm{m}\times1200\,\mathrm{m}$.
All runs use the same mapping and localisation backbone and collision-aware navigation stack, the only difference is the virtual-map resolution policy.

\begin{figure*}[!t]
    \centerline{\includegraphics[width=.85\linewidth]{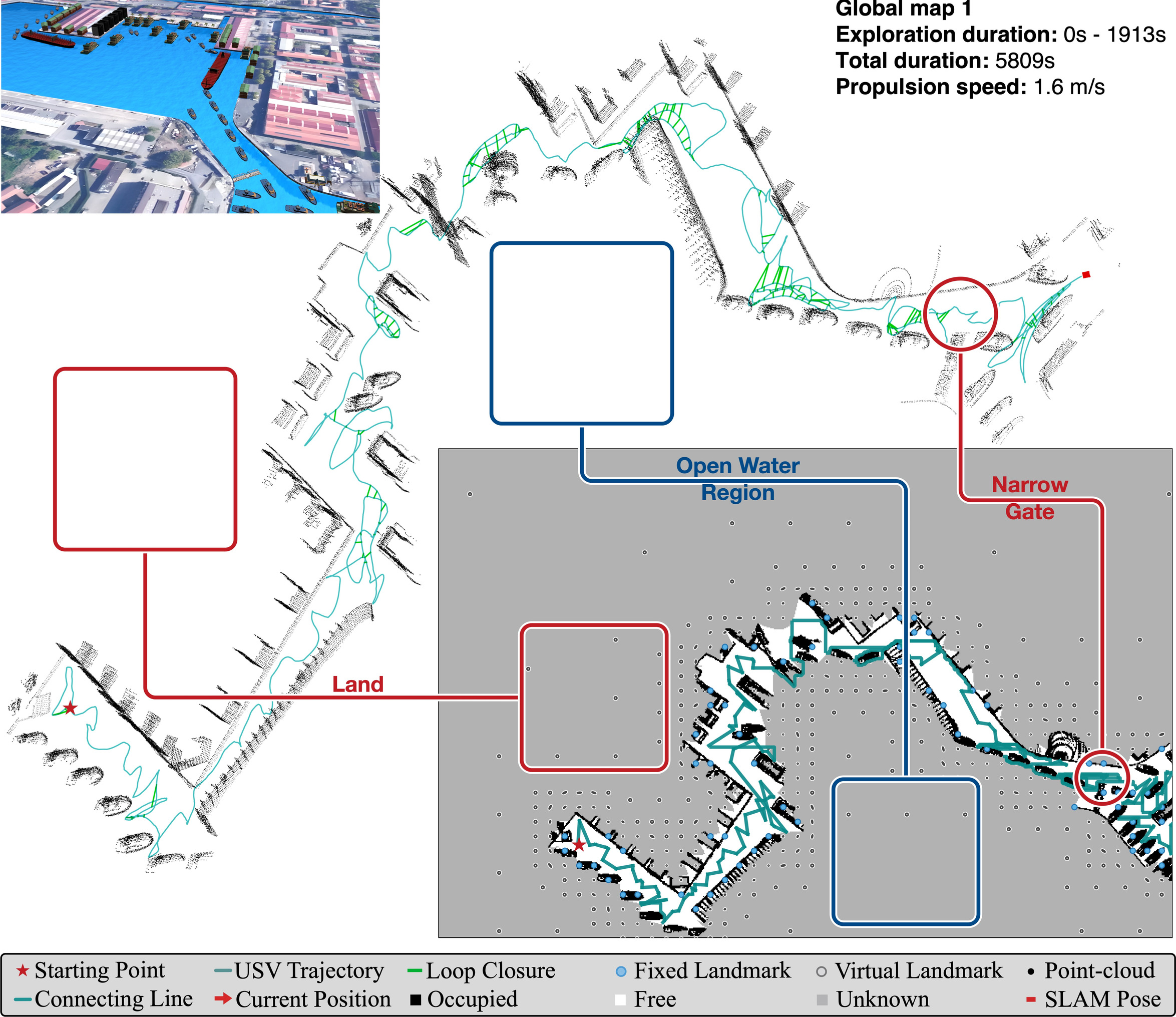}}
    \caption{Global exploration trajectory within the 1000 m $\times$ 1000 m environment (Part I). The USV initiates the survey from the starting point and traverses the open water region.}
    \label{fig:fullRegion1} 
\end{figure*}

\begin{figure*}[!t]
    \centerline{\includegraphics[width=.85\linewidth]{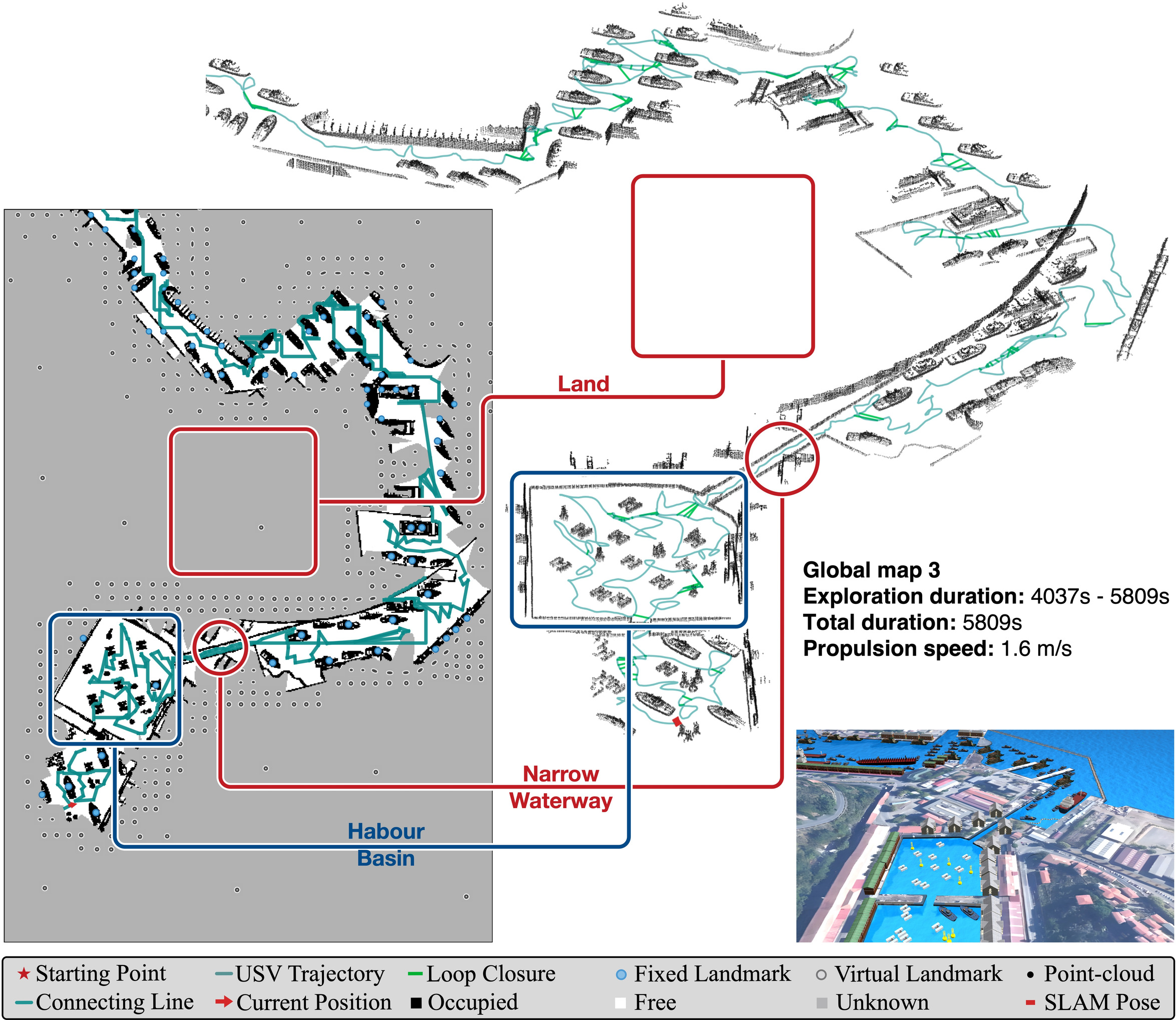}}
    \caption{Global exploration trajectory within the 1000 m $\times$ 1000 m environment (Part III). The mission concludes in the Harbour Basin after the USV navigates through a narrow waterway.}
    \label{fig:fullRegion3} 
\end{figure*}

\begin{figure*}[!t]
    \centerline{\includegraphics[width=.85\linewidth]{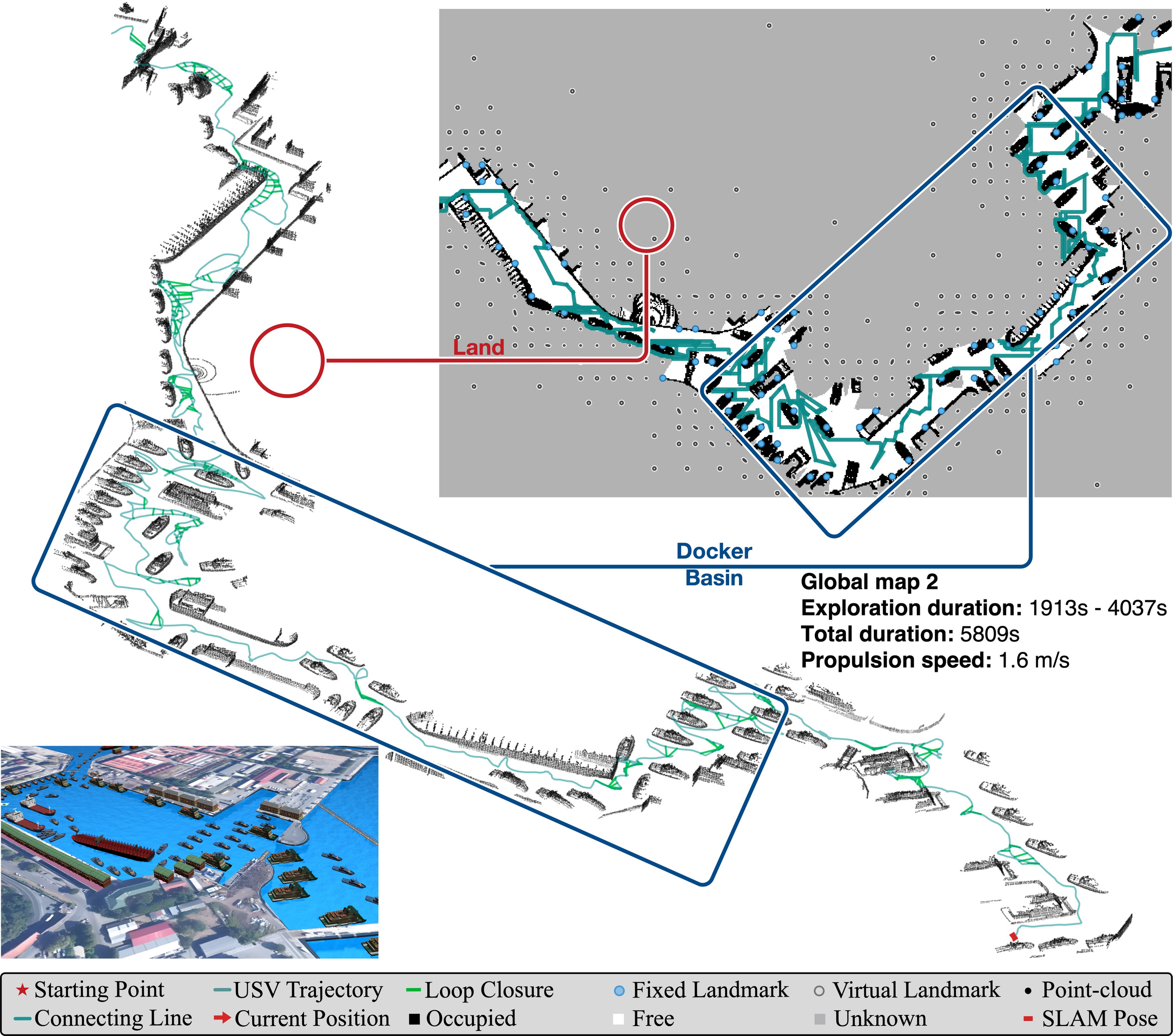}}
    \caption{Global exploration trajectory within the 1000 m $\times$ 1000 m environment (Part II). The path extends into the Docker Basin, where the USV maps the detailed boundaries and docked vessels.}
    \label{fig:fullRegion2} 
\end{figure*}

The reconstruction results are illustrated in Fig.~\ref{fig:fullRegion1}, \ref{fig:fullRegion2} and \ref{fig:fullRegion3}.
Only VRVM completes full-scene exploration on the Raspberry Pi, whereas UVM fails and exhibits resource and timing pathologies (Fig.~\ref{fig:pi_and_host_rss_step}).
On the Raspberry Pi, the acceptable RSS budget is set to $450 \mathrm{MB}$ to balance resource allocation between the exploration algorithm and the lower-level planning and control modules running concurrently on the system.
As shown in Fig.~\ref{fig:uvm_host_rss} and Fig.~\ref{fig:uvm_pi_rss}, UVM exhibits large oscillations in resident set size (RSS) on both the desktop host and the Raspberry Pi, whereas VRVM (Fig.~\ref{fig:vrvm_host_rss} and Fig.~\ref{fig:vrvm_pi_rss}) remains within a substantially narrower range on both platforms. This narrower range of VRVM than UVM provides a more predictable memory footprint, which is critical for preventing system failures on memory-constrained edge devices in long-endurance missions.

Fig.~\ref{fig:uvm_host_step}, \ref{fig:uvm_pi_step}, \ref{fig:vrvm_host_step} and \ref{fig:vrvm_pi_step} show the goal-to-goal planning interval traces for both methods. Using a failure threshold of $10\,\mathrm{s}$, UVM exceeds this limit by a wide margin on both platforms, whereas VRVM maintains goal-to-goal intervals near the $10\,\mathrm{s}$ scale on both the host and the Raspberry Pi. These results indicate the VRVM ability to maintain stable per-step planning intervals as the environmental representation scales.

Fig.~\ref{fig:peak-rss-2x2} compares the change in peak RSS over time for three mapping representations in the Harbour Basin scene: VRVM, UVM, and OctoMap. 
UVM quickly reaches a plateau at approximately $6$–$7\,\mathrm{MB}$ due to uniform allocation, while OctoMap exhibits the largest and continuously increasing memory growth as exploration progresses. 
VRVM maintains the lowest peak RSS and the slowest growth throughout the run. 
This indicates that as the explored area expands, VRVM constrains effective map growth, reducing memory load and supporting stable online operation.

Fig.~\ref{fig:fig9} shows information gain per metre along sampled paths in both the Square Basin scene and the Harbour Basin scene.
In both scenes, VRVM consistently achieves a higher median information gain per metre than UVM, while both exhibit a similar decreasing trend as less informative candidates are evaluated. Together with the RSS and timing results, there results indicate that VRVM reduces computational demand without sacrificing exploration efficiency, and improves information gathered per unit travel under resource constraints.

\begin{figure*}[!t]
    \centering
    \includegraphics[width=\linewidth]{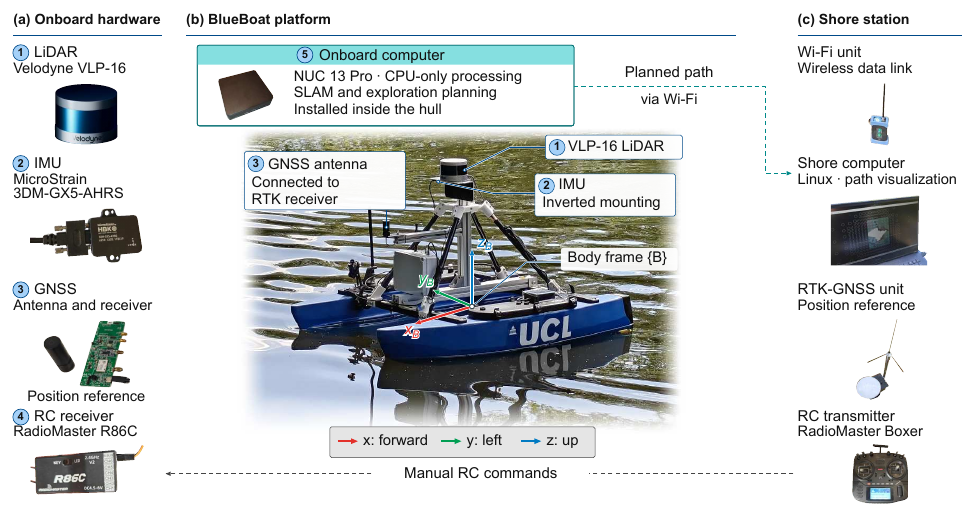}
    \caption{Field experimental setup: (a) onboard sensors and RC receiver;
    (b) BlueBoat platform and the NUC 13 Pro installed inside the hull;
    and (c) shore station. The onboard computer performs SLAM and
    exploration planning using CPU-only processing. Planned paths are
    transmitted to the shore computer via Wi-Fi, and the boat is
    steered manually via RC for safety reasons.}
    \label{fig:field_setup}
\end{figure*}

\subsection{Field Experiments}
\label{subsec:field_experiments}

We integrated sensing and onboard computing systems into the BlueBoat USV platform and conducted field experiments to evaluate VRVM without relying on GNSS. As shown in Fig.~\ref{fig:field_setup}, the platform was equipped with a VLP-16 LiDAR, a MicroStrain 3DM-GX5-AHRS IMU, and a GNSS/RTK receiver. A NUC 13 Pro installed inside the hull handled sensor data acquisition, SLAM, and exploration planning, with all computation performed on the CPU. FAST-LIO2~\citep{xu2022fastlio2} provided online pose estimates to the VRVM-based exploration planner. GNSS/RTK data were not used for online state estimation or planning; RTK-fixed positions were used to evaluate localization errors after the experiments. Planned paths were transmitted via Wi-Fi to a shore-based computer. Due to the presence of dense aquatic vegetation and fallen timber in the test area, an operator manually steered the boat via radio control (RC) for safety, using the planned paths as a reference. To examine localization performance with an alternative SLAM method, we also ran LIO-SAM offline on the same sensor recordings and compared its trajectory and localization errors with those obtained using FAST-LIO2.

\begin{figure*}[!t]
    \centering
    \includegraphics[width=\linewidth]{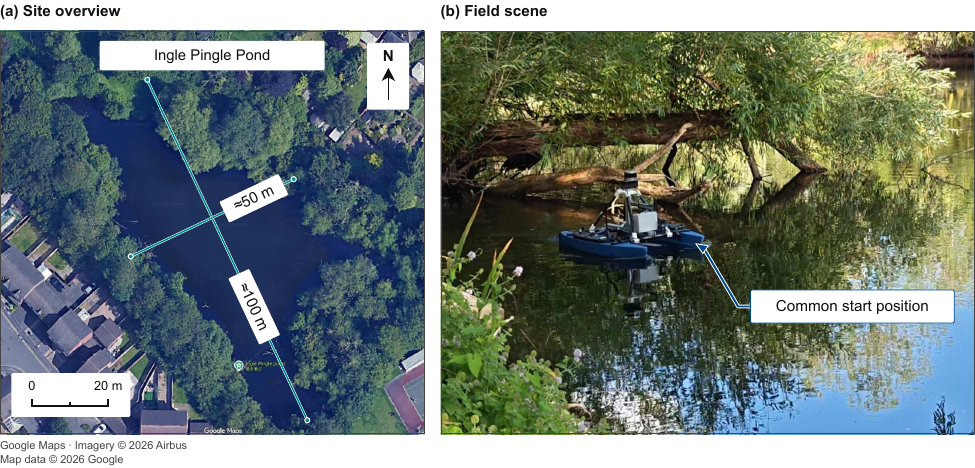}
    \caption{Field experiment site at Ingle Pingle Pond in
Loughborough, Leicestershire, UK.
(a) Satellite view from Google Maps, showing approximate
longitudinal and transverse water spans of 100~m and 50~m,
respectively.
(b) BlueBoat at the common start position.}
    \label{fig:field_site}
\end{figure*}

\begin{figure*}[!t]
    \centering
    \includegraphics[width=\linewidth]{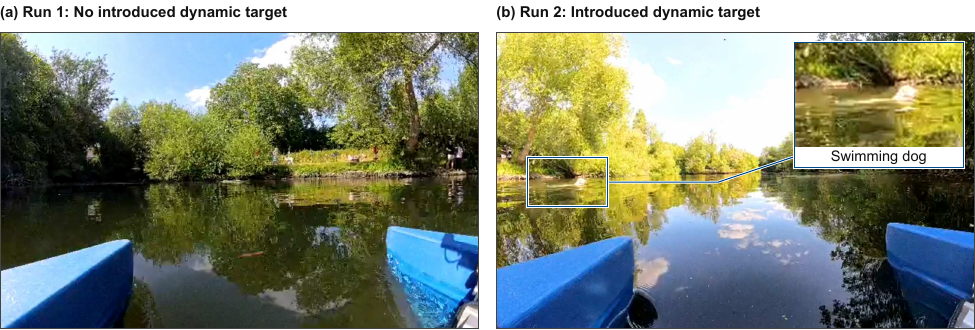}
    \caption{Representative onboard views from the two field trials.
    (a) Run 1, without an intentionally introduced dynamic target.
    (b) Run 2, with a swimming dog deliberately introduced as a
    dynamic target. The inset shows an enlarged view of the boxed region.}
    \label{fig:field_conditions}
\end{figure*}

The experiments took place at Ingle Pingle Pond, Loughborough, UK (Fig.~\ref{fig:field_site}), which was approximately $100\,\mathrm{m}$ long and $50\,\mathrm{m}$ wide.
The two VRVM-guided runs started from approximately the same physical location.
No dynamic objects were deliberately introduced in Run~1.
In Run~2, we threw a tennis ball for a dog to retrieve in shallow water near the shore (Fig.~\ref{fig:field_conditions}).
This run tested SLAM and VRVM in the presence of a dynamic object, while both assumed a static environment.

We compared exploration and mapping results over the first $2340\,\mathrm{s}$ ($39\,\mathrm{min}$) of each run.
We measured map growth using cumulative occupied area.
At each time point, we counted each grid cell once if its occupancy value had exceeded 50 in any map published up to that time, then multiplied the count by the cell area ($0.25\,\mathrm{m} \times 0.25\,\mathrm{m}$).
For localization evaluation, we used the same time interval within each run to align the FAST-LIO2 and LIO-SAM trajectories with the RTK-fixed reference.
Each trajectory was aligned separately by rotation and translation in the horizontal plane.
After this interval, we evaluated errors only at times when an RTK-fixed position and valid position estimates from both methods were available.
We calculated horizontal position RMSE for each method over $10\,\mathrm{s}$ intervals.

\begin{figure*}[!t]
    \centering
    \includegraphics[
        width=0.8\linewidth,
        height=0.656\textheight,
        keepaspectratio
    ]{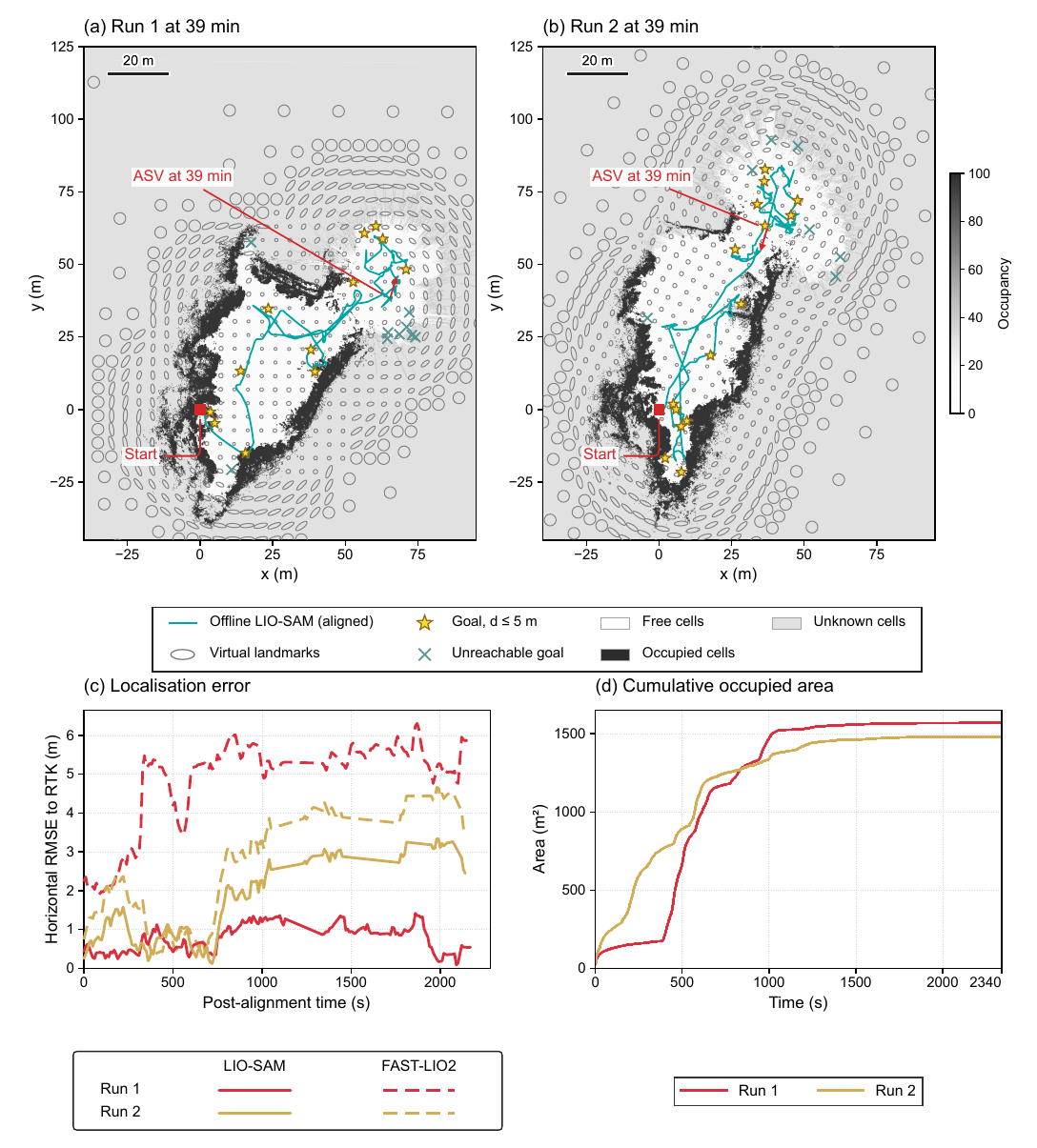}
\caption{Field experiment results.
(a,b) Occupancy grids, virtual landmarks, and aligned offline
LIO-SAM trajectories at 39~min for Runs~1 and~2, shown in
separate display frames. Stars mark goals within 5~m of
the displayed trajectory; crosses mark unreachable goals.
(c) Horizontal localisation RMSE relative to RTK-fixed
measurements, using post-alignment time.
(d) Cumulative occupied area in the published maps over
the first 39~min.}
    \label{fig:field_exploration_localisation}
\end{figure*}

We describe the accumulation of occupied cells in the published maps because it was difficult to directly determine how fully each run covered the real static environment.
Within the $39\,\mathrm{min}$ comparison window, Run~1 published 21 planning goals and Run~2 published 22.
Fig.~\ref{fig:field_exploration_localisation}(a,b) shows the goal locations and trajectories for both runs.
Both runs had goals near the start and in areas farther away, but the goals were selected in different orders.
Run~2 initially explored outward, then selected several consecutive goals closer to the starting point before choosing more distant goals again. This temporary shift was attributed to interference from a dynamic object, which affected the covariance-based uncertainty estimates used for goal selection.
In addition to environmental factors, the stochastic nature of RRT-based candidate-trajectory generation contributes to the differences in goal-selection order and cumulative occupied area between the two runs.
Fig.~\ref{fig:field_exploration_localisation}(d) shows that map growth also differed between the two runs.
At $600\,\mathrm{s}$, the cumulative occupied area in Run~2 was $1120\,\mathrm{m}^2$, higher than the $970\,\mathrm{m}^2$ in Run~1.
After this, the area in Run~1 continued to grow and became larger than that in Run~2.
By $2340\,\mathrm{s}$, Run~1 reached $1573\,\mathrm{m}^2$ and Run~2 reached $1481\,\mathrm{m}^2$, with growth gradually slowing in both runs toward the end.

To compare the localization performance of different state estimation methods on field data, we evaluated the localization errors of smoothing-based LIO-SAM and filtering-based FAST-LIO2 in the two runs without loop closure.
As shown in Fig.~\ref{fig:field_exploration_localisation}(c), over the valid evaluation samples after alignment, offline LIO-SAM had an overall horizontal position RMSE of $0.83\,\mathrm{m}$ in Run~1 and $1.91\,\mathrm{m}$ in Run~2.
For FAST-LIO2, the values were $4.94\,\mathrm{m}$ in Run~1 and $2.75\,\mathrm{m}$ in Run~2.
In Run~1, the LIO-SAM error remained low, while the FAST-LIO2 error increased early in the run.
In Run~2, both methods had higher errors toward the end than at the beginning. 

In Run~2, returns from moving objects may have affected FAST-LIO2's pose estimates during scan-to-map registration.
If these returns were accepted and treated as part of the static environment, they could cause errors in the estimated pose.
Points added to the map using this pose could then be placed at incorrect positions.
If these points were used as a reference when matching later scans to the map, they could also affect future pose estimates.
Therefore, even if later scans matched the map well, the estimated position could still be inaccurate.

\section{Discussion}
\label{sec:discussion} 

\subsection{Scope and applicability of VRVM}
\label{subsec:scope}

The primary contribution of VRVM lies in refining the uncertainty representation in the virtual-map. By coupling variable-resolution map refinement with visibility-limited updates, VRVM preserves the intended interaction between planning and SLAM-derived marginal covariances while shifting computational effort away from map bookkeeping toward trajectory-local updates.

Under identical SLAM, mapping, and navigation stacks, the experimental results indicate bounded memory growth and replanning latency across long-horizon runs, supporting the conclusion that variable resolution combined with visible-set updates improves computational sustainability under constrained onboard resources. In structurally uneven scenes, VRVM further maintains bounded mapping error over longer horizons than geometry-driven baselines, which exhibit earlier localisation degradation, yielding a qualified robustness advantage in the evaluated regime.

\subsection{Mechanisms and information-theoretic interpretation}
The behavioural and computational differences observed in Sec.~\ref{sec:experiments} follow directly from how VRVM modifies both the \emph{support} and the \emph{measure} of uncertainty evaluation. VRVM restricts updates to the sensor-induced visible set (Eq.~\ref{eq:visible_set}), applies monotone information-form updates (Eq.~\ref{eq:info_update}), and evaluates map uncertainty using an area-weighted D-optimality criterion (Eq.~\ref{eq:J_area_x_fixed}). Together, these design choices determine the dominant per-cycle workload and reshape the incentives induced by the map term in Eq.~\ref{eq:utility}.

From a computational perspective, visible-set updates and adaptive refinement confine repeated covariance operations to leaves intersecting the sensing footprint, shifting the dominant work from map dimensions to terms proportional to $|\mathcal{V}(\mathbf{x}_k)|$ and local splits. This explains the bounded memory growth and replanning latency observed under identical upstream SLAM and navigation stacks.

From a behavioural perspective, the D-optimality terms reward actions that produce measurable uncertainty contraction. In near-shore scenes with uneven geometric structure, extended traversals through feature-sparse open water provide limited contraction while increasing drift exposure. Geometry-driven baselines therefore tend to overcommit to low-observability regions, whereas VRVM assigns low utility to candidates whose visible sets are weakly informative, discouraging deep excursions without introducing explicit heuristics. The resulting robustness advantage in sparse and moderate scenes follows from this alignment between valuation and observability.

Area-weighted aggregation further removes split sensitivity by normalising contributions by physical area, ensuring that refinement does not artificially inflate map utility. This makes the objective locally invariant to discretisation changes and ties candidate evaluation to physical uncertainty reduction rather than representation artefacts.

Under this interpretation, VRVM should be viewed as a tractable D-optimality-driven active-exploration approximation within the virtual-map framework. By aligning uncertainty valuation with what is observable and aggregating it in a split-stable manner, VRVM shifts exploration toward a compute-bounded and observability-aware regime without requiring dense mutual-information evaluation.

\subsection{Limitations and future work}

While variable-resolution representation improves computational scalability, it introduces a resolution-dependent limitation. When the largest allowable cell size is classified as free, the corresponding region may be treated as sufficiently explored even though smaller-scale structure has not been resolved. This can suppress further refinement and reduce exploration pressure in areas that are nominally empty at coarse resolution but may contain unmodelled geometric detail.

In addition, the split and occupancy-locking mechanisms introduce sensitivity at the representation level. In dynamic or partially observed settings, transient misclassification can trigger premature refinement or persistent locking, distorting subsequent uncertainty valuation. Incorporating hysteresis in refinement decisions and time-decayed unlocking policies would mitigate such effects without altering the underlying formulation.

The field trials also exposed limitations in obstacle perception.
In some cases, physical contact revealed obstacles that the onboard perception system had not identified, preventing access to planned goals.
We will investigate more integrating sonar for underwater obstacle detection and marine radar for surface obstacle detection with the existing LiDAR system.

Another limitation observed during field testing lies with the planner. Random sampling in RRT may affect goal selection, and its effect on the field results has not been quantified.

Building on the above limitations, future work will extend the field evaluation to repeated experiments in representative water environments, including systematic assessment of wind, wave, and current effects and controlled tests involving dynamic obstacles and vessel traffic. Methodologically, robustness to transient occupancy misclassification will be improved by introducing hysteresis in refinement decisions and time-decayed locking mechanisms.

\section{CONCLUSIONS}
\label{sec:conclusions}

Motivated by long-horizon USV exploration in GNSS-degraded near-shore waters, this work addressed a practical limitation of uncertainty-aware planning: the inefficiency and instability that arise when uncertainty must be evaluated over large, geometrically uneven workspaces under constrained onboard computation. Such environments require balancing coverage expansion with estimator stability in the presence of feature-sparse open water and dense shoreline structure.

We introduced a variable-resolution virtual map (VRVM) framework that confines uncertainty updates to the sensor-visible set and refines representation adaptively, coupled with split-stable, area-weighted valuation within a receding-horizon planner. This design preserves compatibility with factor-graph SLAM while shifting computation from global bookkeeping to locally relevant updates.

Across experiments in simulated representative near-shore scenarios, VRVM demonstrates improved stability and computational sustainability relative to uniform-resolution and geometry-driven baselines. Field trials in a real-world pond further demonstrate VRVM’s ability to support sustained exploration planning without GNSS, with planned paths serving as references for operator-controlled navigation. These results support the conclusion that resolution-adaptive representation and valuation can move uncertainty-aware exploration toward a compute-bounded and failure-resistant operating regime in structurally uneven marine environments.

\section*{Acknowledgment}
This work was supported by the Engineering and Physical Sciences Research Council (EPSRC) under Grants EP/Y000862/1 and EP/X034909/1, and by The Royal Society Kan Tong Po Fellowship (KTP/R1/251117). The work was also  
in part supported by NSF grants 2144624.

The authors thank Professor Cunjia Liu of Loughborough University for providing access to the field-test site and hardware support for the field experiments.

\bibliographystyle{IEEEtranN}
\bibliography{bib}  

@InProceedings{bingham19toward,
  Title                    = {Toward Maritime Robotic Simulation in Gazebo},
  Author                   = {Brian Bingham and Carlos Aguero and Michael McCarrin and Joseph Klamo and Joshua Malia and Kevin Allen and Tyler Lum and Marshall Rawson and Rumman Waqar},
  Booktitle                = {Proceedings of MTS/IEEE OCEANS Conference},
  Year                     = {2019},
  Address                  = {Seattle, WA},
  Month                    = {October},
  pages={1--10}
}

@INPROCEEDINGS{shan2020liosam,
  author={Shan, Tixiao and Englot, Brendan and Meyers, Drew and Wang, Wei and Ratti, Carlo and Rus, Daniela},
  booktitle={2020 IEEE/RSJ International Conference on Intelligent Robots and Systems (IROS)}, 
  title={{LIO-SAM}: Tightly-coupled Lidar Inertial Odometry via Smoothing and Mapping}, 
  year={2020},
  volume={},
  number={},
  pages={5135-5142},
}

@article{hornung2013octomap,
  author  = {Hornung, Armin and Wurm, Kai M. and Bennewitz, Maren and Stachniss, Cyrill and Burgard, Wolfram},
  title   = {{OctoMap}: An Efficient Probabilistic 3D Mapping Framework Based on Octrees},
  journal = {Autonomous Robots},
  year    = {2013},
  volume={34},
  number={3},
  pages={189--206},
}

@INPROCEEDINGS{yamauchi1997frontier,
  author={Yamauchi, B.},
  booktitle={Proceedings 1997 IEEE International Symposium on Computational Intelligence in Robotics and Automation CIRA'97. 'Towards New Computational Principles for Robotics and Automation'}, 
  title={A frontier-based approach for autonomous exploration}, 
  year={1997},
  volume={},
  number={},
  pages={146-151},
  doi={10.1109/CIRA.1997.613851}}

@INPROCEEDINGS{bircher2016rhnbv,
  author={Bircher, Andreas and Kamel, Mina and Alexis, Kostas and Oleynikova, Helen and Siegwart, Roland},
  booktitle={2016 IEEE International Conference on Robotics and Automation (ICRA)}, 
  title={Receding Horizon "Next-Best-View" Planner for 3D Exploration}, 
  year={2016},
  volume={},
  number={},
  pages={1462-1468},
}

@INPROCEEDINGS{charrow2015csqmi,
  author={Charrow, Benjamin and Liu, Sikang and Kumar, Vijay and Michael, Nathan},
  booktitle={2015 IEEE International Conference on Robotics and Automation (ICRA)}, 
  title={Information-theoretic mapping using Cauchy-Schwarz Quadratic Mutual Information}, 
  year={2015},
  volume={},
  number={},
  pages={4791-4798},
}

@INPROCEEDINGS{zhang2020fsmi,
  author={Zhang, Zhengdong and Henderson, Trevor and Sze, Vivienne and Karaman, Sertac},
  booktitle={2019 International Conference on Robotics and Automation (ICRA)}, 
  title={{FSMI}: Fast Computation of Shannon Mutual Information for Information-Theoretic Mapping}, 
  year={2019},
  volume={},
  number={},
  pages={6912-6918},
}

@inproceedings{chen2019autonomous,
  title={Autonomous exploration under uncertainty via graph convolutional networks},
  author={Chen, Fanfei and Wang, Jinkun and Shan, Tixiao and Englot, Brendan},
  booktitle={The International Symposium of Robotics Research},
  pages={676--691},
  year={2019},
  organization={Springer}
}

@inproceedings{chaplot2020anslam,
  author    = {Chaplot, Devendra Singh and Gandhi, Dhiraj and Gupta, Saurabh and Gupta, Abhinav},
  title     = {Learning to Explore Using Active Neural {SLAM}},
  booktitle = {Proc. Int. Conf. Learning Representations (ICLR)},
  year      = {2020}
}

@INPROCEEDINGS{wang2019virtualmap,
  author={Wang, Jinkun and Shan, Tixiao and Englot, Brendan},
  booktitle={2019 IEEE/RSJ International Conference on Intelligent Robots and Systems (IROS)}, 
  title={Virtual Maps for Autonomous Exploration with Pose {SLAM}}, 
  year={2019},
  volume={},
  number={},
  pages={4899-4906},
}

@INPROCEEDINGS{colladogonzalez2024vmauv,
  author={Collado-Gonzalez, Ivana and McConnell, John and Wang, Jinkun and Szenher, Paul and Englot, Brendan},
  booktitle={2024 IEEE International Conference on Robotics and Automation (ICRA)}, 
  title={Real-Time Planning Under Uncertainty for AUVs Using Virtual Maps}, 
  year={2024},
  volume={},
  number={},
  pages={8334-8340},
}

@Article{saldarriaga2025vrx,
AUTHOR = {Saldarriaga-Mesa, Brayan and Montesdeoca, Julio and Báez, Dennys and Roberti, Flavio and Toibero, Juan Marcos},
TITLE = {Open-Access Simulation Platform and Motion Control Design for a Surface Robotic Vehicle in the {VRX} Environment},
JOURNAL = {Robotics},
VOLUME = {14},
YEAR = {2025},
NUMBER = {10},
ARTICLE-NUMBER = {147},
ISSN = {2218-6581},
}

@article{engstrom2022lidarslam,
title = {A lidar-only SLAM algorithm for marine vessels and autonomous surface vehicles},
journal = {IFAC-PapersOnLine},
volume = {55},
number = {31},
pages = {229-234},
year = {2022},
note = {14th IFAC Conference on Control Applications in Marine Systems, Robotics, and Vehicles CAMS 2022},
issn = {2405-8963},
author = {Artur Engström and Domenic Geiseler and Krister Blanch and Ola Benderius and Iván García Daza},
}

@article{sawada2023shipslam,
  author  = {Kouki Sawada and Yasuhisa Hirata},
  title   = {Mapping and localization for autonomous ship using {LiDAR} {SLAM} on the sea},
  journal = {Journal of Marine Science and Technology},
  year    = {2023}
}

@inproceedings{huang2024multi,
  title={Multi-Robot Autonomous Exploration and Mapping Under Localization Uncertainty with Expectation-Maximization},
  author={Huang, Yewei and Lin, Xi and Englot, Brendan},
  booktitle={2024 IEEE International Conference on Robotics and Automation (ICRA)},
  pages={7236--7242},
  year={2024},
  organization={IEEE}
}

@article{kaess2012isam2,
  author  = {Michael Kaess and Hordur Johannsson and Richard Roberts and Viorela Ila and John J. Leonard and Frank Dellaert},
  title   = {{iSAM2}: Incremental Smoothing and Mapping Using the {B}ayes Tree},
  journal = {The International Journal of Robotics Research},
  volume  = {31},
  number  = {2},
  pages   = {216--235},
  month   = feb,
  year    = {2012},
}

@book{horn2013matrix,
  author    = {Roger A. Horn and Charles R. Johnson},
  title     = {Matrix Analysis},
  edition   = {2},
  publisher = {Cambridge University Press},
  year      = {2013}
}

@Article{makar2023harbour_gnss,
AUTHOR = {Makar, Artur},
TITLE = {Limitations of Multi-{GNSS} Positioning of {USV} in Area with High Harbour Infrastructure},
JOURNAL = {Electronics},
VOLUME = {12},
YEAR = {2023},
NUMBER = {3},
ARTICLE-NUMBER = {697},
ISSN = {2079-9292},
}

@inproceedings{pandele2020maritime,
  title={Maritime Environment GNSS Multipath Analysis in the Framework of the MARGOT Project},
  author={Pandele, Alexandru and Croitoru, Antonia and Hulea, Andrei and Cherciu, Costi and Radutu, Alina and Stefanescu, Irina and Urbanska, Katarzyna and Andrescu, Dumitru and Dragasanu, Claudiu and Trusculescu, Marius and others},
  booktitle={Proceedings of the 33rd International Technical Meeting of the Satellite Division of The Institute of Navigation (ION GNSS+ 2020)},
  pages={757--778},
  year={2020}
}

@Article{wang2024usv_laserslam,
AUTHOR = {Wang, Yang and Liu, Chao and Liu, Jiahe and Wang, Jinzhe and Liu, Jianbin and Zheng, Kai and Zheng, Rencheng},
TITLE = {A Laser-Based SLAM Algorithm of the Unmanned Surface Vehicle for Accurate Localization and Mapping in an Inland Waterway Scenario},
JOURNAL = {Journal of Marine Science and Engineering},
VOLUME = {12},
YEAR = {2024},
NUMBER = {12},
ARTICLE-NUMBER = {2311},
ISSN = {2077-1312},
}

@article{marchel2020slam_nav_aids,
  author  = {Marchel, {\L}ukasz and Naus, Krzysztof and Specht, Mariusz},
  title   = {Optimisation of the Position of Navigational Aids for the Purposes of {SLAM} Technology for Accuracy of Vessel Positioning},
  journal = {Journal of Navigation},
  year    = {2020},
  volume  = {73},
  number  = {2},
  pages   = {282--295},
}

@INPROCEEDINGS{stachniss2004activeloopclosing,
  author={Stachniss, C. and Hahnel, D. and Burgard, W.},
  booktitle={2004 IEEE/RSJ International Conference on Intelligent Robots and Systems (IROS) (IEEE Cat. No.04CH37566)}, 
  title={Exploration with active loop-closing for FastSLAM}, 
  year={2004},
  volume={2},
  number={},
  pages={1505-1510 vol.2},
}

@article{Nelson-2018-120036,
  author  = {Erik Nelson and Micah Corah and Nathan Michael},
  title   = {Environment Model Adaptation for Mobile Robot Exploration},
  journal = {Autonomous Robots},
  volume  = {42},
  number  = {5},
  pages   = {257--272},
  year    = {2018}
}

@article{forster2016manifold,
  title={On-manifold preintegration for real-time visual--inertial odometry},
  author={Forster, Christian and Carlone, Luca and Dellaert, Frank and Scaramuzza, Davide},
  journal={IEEE Transactions on Robotics},
  volume={33},
  number={1},
  pages={1--21},
  year={2016},
  publisher={IEEE}
}

@inproceedings{wang2019autonomous,
  title={Autonomous exploration with expectation-maximization},
  author={Wang, Jinkun and Englot, Brendan},
  booktitle={Robotics Research: The 18th International Symposium ISRR},
  pages={759--774},
  year={2019},
  organization={Springer}
}

@inproceedings{urmson2003approaches,
  title={Approaches for heuristically biasing {RRT} growth},
  author={Urmson, Chris and Simmons, Reid},
  booktitle={2003 IEEE/RSJ International Conference on Intelligent Robots and Systems (IROS)(Cat. No. 03CH37453)},
  volume={2},
  pages={1178--1183},
  year={2003},
  organization={IEEE}
}

@inproceedings{leung2006active,
  title={Active {SLAM} using model predictive control and attractor based exploration},
  author={Leung, Cindy and Huang, Shoudong and Dissanayake, Gamini},
  booktitle={2006 IEEE/RSJ International Conference on Intelligent Robots and Systems (IROS)},
  pages={5026--5031},
  year={2006},
  organization={IEEE}
}

@article{kriegel2015efficient,
  title={Efficient next-best-scan planning for autonomous 3D surface reconstruction of unknown objects},
  author={Kriegel, Simon and Rink, Christian and Bodenm{\"u}ller, Tim and Suppa, Michael},
  journal={Journal of Real-Time Image Processing},
  volume={10},
  number={4},
  pages={611--631},
  year={2015},
  publisher={Springer}
}

@inproceedings{carlone2010application,
  title={An application of Kullback-Leibler divergence to active SLAM and exploration with particle filters},
  author={Carlone, Luca and Du, Jingjing and Ng, Miguel Kaouk and Bona, Basilio and Indri, Marina},
  booktitle={2010 IEEE/RSJ International Conference on Intelligent Robots and Systems},
  pages={287--293},
  year={2010},
  organization={IEEE}
}

@inproceedings{bai2017toward,
  title={Toward autonomous mapping and exploration for mobile robots through deep supervised learning},
  author={Bai, Shi and Chen, Fanfei and Englot, Brendan},
  booktitle={2017 IEEE/RSJ International Conference on Intelligent Robots and Systems (IROS)},
  pages={2379--2384},
  year={2017},
  organization={IEEE}
}

@inproceedings{cao2025dare,
  title={Dare: Diffusion policy for autonomous robot exploration},
  author={Cao, Yuhong and Lew, Jeric and Liang, Jingsong and Cheng, Jin and Sartoretti, Guillaume},
  booktitle={2025 IEEE International Conference on Robotics and Automation (ICRA)},
  pages={11987--11993},
  year={2025},
  organization={IEEE}
}

@article{placed2023survey,
  title={A survey on active simultaneous localization and mapping: State of the art and new frontiers},
  author={Placed, Julio A and Strader, Jared and Carrillo, Henry and Atanasov, Nikolay and Indelman, Vadim and Carlone, Luca and Castellanos, Jos{\'e} A},
  journal={IEEE Transactions on Robotics},
  volume={39},
  number={3},
  pages={1686--1705},
  year={2023},
  publisher={IEEE}
}

@article{asgharivaskasi2025riemannian,
  title={Riemannian optimization for active mapping with robot teams},
  author={Asgharivaskasi, Arash and Girke, Fritz and Atanasov, Nikolay},
  journal={IEEE Transactions on Robotics},
  year={2025},
  publisher={IEEE},
  pages={1077--1097}
}

@article{wang2022virtual,
  title={Virtual maps for autonomous exploration of cluttered underwater environments},
  author={Wang, Jinkun and Chen, Fanfei and Huang, Yewei and McConnell, John and Shan, Tixiao and Englot, Brendan},
  journal={IEEE Journal of Oceanic Engineering},
  volume={47},
  number={4},
  pages={916--935},
  year={2022},
  publisher={IEEE}
}

@Article{shen2023usvfusion,
AUTHOR = {Shen, Wei and Yang, Zhisong and Yang, Chaoyu and Li, Xin},
TITLE = {A LiDAR SLAM-Assisted Fusion Positioning Method for USVs},
JOURNAL = {Sensors},
VOLUME = {23},
YEAR = {2023},
NUMBER = {3},
ARTICLE-NUMBER = {1558},
PubMedID = {36772598},
ISSN = {1424-8220},
}

@inproceedings{connolly1985determination,
  title={The determination of next best views},
  author={Connolly, Cl},
  booktitle={1985 IEEE International Conference on Robotics and Automation (ICRA)},
  volume={2},
  pages={432--435},
  year={1985},
  organization={IEEE}
}

@article{song2024efficient,
  title={An Efficient Autonomous Exploration Framework for Unmanned Surface Vehicles in Unknown Waters.},
  author={Song, Baojian and Zhang, Jiahao and Han, Xinjie and Fan, Yunsheng and Sun, Zhe and Wang, Yingjie},
  journal={Journal of Marine Science \& Engineering},
  volume={12},
  number={9},
  year={2024}
}

@article{carrillo2018autonomous,
  title={Autonomous robotic exploration using a utility function based on R{\'e}nyi’s general theory of entropy},
  author={Carrillo, Henry and Dames, Philip and Kumar, Vijay and Castellanos, Jos{\'e} A},
  journal={Autonomous Robots},
  volume={42},
  number={2},
  pages={235--256},
  year={2018},
  publisher={Springer}
}

@inproceedings{bourgault2002information,
  title={Information based adaptive robotic exploration},
  author={Bourgault, Frederic and Makarenko, Alexei A and Williams, Stefan B and Grocholsky, Ben and Durrant-Whyte, Hugh F},
  booktitle={2002 IEEE/RSJ International Conference on Intelligent Robots and Systems (IROS)},
  volume={1},
  pages={540--545},
  year={2002},
  organization={IEEE}
}

@inproceedings{saulnier2020information,
  title={Information theoretic active exploration in signed distance fields},
  author={Saulnier, Kelsey and Atanasov, Nikolay and Pappas, George J and Kumar, Vijay},
  booktitle={2020 IEEE International Conference on Robotics and Automation (ICRA)},
  pages={4080--4085},
  year={2020},
  organization={IEEE}
}

@inproceedings{oleynikova2017voxblox,
  title={Voxblox: Incremental 3d euclidean signed distance fields for on-board mav planning},
  author={Oleynikova, Helen and Taylor, Zachary and Fehr, Marius and Siegwart, Roland and Nieto, Juan},
  booktitle={2017 IEEE/RSJ International Conference on Intelligent Robots and Systems (IROS)},
  pages={1366--1373},
  year={2017},
  organization={IEEE}
}

@inproceedings{vallve2014dense,
  title={Dense entropy decrease estimation for mobile robot exploration},
  author={Vallv{\'e}, Joan and Andrade-Cetto, Juan},
  booktitle={2014 IEEE International Conference on Robotics and Automation (ICRA)},
  pages={6083--6089},
  year={2014},
  organization={IEEE}
}

@inproceedings{vallve2015active,
  title={Active pose SLAM with RRT},
  author={Vallv{\'e}, Joan and Andrade-Cetto, Juan},
  booktitle={2015 IEEE International Conference on Robotics and Automation (ICRA)},
  pages={2167--2173},
  year={2015},
  organization={IEEE}
}

@inproceedings{popovic2020informative,
  title={Informative path planning for active field mapping under localization uncertainty},
  author={Popovi{\'c}, Marija and Vidal-Calleja, Teresa and Chung, Jen Jen and Nieto, Juan and Siegwart, Roland},
  booktitle={2020 IEEE International Conference on Robotics and Automation (ICRA)},
  pages={10751--10757},
  year={2020},
  organization={IEEE}
}

@article{cao2024deep,
  title={Deep reinforcement learning-based large-scale robot exploration},
  author={Cao, Yuhong and Zhao, Rui and Wang, Yizhuo and Xiang, Bairan and Sartoretti, Guillaume},
  journal={IEEE Robotics and Automation Letters},
  volume={9},
  number={5},
  pages={4631--4638},
  year={2024},
  publisher={IEEE}
}

@inproceedings{chen2025gleam,
  title={GLEAM: Learning Generalizable Exploration Policy for Active Mapping in Complex 3D Indoor Scene},
  author={Chen, Xiao and Wang, Tai and Li, Quanyi and Huang, Tao and Pang, Jiangmiao and Xue, Tianfan},
  booktitle={Proceedings of the IEEE/CVF International Conference on Computer Vision},
  pages={5558--5568},
  year={2025}
}

@INPROCEEDINGS{zhu2018deep,
  author={Zhu, Delong and Li, Tingguang and Ho, Danny and Wang, Chaoqun and Meng, Max Q.-H.},
  booktitle={2018 IEEE International Conference on Robotics and Automation (ICRA)}, 
  title={Deep Reinforcement Learning Supervised Autonomous Exploration in Office Environments}, 
  year={2018},
  volume={},
  number={},
  pages={7548-7555},
}

@ARTICLE{li2025learning,
  author={Li, Zhi and Zheng, Kairao and Yuan, Yiqing and Huang, Junlong and Zhang, Xiaoxun and Wu, Jinze and Cheng, Hui},
  journal={IEEE Robotics and Automation Letters}, 
  title={Learning to Explore Efficiently: Heterogeneous Topological Graphs and Lightweight Global Reasoning for Robotic Exploration}, 
  year={2025},
  volume={10},
  number={12},
  pages={12357-12364},
}

@ARTICLE{ral2019NBV,
  author={Selin, Magnus and Tiger, Mattias and Duberg, Daniel and Heintz, Fredrik and Jensfelt, Patric},
  journal={IEEE Robotics and Automation Letters}, 
  title={Efficient Autonomous Exploration Planning of Large-Scale 3-D Environments}, 
  year={2019},
  volume={4},
  number={2},
  pages={1699-1706},
}

@article{ijrr2013efficient,
  title={Efficient frontier detection for robot exploration},
  author={Keidar, Matan and Kaminka, Gal A.},
  journal={The International Journal of Robotics Research},
  volume={33},
  number={2},
  pages={215--236},
  year={2013},
  publisher={SAGE Publications Sage UK: London, England},
}

@book{fossen2011handbook,
  title={Handbook of marine craft hydrodynamics and motion control},
  author={Fossen, Thor I},
  year={2011},
  publisher={John wiley \& sons}
}

@article{cheng2025asymmetric,
  title={Asymmetric information enhanced mapping framework for multirobot exploration based on deep reinforcement learning},
  author={Cheng, Jiyu and Fan, Junhui and Li, Xiaolei and Rosin, Paul L and Li, Yibin and Zhang, Wei},
  journal={IEEE Transactions on Robotics},
  year={2025},
  publisher={IEEE},
  pages={6250--6266}
}

@article{wang2024exploration,
  title={An exploration-enhanced search algorithm for robot indoor source searching},
  author={Wang, Miao and Xin, Bin and Jing, Mengjie and Qu, Yun},
  journal={IEEE Transactions on Robotics},
  volume={40},
  pages={4160--4178},
  year={2024},
  publisher={IEEE}
}

@article{zhang2024falcon,
  title={Falcon: Fast autonomous aerial exploration using coverage path guidance},
  author={Zhang, Yichen and Chen, Xinyi and Feng, Chen and Zhou, Boyu and Shen, Shaojie},
  journal={IEEE Transactions on Robotics},
  volume={41},
  pages={1365--1385},
  year={2024},
  publisher={IEEE}
}

@article{lindqvist2024tree,
  title={A tree-based next-best-trajectory method for 3-D UAV exploration},
  author={Lindqvist, Bj{\"o}rn and Patel, Akash and L{\"o}fgren, Kalle and Nikolakopoulos, George},
  journal={IEEE Transactions on Robotics},
  volume={40},
  pages={3496--3513},
  year={2024},
  publisher={IEEE}
}

@article{zheng2025aage,
  title={AAGE: Air-assisted ground robotic autonomous exploration in large-scale unknown environments},
  author={Zheng, Lanxiang and Wei, Mingxin and Mei, Ruidong and Xu, Kai and Huang, Junlong and Cheng, Hui},
  journal={IEEE Transactions on Robotics},
  year={2025},
  publisher={IEEE},
  volume={41},
  number={},
  pages={1918-1937},
}

@article{border2024surface,
  title={The surface edge explorer (see): A measurement-direct approach to next best view planning},
  author={Border, Rowan and Gammell, Jonathan D},
  journal={The International Journal of Robotics Research},
  volume={43},
  number={10},
  pages={1506--1532},
  year={2024},
  publisher={SAGE Publications Sage UK: London, England}
}

@article{best2024multi,
  title={Multi-robot, multi-sensor exploration of multifarious environments with full mission aerial autonomy},
  author={Best, Graeme and Garg, Rohit and Keller, John and Hollinger, Geoffrey A and Scherer, Sebastian},
  journal={The International Journal of Robotics Research},
  volume={43},
  number={4},
  pages={485--512},
  year={2024},
  publisher={SAGE Publications Sage UK: London, England}
}

@article{brugali2025mobile,
  title={Mobile robots exploration strategies and requirements: A systematic mapping study},
  author={Brugali, Davide and Muratore, Luca and De Luca, Alessio},
  journal={The International Journal of Robotics Research},
  volume={44},
  number={9},
  pages={1461--1506},
  year={2025},
  publisher={SAGE Publications Sage UK: London, England}
}

@article{vutetakis2025active,
  title={Active perception network for non-myopic online exploration and visual surface coverage},
  author={Vutetakis, David and Xiao, Jing},
  journal={The International Journal of Robotics Research},
  volume={44},
  number={2},
  pages={247--272},
  year={2025},
  publisher={Sage Publications Sage UK: London, England}
}

@article{chen2024adaptive,
  title={Adaptive robotic information gathering via non-stationary Gaussian processes},
  author={Chen, Weizhe and Khardon, Roni and Liu, Lantao},
  journal={The International Journal of Robotics Research},
  volume={43},
  number={4},
  pages={405--436},
  year={2024},
  publisher={SAGE Publications Sage UK: London, England}
}

@article{xu2022fastlio2,
  author  = {Xu, Wei and Cai, Yixi and He, Dongjiao and Lin, Jiarong and Zhang, Fu},
  title   = {{FAST-LIO2}: Fast Direct {LiDAR}-Inertial Odometry},
  journal = {IEEE Transactions on Robotics},
  year    = {2022},
  volume  = {38},
  number  = {4},
  pages   = {2053--2073},
  doi     = {10.1109/TRO.2022.3141876}
}

\end{document}